\documentclass{article}

\PassOptionsToPackage{numbers, compress}{natbib}

\usepackage[preprint]{neurips_2026}

\usepackage[utf8]{inputenc} 
\usepackage[T1]{fontenc}    
\usepackage{hyperref}       
\usepackage{url}            
\usepackage{booktabs}       
\usepackage{amsfonts}       
\usepackage{nicefrac}       
\usepackage{microtype}      
\usepackage{xcolor}         

\usepackage{graphicx}
\usepackage{fontawesome5}
\usepackage{siunitx}
\DeclareSIUnit{\pixel}{px}
\DeclareSIUnit{\nothing}{\relax}
\usepackage{cleveref}
\usepackage{enumitem}
\usepackage{xspace}
\usepackage{float}
\usepackage[super]{nth}
\usepackage{pgf}
\usepackage{rotating}
\usepackage{multirow}
\usepackage{marvosym}
\usepackage{placeins}
\usepackage{makecell}

\usepackage{graphicx}   
\usepackage{subcaption} 
\usepackage{mwe}
\usepackage[percent]{overpic} 
\makeatletter
\DeclareRobustCommand\onedot{\futurelet\@let@token\@onedot}
\def\@onedot{\ifx\@let@token.\else.\null\fi\xspace}

\def\eg{\emph{e.g}\onedot}

\def\etc{\emph{etc}\onedot}

\def\etal{\emph{et al}\onedot}
\makeatother

\title{The Road Ahead in Autonomous Driving:\\ The KITScenes Multimodal Dataset}

\author{%
Richard Schwarzkopf$^{1,2}$\thanks{Joint first authors} \quad Fabian Immel$^{1,2}$$^{\ast}$\\[1mm]
\textbf{Alexander Blumberg$^{2}$ \quad Jonas Merkert$^2$ \quad Nils Rack$^{2}$ \quad Kaiwen Wang$^{2}$}\\
\textbf{Fabian Konstantinidis$^{2}$ \quad Julian Truetsch$^{1,2}$ \quad Carlos Fernandez$^{2}$ \quad Annika Bätz$^{2}$}\\
\textbf{Kevin Rösch$^{1,2}$ \quad Marlon Steiner$^{2}$ \quad Willi Poh$^{2}$ \quad  Yinzhe Shen$^{2}$ \quad Royden Wagner$^{2}$ }\\
\textbf{\quad Felix Hauser$^{2}$ \quad Dominik Strutz$^{2}$ \quad Jaime Villa$^{3}$ \quad Gleb Stepanov$^{2}$}\\
\textbf{Holger Caesar$^4$ \quad Ömer Şahin Taş$^{1,2}$ \quad Frank Bieder$^{1,2}$} \\[1mm]
\textbf{Jan-Hendrik Pauls$^{2}$\thanks{Project lead and \Letter\ Corresponding author}\, $\text{\Letter}$ \quad Christoph Stiller$^{1,2}$}\\[2mm]
$^1$FZI Research Center for Information Technology \quad $^2$Karlsruhe Institute of Technology\\
\qquad\quad\texttt{\{immel, schwarzkopf\}@fzi.de} \qquad \qquad\ \ \texttt{jan-hendrik.pauls@kit.edu}\\[1mm]
 \quad $^3$ University Charles III of Madrid \quad $^4$Delft University of Technology
}

\begin{document}

\maketitle

\begin{abstract}

Existing autonomous driving datasets have enabled major progress, but fall short in sensor fidelity, map completeness, or geographic diversity.
We present KITScenes Multimodal, a European dataset built around high-fidelity sensors and maps.
Our fully synchronized sensor suite combines high-resolution global-shutter cameras, long-range lidar beyond \SI{400}{\meter}, 4D imaging radar, and redundant GNSS/INS localization.
Our HD maps are, to our knowledge, the most complete of any sensor dataset, validated through autonomous driving trials on open-source software.
For the first time in a public dataset, all driving-relevant traffic elements, such as traffic lights, are mapped in 3D to a reprojection-accurate level with full topological connectivity.
Recorded in cities with irregular street layouts and mixed traffic modes, our dataset complements existing datasets by broadening the available geographic diversity. 
We also introduce four benchmarks, each advancing spatial learning for embodied AI: online HD map construction, long-range depth estimation, novel view synthesis, and end-to-end driving.
Project page: \url{https://kitscenes.com/multimodal/}.

\end{abstract}

\section{Introduction}

Autonomous driving datasets~\cite{geiger2012kitti, caesar2020nuscenes, sun2020waymo_perception, ettinger2021waymo_motion} have enabled significant progress in both computer vision and autonomous driving research.
However, existing datasets still fall short of capturing the complexity required for spatially aware driving in dense urban environments.
Some lack public annotations or topology-aware map references~\cite{caesar2020nuscenes, sun2020waymo_perception}, while others focus on comparatively simple driving scenarios such as motorways~\cite{fent2024truckscenes, ghilotti2026truckdrive}.
As autonomous driving systems move toward deeper spatial understanding, datasets must support reasoning not only about objects, but also about geometry, road structure, and their geospatial relationships.
High-fidelity datasets enriched with geospatial annotations, HD maps, and 3D labels are essential for evaluating such capabilities.

High-fidelity datasets with geospatial annotations have a limited geographic footprint, with coverage heavily skewed toward North America and Asia.
KITTI~\cite{geiger2012kitti}, though seminal, is small-scale; ZOD~\cite{alibeigi2023zenseact} annotates only single keyframes with image-space labels; and large-scale recording efforts such as those from Nvidia~\cite{nvidia2025physicalai_av} still lack public annotations.
Consequently, complex European urban environments remain underrepresented in current autonomous driving benchmarks, arguably being the most difficult to spatially reason about.
This leaves a clear need for datasets that combine high-fidelity sensing, complete geospatial context, and dense 3D annotations.

\begin{figure}
    \centering
    \includegraphics[width=1\linewidth]{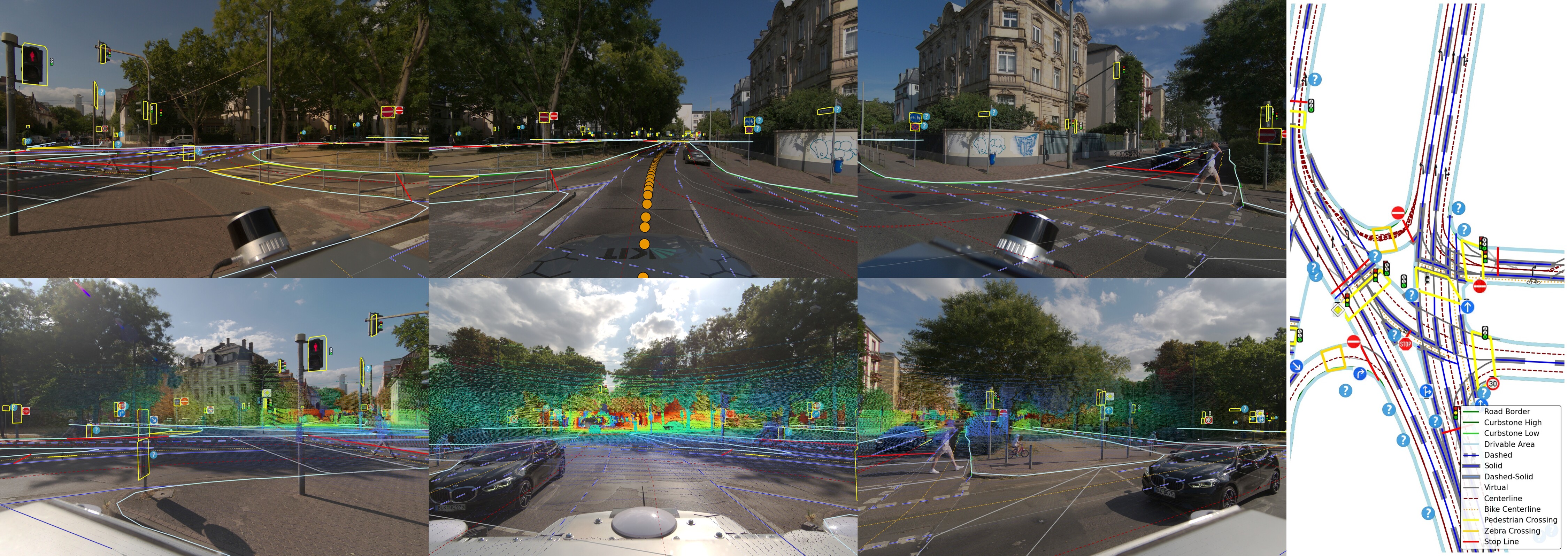}
    \caption{A showcase of 3D HD map elements and the ground truth reprojected into 6 out of 9 cameras, the dense long-range lidar pointcloud reprojected into the rear cameras, and a top-view of the HD map benchmark labels. Best viewed zoomed in.}
    \label{fig:teaser}
\end{figure}

We present \textbf{KITScenes Multimodal}, a dataset recorded across diverse European urban environments using a state-of-the-art robotaxi sensor platform.
Our dataset addresses the geographic gap in existing benchmarks while simultaneously raising the bar on both sensor fidelity and geospatial understanding.
Our sensor platform combines high-resolution cameras (up to \SI{16.2}{\mega\pixel}), long-range lidar with effective range beyond \SI{400}{\meter}, 4D imaging radar, and redundant GNSS, all hardware-synchronized and processed with high-fidelity pipelines that make the data suitable for applications such as neural rendering and novel view synthesis.
Besides high fidelity sensor data, we provide the most complete HD maps of any public autonomous driving dataset.
Annotated in Lanelet2~\cite{poggenhans2018lanelet2}, our maps visualized in \Cref{fig:teaser} cover all regulatory road feature and traffic sign classes, and host our annotated 3D traffic lights, signs, and poles with reprojection-accurate localization.

To demonstrate the unique strengths of the dataset, we introduce four benchmarks: (1)~\textbf{Complete online HD map perception}, evaluating relational Lanelet2 map prediction from sensor data; (2)~\textbf{long-range depth estimation}, targeting depth beyond \SI{200}{\meter} where current methods degrade severely; (3)~\textbf{novel view synthesis}, exploiting our high-fidelity imagery and dense lidar for 3D scene reconstruction; and (4)~\textbf{multimodal end-to-end models for autonomous driving}, predicting future trajectories and scene evolution from camera, lidar, and radar inputs.
Our contributions include:
\begin{itemize}
  \item A multimodal European driving dataset, 
   recorded in three cities with a high-fidelity robotaxi sensor suite: \SI{72.5}{\mega\pixel} of synchronized global-shutter cameras, seven lidars with over $3\times$ the point density and nearly double the range of the next closest dataset, three 4D imaging radars, and redundant GNSS/INS.
  \item Production-grade Lanelet2 HD maps covering \SI{60}{\kilo\meter\squared} 
        with 29 road-feature classes, 120 traffic-sign classes, and 3D traffic lights, signs, and poles localized to reprojection accuracy.
        The maps include all regulatory elements required for autonomous navigation and are validated for use in the open-source Autoware~\cite{autoware} stack, both online and in simulation. 
  \item Four benchmarks designed to expose the limits of current methods on the path to Level 4 autonomy, targeting capabilities existing datasets cannot benchmark at this fidelity: holistic HD map prediction, depth estimation beyond
  \SI{200}{\meter}, high-fidelity novel view synthesis, and multi-modal end-to-end driving.
\end{itemize}

\section{Related Work}

\paragraph{Autonomous Driving Datasets for Perception.}

The past decade has seen a rapid growth of autonomous driving datasets.
Foundational datasets such as nuScenes~\cite{caesar2020nuscenes}, Waymo Open~\cite{sun2020waymo_perception}, and Argoverse~2~\cite{wilson2021argoverse2} established the multimodal paradigm with complementary sensor configurations and annotation schemes.
Further datasets~\cite{huang2018apolloscape,mao2021once} broaden the range of traffic layouts and driving conditions, although detailed map annotations and deployment-oriented perception support remain limited.
KITTI~\cite{geiger2012kitti} and KITTI-360~\cite{Liao2022PAMI} remain influential but limited in scale and sensor diversity by current standards.
ZOD~\cite{alibeigi2023zenseact} provides large-scale recordings, yet annotates only a single keyframe per scenario and mainly provides image-space labels.
MAN TruckScenes~\cite{fent2024truckscenes} focuses on motorway trucking rather than complex urban perception.
While TruckDrive~\cite{ghilotti2026truckdrive} features long-range sensors, it likewise targets trucking scenarios, relies on automotive RCCB cameras, and has not released any public data to date.
Large-scale fleet recordings such as Nvidia Physical AI AV~\cite{nvidia2025physicalai_av} provide broad real-world coverage but lack public annotations.
A quantified comparison of the sensor setups is shown in \Cref{tab:sensor_comparison}.

\paragraph{HD Maps and Map Perception Benchmarks.}

Map representations accompanying public datasets vary substantially in completeness.
nuScenes~\cite{caesar2020nuscenes} and Argoverse~2~\cite{wilson2021argoverse2} expose lane geometry via dataset-specific APIs but omit regulatory structure from traffic lights and signs. OpenLane-V2~\cite{wang2023openlanev2} adds lane-topology links, but as image-space annotations rather than metric 3D maps.
To our knowledge, no prior dataset provides HD maps that are simultaneously reprojection-accurate, complete in regulatory structure (traffic signs, lights, lane assignments), and validated in a planning stack (\Cref{tab:hd_map_comparison}).
As a consequence, so far online HD map construction methods~\cite{li2022hdmapnet,liu2023vectormapnet,liao2022maptr,maptrv2,yuan2024streammapnet,qiao2023bemapnet,ding2023pivotnet,wang2024stream_sqd_mapnet,chen2024maptracker,zhang2024enhancing_HR_mapnet,shi2024globalmapnet,zhang2025mapexpert,yang2025histrackmap,erdougan2025mapping_skeptic} are evaluated on simple geometric primitives only (lane dividers without type, pedestrian crossings, road borders).
Lanelet2~\cite{poggenhans2018lanelet2} has emerged as the open academic standard for HD maps, encoding geometry, topology, and 3D regulatory elements in a single graph; it is the native input of Autoware~\cite{autoware} and translatable to learning-friendly representations using~\cite{immel2024lanelet2mlconverter}.

\paragraph{Long-range Perception, Neural Rendering, and End-to-End Driving.}

Depth estimation is predominantly benchmarked on KITTI~\cite{geiger2012kitti} and DDAD~\cite{guizilini2020ddad}; recent foundation models~\cite{depthanything3,ganesan2026unidacuniversalmetricdepth} achieve strong near-range performance, but existing benchmarks rarely assess depth beyond 80--100\,m.
Neural scene representations for driving like NeRF-based~\cite{wu2023mars,yang2024emernerf} and 3D Gaussian Splatting methods~\cite{yan2024street,chen2025omnire,yu2026_recondrive}, are similarly constrained by input image fidelity and lidar density.
End-to-end driving models~\cite{hu2023uniad,jiang2023vad} and world models are evaluated on a small family of closely related datasets~\cite{caesar2020nuscenes,caesar2021nuplan,openscene2023,dauner2024navsim}, limiting the sensor configurations and geographies under which they are assessed.

\section{The KITScenes Multimodal Dataset}

\subsection{High-Resolution Long-Range Multi-Modal Sensor Setup}
\label{sec:dataset:sensors}

\begin{figure}[t]
  \centering
  \includegraphics[width=\linewidth]{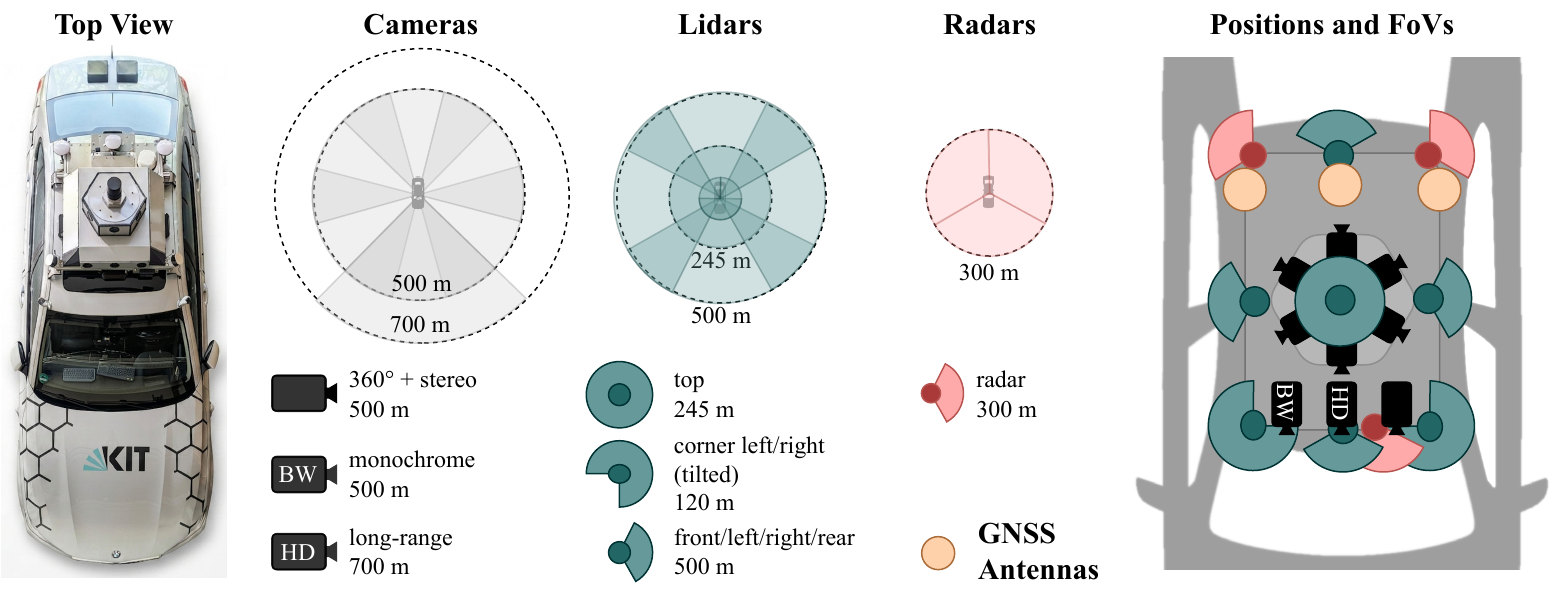}
  \caption{KITScenes Multimodal Sensor Setup. Our sensor rack (left) is depicted along with nominal sensing range (center), as well as sensor positions and their field of view (right).
  }
  \label{fig:sensor_setup}
\end{figure}

KITScenes Multimodal uses a fully synchronized sensor suite.
\Cref{fig:sensor_setup} depicts the sensor positions and their nominal fields of view. To enable sensor fusion up to maximum effective sensing range, we perform intrinsic and extrinsic calibration across all modalities, achieving subpixel intrinsic and \SI{1}{\centi\meter} and \SI{0.1}{\degree} extrinsic accuracy. Further details are listed in \Cref{sec:appx:sensor_setup_details} and \Cref{sec:appx:calibration_details}.

\paragraph{Cameras.}
The camera suite comprises six \SI{7.1}{\mega\pixel} surround cameras providing full \SI{360}{\degree} coverage, one \SI{16.2}{\mega\pixel} high-resolution long-range camera, and a tilted forward-facing stereo setup, yielding a combined resolution of \SI{72.5}{\mega\pixel} per frame, which is more than twice that of the next closest dataset (\Cref{tab:sensor_comparison}).
Existing setups put their focus on dynamic object perception~\cite{caesar2020nuscenes,sun2020waymo_perception,fent2024truckscenes,wilson2021argoverse2}, triggering the cameras when the lidar swept across the image center to ensure a minimal delay between both modalities.
All cameras use global shutter sensors and are hardware-synchronized, ensuring pixel-accurate temporal alignment. The images are anonymized and compressed with JPEGLI~\cite{szabadka2024jpegli}, a state-of-the-art visually lossless codec described in \Cref{appx:sensor_processing_privacy}.
This is the foundation for our high fidelity ground truth for neural rendering and novel view synthesis.
At the same time, we ensure lidar coverage by redundantly combining multiple lidars with varying sweeping directions.

\paragraph{Lidar.}
Seven lidar sensors provide \SI{360}{\degree} coverage with substantial overlap between adjacent units.
As shown in \Cref{tab:sensor_comparison}, the fused point cloud contains on average more than \SI{900}{\kilo\nothing} points per frame with peaks above \SI{1.2}{\mega\nothing} points, tripling the effective point density over existing datasets.
The use of \SI{1550}{\nano\meter} lidars enables an average maximum range of more than \SI{400}{\meter}, nearly doubling that of the next-best dataset.
This long-range capability is essential for both online long-range perception and for providing ground truth for benchmarks, such as depth estimation.
\Cref{fig:lidar_point_distribution} compares the per-distance-bin return density for KITScenes and existing autonomous driving datasets, showing that KITScenes provides higher effective point density in every bin and extends usable range beyond \SI{250}{\meter}, where prior datasets fall to zero.

\begin{table*}
\caption{KITScenes Multimodal sets a new state of the art for temporally consistent high-resolution high-fidelity RGB surround vision, highly dense long-range lidar, and ranging modality coverage. We triple the average lidar point density and almost double the typical maximum range; see also \Cref{fig:lidar_point_distribution}.}
\setlength{\tabcolsep}{3pt}
\label{tab:sensor_comparison}
\centering
\small
\resizebox{\linewidth}{!}{%
\begin{tabular}{l r r r r c c c c r r r r}
\toprule
& \multicolumn{7}{c}{\bfseries Cameras} & \bfseries Radar & \multicolumn{4}{c}{\bfseries Lidar} \\
\cmidrule(r){2-8}\cmidrule(lr){9-9}\cmidrule(l){10-13}
\textbf{Dataset}
  & \textbf{\faCamera} & \textbf{\faBinoculars} & \textbf{MPix} & \textbf{FOV}
  & \textbf{Shutter} & \textbf{Cam.\ Sync} & \textbf{Comp.}
  & \textbf{Config}
  & \textbf{\#} & \textbf{Avg pts} & \textbf{Max pts} & \textbf{Max range} \\
\midrule
nuScenes~\cite{caesar2020nuscenes}             & 6          & --         & 8.4            & \textbf{360$^\circ$} & Rolling          & to lidar             & JPEG                             & 5$\times$3D            & 1          & 34.7\,k           & 34.8\,k            & 102.1\,m \\
ONCE~\cite{mao2021once}                 & 7          & --         & 14.5           & \textbf{360$^\circ$} & Rolling          & to lidar             & JPEG                             & --            & 1          & 64.7\,k           & 69.7\,k            & 196.8\,m \\
nuPlan Sensors~\cite{caesar2021nuplan}               & \textbf{8} & --         & 19.2           & \textbf{360$^\circ$} & Rolling          & to lidar             & JPEG                             & --            & 5          & 93.0\,k           & 100.3\,k           & 215.5\,m \\
Argoverse 2 Sensor~\cite{wilson2021argoverse2}          & 7          & \textbf{1} & 28.6           & \textbf{360$^\circ$} & Rolling          & to lidar             & JPEG                            & --            & 2          & 96.9\,k           & 106.3\,k           & 217.4\,m \\
WOD Perception~\cite{sun2020waymo_perception}                  & 5          & --         & 10.4           & 230$^\circ$          & Rolling          & to lidar             & JPEG                             & --            & 5          & 175.5\,k          & 215.9\,k           & 75.0\,m  \\
MAN TruckScenes~\cite{fent2024truckscenes}     & 4          & --         & 9.3            & \textbf{360$^\circ$} & Rolling          & to lidar             & JPEG                             & 6$\times$4D           & 6          & 231.7\,k          & 296.7\,k           & 221.6\,m \\
Zenseact Open~\cite{alibeigi2023zenseact}        & 1          & --         & 8.3            & 120$^\circ$          & Rolling          & --                   & PNG                              & --            & 3          & 253.7\,k          & 311.1\,k           & 244.0\,m \\
Nvidia PhysicalAI AV~\cite{nvidia2025physicalai_av}     & 7          & --         & 14.5           & \textbf{360$^\circ$} & Rolling          & no                   & H.264                            & \textbf{9$\times$4D}   & 1          & 297.2\,k          & 344.1\,k           & 206.0\,m \\
\textbf{KITScenes Multimodal}   & 7          & \textbf{1} & \bfseries 72.5 & \textbf{360$^\circ$} & \bfseries Global & \textbf{all cameras} & JPEGLI & 3$\times$4D          & \textbf{7} & \textbf{906.4\,k} & \textbf{1235.2\,k} & \textbf{409.2\,m} \\
\bottomrule
\end{tabular}%
} 
\\[0.5ex]
\makebox[\linewidth][l]{\footnotesize \textbf{\faCamera} Monocular cameras, \textbf{\faBinoculars} Stereo camera pair, MPix = Total resolution per frame, Comp.\ = Image compression}
\end{table*}

\subsection{HD Map Annotation}
\label{sec:dataset:maps}

\begin{table}[t]
  \caption{%
    Comparison of related datasets comprised of HD maps and sensor data, datasets from \Cref{tab:sensor_comparison} without HD maps are not listed. 
    Legend: \textcolor{green!55!black}{\faCheck}\,yes, \textcolor{black}{\faCheck}\,partial/limited, \textcolor{red!75!black}{\faTimes} no; 
    ( ):~unreleased data;  %
    $\uparrow$:~large coverage  based on dataset description which is not reported or reproduced area coverage. 
  }
  \label{tab:hd_map_comparison}
  \centering
  \footnotesize
  \begingroup
  \newcommand{\y}{\textcolor{green!55!black}{\faCheck}}
  \newcommand{\n}{\textcolor{red!75!black}{\faTimes}}
  \newcommand{\p}{\textcolor{black}{\faCheck}}
  \newcommand{\open}{\mbox{---}}
  \newcommand{\ystar}{\textcolor{green!55!black}{\faCheck}\raisebox{0.35ex}{\scriptsize$*$}}
  \newcommand{\ymapnv}{( \textcolor{green!55!black!25!white}{\faCheck} )}
  \newcommand{\nmapnv}{( \textcolor{red!75!black!25!white}{\faTimes} )}
  \setlength{\tabcolsep}{2.2pt}
  \resizebox{\linewidth}{!}{%
  \begin{tabular}{l *{13}{c}}
    \toprule
    & \textbf{Dataset} & \shortstack{\textbf{Area}\\[-0.1em]\textbf{(\si{km^2})}} & \textbf{Region} & \shortstack{\textbf{All}\\[-0.08em]\textbf{sensors}} & \shortstack{\faCamera\\[-0.12em]\scriptsize360$^\circ$} & \shortstack{\textbf{3D}\\[-0.1em]\textbf{lanes}} & \shortstack{\textbf{Lane border}\\[-0.08em]\textbf{type}} & \shortstack{\textbf{Bike}\\[-0.1em]\textbf{Lanes}} & \shortstack{\textbf{3D Traffic}\\[-0.1em]\textbf{elements}} & \shortstack{\textbf{Full}\\[-0.1em]\textbf{topology}} & \shortstack{\textbf{Human}\\[-0.08em]\textbf{HD map}} & \shortstack{\textbf{OSS}\\[-0.08em]\textbf{AD stk.}} \\
    \midrule
    \multirow{4}{*}{\rotatebox[origin=c]{90}{\textbf{\parbox{1.4cm}{\centering\scriptsize Limited spa-\\tial learning}}}}
    & WOD Perception~\cite{sun2020waymo_perception} & 76\,km$^2$ & US & \y & \n & \y & \y & \y & \n & \n & \y & \n \\
    & nuPlan Sensors$^\dagger$~\cite{caesar2021nuplan} & $\uparrow$ & US, Asia & \y$^\dagger$ & \y & \n & \p & \n & \n & \p$^\dagger$ & \y & \n \\
    & AV2 TbV~\cite{av2_trust_but_verify} & 42\,km$^2$ & US & \n & \y & \y & \y & \y & \n & \n & \y & \n \\
    & Nvidia PhysicalAI AV~\cite{nvidia2025physicalai_av} & $\uparrow\uparrow\uparrow$ & US, EU & \y & \y & \ymapnv & \nmapnv & \nmapnv & \ymapnv & \nmapnv & \nmapnv & \n \\
    \specialrule{0.4pt}{0.35em}{0.35em}
    \multirow{4}{*}{\rotatebox[origin=c]{90}{\textbf{\parbox{1.4cm}{\centering\scriptsize Full spatial\\ learning}}}}
    & nuScenes~\cite{caesar2020nuscenes} & 5\,km$^2$ & US, Asia & \y & \y & \n & \p & \n & \n & \n & \y & \n \\
    & Argoverse 2 Sensor~\cite{wilson2021argoverse2} & 17\,km$^2$ & US & \y & \y & \y & \y & \y & \n & \n & \y & \n \\
    & OpenLane-V2~\cite{wang2023openlanev2} & 22\,km$^2$ & US, Asia & \n & \y & \p & \p & \n & \n & \p$^\dagger$ & \y & \n \\
    &  \textbf{KITScenes Multimodal} & 50\,km$^2$$^\dagger$ & EU & \y & \y & \y & \y & \y & \y & \y & \y & \y \\
    \bottomrule
  \end{tabular}%
  }
  \endgroup
  \vspace{0.45em}
  \begin{minipage}{\linewidth}
    \footnotesize\setlength{\parindent}{0pt}\setlength{\parskip}{0.35em}
    \textit{$^\dagger$Remarks:} 
    nuPlan Sensor~\cite{caesar2021nuplan}: shorthand for the 10\% of scenes in nuPlan with available sensor data. traffic light states available trough offline state estimation, no linkage to sensor data.
    NVIDIA PhysicalAI AV: entries transparent filled based on current publically available release plans, not verified. 
    OpenLaneV2: built on top of sensor data of AV2 and nuScenes, with limited set of labeled traffic element 2D bounding boxes in a visible range of 25x50m at 2Hz. 
    KITScenes area: union of 200\,m\,$\times$\,200\,m boxes around the driven poses, computed in the same way as for AV2 TbV; the annotated HD maps cover 60\,km$^2$.
    \textbf{All sensors}: full suite and quality of original sensor dataset available.
    \textbf{OSS AD stack}: Native support of HD map for simulation and closed-loop driving with open-source software autonomous driving stack.  \textbf{Full spatial learning}: support for full resolution multimodal 360$^\circ$ surround view learning with a at least a base set of BEV annotations. 
  \end{minipage}
\end{table}

We provide pixel-accurate 3D maps that can be directly used in the open-source Autoware~\cite{autoware} stack, both for simulation and real-world autonomous driving. 
All maps are annotated in Lanelet2~\cite{poggenhans2018lanelet2}, an established open-source format for relational semantic HD maps.
Beyond geometry, each map encodes the full regulatory structure required for autonomous driving: 
Road level polylines are annotated with one of 29 classes, (\eg, road border, dashed, zebra-crossing \etc) traffic signs are classified based on 220 German road traffic code classes~\cite{carnot2026gtsign} (with 120 observed), traffic lights types are grouped into four categories (car, bike, pedestrian, misc).
All traffic signs and lights are explicitly assigned to the lanes they govern via topological links in the Lanelet2 format.
Traffic lights, road signs, and poles are annotated based on lidar and camera data as 3D shapes including orientation that are reprojection-accurate to the calibrated camera images~\cite{pauls2021automatic}.
This reprojection accuracy directly connects map labels to image pixels, enabling HD map annotations to be used as pixel-level training signal for perception models without any additional alignment step, as shown in \Cref{sec:benchmarks:online_hd_map_perception}.
We provide routable SD maps, extracted from OpenStreetMap~\cite{openstreetmap} snapshots matching the recording times, which serve both as a source of routing information and as an HD map prior.

\subsection{Dataset Statistics}
\label{sec:dataset:statistics}

\begin{figure}[t]
    \centering
        \includegraphics[width=\linewidth]{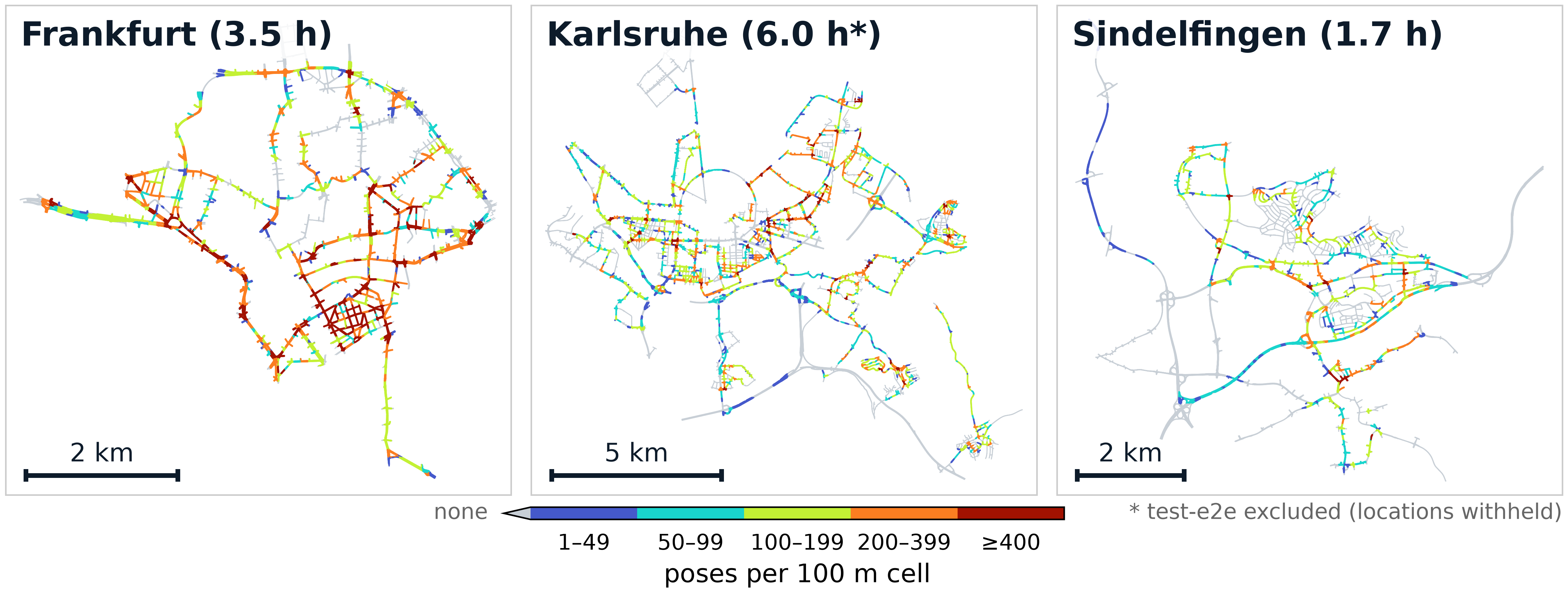}
    \caption{Spatial and temporal coverage for the three KITScenes Multimodal cities. The color indicates the number of poses at \SI{10}{\hertz} within a \SI{100}{m} grid cell over our HD map; gray is not covered.}
    \label{fig:spatial_coverage_density_FRA_KA}
\end{figure}

Our current release (v2.0.0) contains \num{2079} scenarios ranging from \qtyrange{7}{70}{\second}, totaling \SI{11.8}{\hour} and \SI{339}{\kilo\meter} of synchronized multimodal recording at \SI{10}{\hertz}.  
Details on the split and label statistics can be found in \Cref{sec:appx:statistics}.
The dataset currently spans Karlsruhe, Frankfurt, and Sindelfingen, chosen for their unique environments of a planned 18\textsuperscript{th} century radial layout, a dense metropolitan financial district core, and a suburban-industrial mix.
Our maps of a fourth city, Stuttgart, are withheld as a geographically separate split for future held-out test leaderboards.
\Cref{fig:spatial_coverage_density_FRA_KA} shows the coverage and density of the recordings, which took place from summer 2025 to spring 2026 to capture seasonal changes.
Additional visualizations are provided in \Cref{fig:scene_coverage,fig:campaign_overlap,fig:splits_FRA_KA}.

\section{Benchmarks}

Our benchmarks span spatial learning from map-level scene understanding to multimodal end-to-end driving.
They expose limitations of existing methods that prior datasets cannot reveal.

\subsection{Online HD Map Construction}
\label{sec:benchmarks:online_hd_map_perception}

\begin{figure}[t]
    \centering
    \begin{subfigure}[b]{0.56\linewidth}
        \centering
        \includegraphics[width=\linewidth]{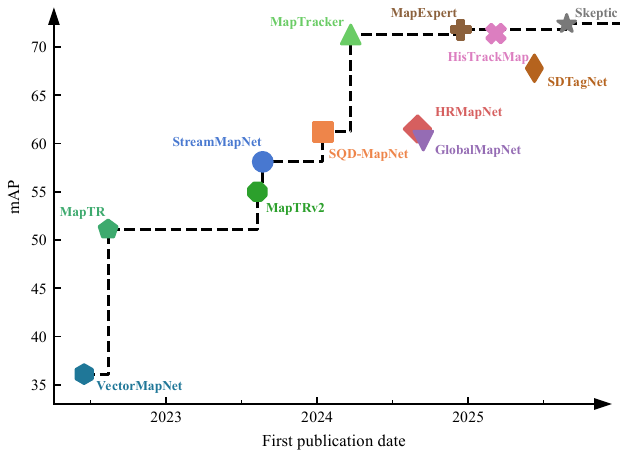}
        \caption{Historical SOTA progression of online HD map construction models~\cite{liu2023vectormapnet,liao2022maptr,maptrv2,yuan2024streammapnet,wang2024stream_sqd_mapnet,chen2024maptracker,zhang2024enhancing_HR_mapnet,shi2024globalmapnet,zhang2025mapexpert,yang2025histrackmap,immel2026sdtagnet,erdougan2025mapping_skeptic} on the AV2~\cite{wilson2021argoverse2} geo-split of~\cite{yuan2024streammapnet} at \qtyproduct{60x30}{\meter} range. Performance saturated on the current datasets and task complexity.}
        \label{fig:history_online_hd_map_construction}
    \end{subfigure}
    \hfill
    \begin{subfigure}[b]{0.42\linewidth}
        \centering
        \includegraphics[width=\linewidth]{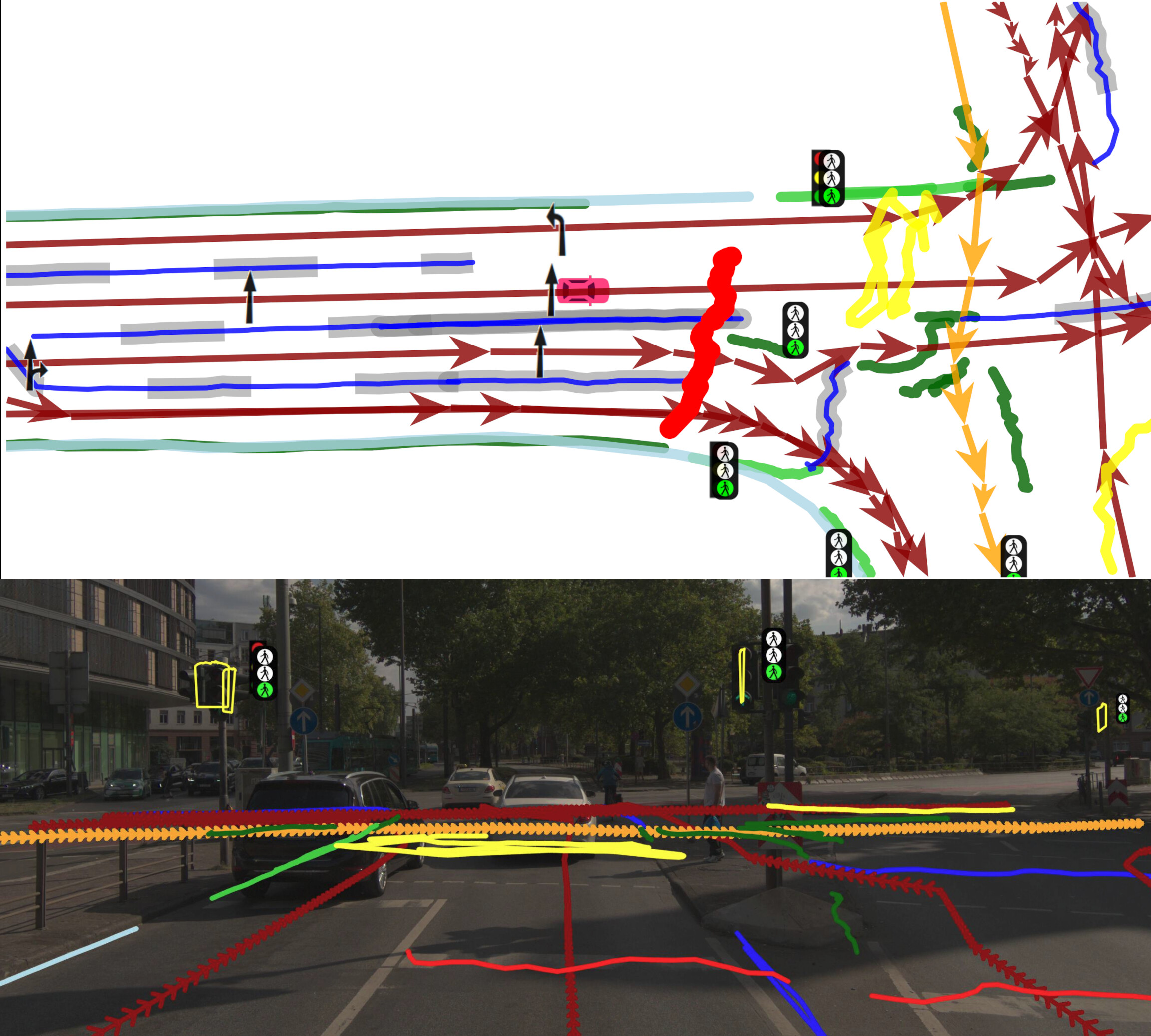}
        \caption{Example online HD map construction prediction of SDTagNet~\cite{immel2026sdtagnet} on a validation sample. While showing new capabilities such as 3D detection of non-ground elements thanks to the extensive map labels, a large gap in prediction of complete 3D HD maps remains.}
        \label{fig:example_online_hd_map_construction}
    \end{subfigure}
    \caption{Historical progression of online HD map construction models and qualitative online HD map construction results on KITScenes Multimodal.}
    \label{fig:history_example_online_hd_map_construction}
\end{figure}

Online HD map construction aims to predict a structured, drivable map directly from onboard sensor data, without relying on pre-built prior maps.
Existing benchmarks evaluate the prediction of simple geometric primitives such as lane dividers and pedestrian crossings~\cite{maptrv2}, leading to saturation of existing benchmarks, as shown in \Cref{fig:history_online_hd_map_construction}. 
We enable a substantially more complete formulation: our Lanelet2 maps encode lane topology, regulatory elements, traffic signs, and traffic lights with their lane assignments, allowing models to be evaluated on the complete Lanelet2 map structure on a substantially bigger and diverse map area than established benchmarks.
As a baseline for topology prediction, we extend MapQR~\cite{liu2024mapqr} and StreamMapNet~\cite{yuan2024streammapnet} with a graph neural network (GNN) head that 
predicts pairwise relations between all predicted map elements (called MapQR-Topo and StreamMapNet-Topo).
Implementation details are described in \Cref{sec:appx:benchmark_details:online_hd_map_construction}.

\paragraph{Results.}
In~\Cref{tab:detection_results}, we evaluate MapTRv2~\cite{maptrv2} as a widely adopted camera-only baseline and its improved derivative MapQR~\cite{liu2024mapqr}, StreamMapNet~\cite{yuan2024streammapnet} as a temporal baseline, and SDTagNet~\cite{immel2026sdtagnet} as a representative of methods that leverage our SD map priors.
Due to the significant gap between existing online HD map construction datasets and KITScenes Multimodal, we train the networks from scratch (see \Cref{sec:appx:benchmark_details:online_hd_map_construction} for details).

All approaches exhibit a large performance drop compared to existing benchmarks, revealing a gap hidden by the currently limited task scope, with SDTagNet benefiting most from the richer formulation.
This suggests that structured prior knowledge becomes increasingly valuable as the task approaches real-world complexity.
Example outputs are provided in \Cref{fig:example_online_hd_map_construction}.
A qualitative example of the predicted topology by MapQR-Topo is shown in \Cref{fig:mapqr_pred}.

\begin{table}[H]
  \centering
  \caption{Evaluation of online HD map construction models. For readability, the classes are grouped into 6 categories for the average precision: Lane Markings (LM), Lane Centerlines (LC), Road Infrastructure (RI), Traffic Lights (TL), Traffic Signs (TS) and Road Markings (RM); each category AP is the mean over its classes with ground truth in the evaluated validation frames. For the topology prediction baselines we additionally report the topology score. Evaluation details in \Cref{sec:appx:benchmark_details:online_hd_map_construction}.}
  \label{tab:detection_results}
  \resizebox{\textwidth}{!}{%
  \begin{tabular}{@{}lcccccc|c|c@{}}
    \toprule
    \textbf{Model} & $\text{AP}_{LM}$ & $\text{AP}_{LC}$ & $\text{AP}_{RI}$ & $\text{AP}_{TL}$ & $\text{AP}_{TS}$ & $\text{AP}_{RM}$ & mAP & $\text{AP}_{Topo}$ \\
     No. of classes in AP bucket & \footnotesize(9) & \footnotesize(2) & \footnotesize(5) & \footnotesize(3) & \footnotesize(18) & \footnotesize(9) & \footnotesize(46) & \\
    \midrule
    MapTRv2~\cite{maptrv2}             & 7.6 & 28.2 & 17.8  & 8.1  & 1.8 & 11.6  & 8.2 & - \\
    SDTagNet~\cite{immel2026sdtagnet}       & 9.9 & 26.5 & 15.5 & 11.4 & 2.3 & 14.5 & 9.3 & -\\
    StreamMapNet-Topo~\cite{yuan2024streammapnet}   & 8.6 & 27.0 & 15.9 & 12.2 & 1.9 & 7.3  & 7.7 & 26.1 \\
    MapQR-Topo~\cite{liu2024mapqr}                         & 8.1 & 26.8 & 17.7  & 6.6  & 1.5 & 7.8  & 7.2 & 28.2 \\
    \bottomrule
  \end{tabular}%
  }
\end{table}

\subsection{Long-range Depth Estimation}
\label{sec:benchmarks:monocular_depth}

\begin{figure}[h]
    \centering
    \begin{minipage}[t]{0.49\textwidth}
    \centering
    \caption{Lidar distribution across major autonomous driving datasets. Lines show the per-bin mean over 500 train samples. KITScenes Multimodal sets a new benchmark in both density and range.}
    \includegraphics[width=\columnwidth]{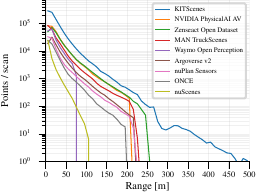}
    \label{fig:lidar_point_distribution}
    \end{minipage}
    \hfill
    \begin{minipage}[t]{0.49\textwidth}
    \centering
    \caption{Depth distributions of pixels with valid lidar depth. Depending on the method and compared to the ground truth, a systematic shortfall compared to the ground truth is observable starting at \qtyrange[range-phrase=--]{75}{125}{\meter}}
    \includegraphics[width=\columnwidth]{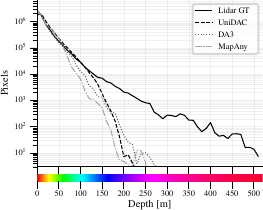}
    \label{fig:depth_distribution}
    \end{minipage}

    \includegraphics[width=\textwidth, trim=0 130 0 200, clip]{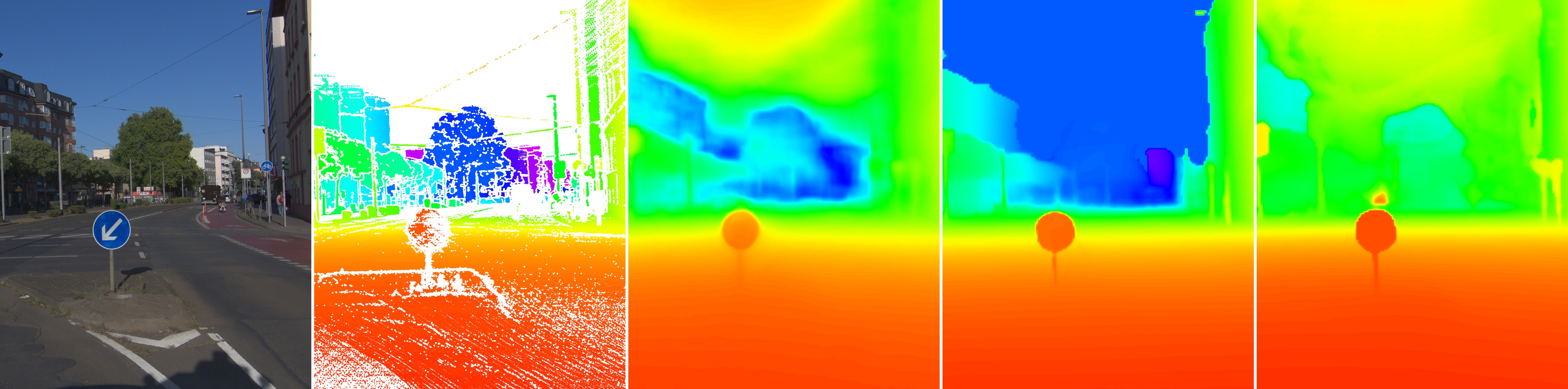}
    \newcolumntype{C}[1]{>{\centering\arraybackslash}p{#1}}
    {\small
    \begin{tabular}{C{0.17\textwidth}C{0.17\textwidth}C{0.17\textwidth}C{0.17\textwidth}C{0.17\textwidth}}
        RGB & KITScenes Lidar & UniDAC & Depth Anything 3 & MapAnything
    \end{tabular}
    }
    \vspace{-8pt}
    \caption{Qualitative comparison of monocular depth estimation methods. The depth scale is introduced in \Cref{fig:depth_distribution}. All methods systematically underestimate depth at long range, with MapAnything exhibiting the largest deviation.}
    \label{fig:qualitative_depth}
\end{figure}

Depth estimation has made rapid progress on near-range benchmarks, yet autonomous driving at highway speeds and in complex intersections requires reliable depth estimates well beyond \SI{100}{\meter}.
We show that current depth estimation models trained and evaluated on existing datasets fail to generalize to long-range distances, as their training signal is dominated by close-range lidar returns.
As shown in \Cref{fig:lidar_point_distribution}, we provide a dedicated benchmark for long-range depth estimation, enabling the first systematic evaluation of depth estimation at ranges that extend beyond \SI{400}{\meter}.

We report established metrics for monocular and stereo  depth evaluation: mean absolute relative error (AbsRel), threshold accuracy ($\delta_1$) and end-point error (EPE).
Scores are reported stratified into close range (\qtyrange[range-phrase=--]{0}{100}{\meter}), medium range (\qtyrange[range-phrase=--]{100}{200}{\meter}) and far range (\qty{>200}{\meter}), and overall. 
Evaluation setup, ground truth generation and further evaluations are provided in \Cref{sec:appx:benchmark_details:depth}.

\paragraph{Monocular Depth Results.}
We evaluate UniDAC~\cite{ganesan2026unidacuniversalmetricdepth}, Depth~Anything~3~\cite{depthanything3}, and MapAnything~\cite{keetha2026mapanything}, all reported to achieve dataset-agnostic SOTA monocular depth estimation, in a zero-shot protocol.
They provide strong performance at close range, but fall short as early as \qty{75}{\meter} (see~\Cref{fig:depth_distribution}).
\Cref{tab:depth-qualitative} reveals a critical limitation of aggregate evaluation:
overall metrics mask severe performance inversions across depth ranges.
MapAnything dominates the \qtyrange[range-phrase=--]{0}{100}{\meter} range and ranks first overall, yet degrades
significantly beyond it. UniDAC, ranked last overall, is in fact the strongest
long-range estimator by a significant margin. Regardless, no method achieves reliable performance
beyond \qty{200}{\meter}.
With its comprehensive lidar setup, KITScenes exposes such limitations, providing the long-range ground truth to benchmark methods where current autonomous driving datasets fall short.
A qualitative comparison is shown in \Cref{fig:qualitative_depth}.

\begin{table}[h]
  \caption{Range-stratified metric depth estimation exposes a ranking inversion: MapAnything dominates overall and at \qtyrange[range-phrase=--]{0}{100}{\meter} but degrades severely beyond it, while UniDAC, ranked last overall, is the strongest long-range estimator. Regardless, all methods perform poorly beyond \SI{200}{\meter}.}
  \label{tab:depth-qualitative}
  \centering
  \resizebox{\textwidth}{!}{%
  \begin{tabular}{@{}l c c c c c c c c@{}}
    \toprule
    Method & \multicolumn{2}{c}{\qtyrange[range-phrase=--]{0}{100}{\meter}} & \multicolumn{2}{c}{\qtyrange[range-phrase=--]{100}{200}{\meter}} & \multicolumn{2}{c}{\qty{>200}{\meter}} & \multicolumn{2}{c}{Overall} \\
    \cmidrule(lr){2-3}\cmidrule(lr){4-5}\cmidrule(lr){6-7}\cmidrule(lr){8-9}
    & AbsRel $\downarrow$ & $\delta_{1}$ $\uparrow$ & AbsRel $\downarrow$ & $\delta_{1}$ $\uparrow$ & AbsRel $\downarrow$ & $\delta_{1}$ $\uparrow$ & AbsRel $\downarrow$ & $\delta_{1}$ $\uparrow$ \\
    \midrule
  UniDAC~\cite{ganesan2026unidacuniversalmetricdepth} & 0.393 & 23.42 & \textbf{0.327} & \textbf{35.49} & \textbf{0.567} & \textbf{1.22} & 0.392 & 23.57 \\
  Depth Anything 3~\cite{depthanything3} & 0.273 & 51.13 & 0.481 & 11.44 & 0.708 & 0.56 & 0.277 & 50.41 \\
  MapAnything~\cite{keetha2026mapanything} & \textbf{0.153} & \textbf{81.06} & 0.516 & 10.73 & 0.778 & 0.00 & \textbf{0.160} & \textbf{79.79} \\
    \bottomrule
  \end{tabular}%
  }
\end{table}

\paragraph{Stereo Depth Results.}
Similarly, we evaluate Semi-Global Block Matching (SGBM)~\cite{hirschmuller2008stereo}, DEFOM-Stereo~\cite{jiang2025defom}, and FoundationStereo~\cite{wen2025foundationstereo} for zero-shot depth estimation.
Results in \Cref{tab:stereo-depth-qualitative} show that state-of-the-art stereo approaches outperform their monocular counterparts, with FoundationStereo achieving the best scores at every range. As with monocular methods, accuracy degrades with range, albeit less severely.

\begin{table}[h]
  \caption{Range-stratified stereo depth estimation: FoundationStereo outperforms DEFOM-Stereo and the SGBM baseline at every range. Depth accuracy degrades with range, though less severely than for the monocular methods in \Cref{tab:depth-qualitative}.}
  \label{tab:stereo-depth-qualitative}
  \centering
  \resizebox{\textwidth}{!}{%
  \begin{tabular}{@{}l c c c c c c c c c c c c@{}}
    \toprule
    Method & \multicolumn{3}{c}{\qtyrange[range-phrase=--]{0}{100}{\meter}} & \multicolumn{3}{c}{\qtyrange[range-phrase=--]{100}{200}{\meter}} & \multicolumn{3}{c}{\qty{>200}{\meter}} & \multicolumn{3}{c}{Overall} \\
    \cmidrule(lr){2-4}\cmidrule(lr){5-7}\cmidrule(lr){8-10}\cmidrule(lr){11-13}
    & EPE $\downarrow$ & AbsRel $\downarrow$ & $\delta_{1}$ $\uparrow$ & EPE $\downarrow$ & AbsRel $\downarrow$ & $\delta_{1}$ $\uparrow$ & EPE $\downarrow$ & AbsRel $\downarrow$ & $\delta_{1}$ $\uparrow$ & EPE $\downarrow$ & AbsRel $\downarrow$ & $\delta_{1}$ $\uparrow$ \\
    \midrule
  SGBM (WLS)~\cite{hirschmuller2008stereo} & 27.582 & 0.243 & 79.24 & 18.008 & 0.266 & 65.24 & 12.784 & 0.276 & 58.42 & 27.416 & 0.243 & 79.00 \\
  DEFOM-Stereo~\cite{jiang2025defom} & 4.790 & 0.155 & 93.11 & 2.805 & 0.192 & 76.77 & 1.875 & 0.206 & 70.57 & 4.756 & 0.156 & 92.83 \\
  FoundationStereo~\cite{wen2025foundationstereo} & \textbf{2.199} & \textbf{0.131} & \textbf{96.02} & \textbf{1.555} & \textbf{0.166} & \textbf{83.14} & \textbf{1.182} & \textbf{0.179} & \textbf{75.76} & \textbf{2.188} & \textbf{0.132} & \textbf{95.80} \\
    \bottomrule
  \end{tabular}%
  }
\end{table}

\subsection{Novel View Synthesis}
\label{sec:benchmarks:nvs}

Neural scene representations and novel view synthesis (NVS) methods have emerged as powerful tools for autonomous driving simulation and data augmentation.
Common NVS methods~\cite{yan2024street,chen2025omnire,yu2026_recondrive} are evaluated using pixel-based metrics, but this strongly relies on the availability of ground truth images at target viewpoints, which are typically restricted to the originally driven trajectory.
While lateral novel view synthesis is critical for autonomous driving simulation, its quality is often judged only through qualitative inspection~\cite{yu2026_recondrive} and image-based metrics~\cite{unisim, ni2025recondreamer}. 
However, those often fail to reveal subtle structural distortions that can significantly impact downstream perception tasks.
To probe geometric fidelity at novel lateral poses, we introduce a map-based NVS evaluation benchmark using traffic sign recall.

We synthesize the front ring camera at six lateral offsets $\Delta y\in\{\pm1,\pm2,\pm3\}$\,m and project ground-truth traffic signs from our HD map into each shifted viewpoint, applying lidar-based occlusion filtering to retain only unoccluded signs (see \Cref{fig:nvs_visualization}).
We score signs whose box lies fully inside the $280{\times}498$ evaluation image and whose shortest side is at least \SI{8}{\pixel}.
Running the same detector on the real photograph at $\Delta y=0$ provides a detector reference.
We evaluate the released checkpoints of ReconDrive~\cite{yu2026_recondrive} and AnyView~\cite{anyview} in a zero-shot protocol.
A full description is given in~\Cref{sec:appx:benchmark_details:nvs}.

\paragraph{Results.}

As shown in~\Cref{tab:lateral_recall}, ReconDrive's VGGT-based geometric backbone provides much better 3D understanding, enabling better spatial reconstruction and novel view synthesis compared to the video diffusion model AnyView.
More details, further qualitative comparison in \Cref{fig:qualitative_lateral_nvs} and standard photometric metrics are provided in~\Cref{sec:appx:benchmark_details:nvs}.

\begin{table}[h]
  \centering
  \caption{%
    Traffic sign recall on the front camera at six lateral offsets.
    Orig. is the detector's recall on the real unshifted image, acting as reference, not as ground truth for shifted views.
    Parentheses give the change relative to the original.
    $\uparrow$ denotes higher is better.
  }
  \label{tab:lateral_recall}
  \setlength{\tabcolsep}{4pt}
  \resizebox{\linewidth}{!}{
  \begin{tabular}{l c cccccc}
    \toprule
      Recall in \% $\uparrow$
      & $-3\,\text{m}$ & $-2\,\text{m}$ & $-1\,\text{m}$
    & Orig. 
      & $+1\,\text{m}$ & $+2\,\text{m}$ & $+3\,\text{m}$ \\
    \midrule
    ReconDrive~\cite{yu2026_recondrive}
      & 18.8 \scriptsize{(-73.7\%)} & 31.7 \scriptsize{(-55.5\%)} & 47.3 \scriptsize{(-33.6\%)}
       & 71.3
      & 43.0 \scriptsize{(-39.7\%)} & 26.6 \scriptsize{(-62.7\%)} & 15.8 \scriptsize{(-77.8\%)} \\
    AnyView~\cite{anyview}
      & 0.6 \scriptsize{(-99.1\%)} & 0.8 \scriptsize{(-98.9\%)} & 1.8 \scriptsize{(-97.4\%)}
       & 71.3
      & 2.2 \scriptsize{(-96.9\%)} & 0.5 \scriptsize{(-99.3\%)} & 0.4 \scriptsize{(-99.5\%)} \\
    \bottomrule
  \end{tabular}
  }
\end{table}

\subsection{End-to-End Driving}
\label{sec:benchmarks:e2e}

End-to-end (e2e) driving and neural world models are evaluated primarily on nuScenes~\cite{caesar2020nuscenes}, nuPlan~\cite{caesar2021nuplan}, and their derivatives~\cite{openscene2023,dauner2024navsim}, narrowing the sensor configurations, geographies, and map-grounded behaviours under which they are assessed.
KITScenes Multimodal supports various input tiers on identical scenes, i.e., a single front-view camera, the full \SI{360}{\degree} surround-view, and the complete multi-modal suite with lidar and radar, enabling controlled modality ablations with a novel combination of benchmark metrics.
The evaluated baseline models range from single to multi-view camera input, partly fused with lidar BEV features, yet none exploits the full suite including radar or SD map priors, leaving open challenges for multi-modal e2e training.
Evaluation setup, split details, and the held-out \texttt{test-e2e} leaderboard split are described in \Cref{sec:appx:benchmark_details:e2e}.

Beyond standard ADE and FDE~\cite{alahi2016social}, we leverage our centimeter-accurate Lanelet2 maps and a lidar-derived occupancy layer to evaluate three safety metrics: \emph{map compliance}, \emph{collision-free rate}, and \emph{centerline distance}, serving as a proxy for safety properties usually assessed only in closed-loop simulation.
To decouple correctness from a single expert trajectory, we additionally adopt the \emph{Multi-Maneuver Score} (MMS)~\cite{wagner2026longtail}, scoring each prediction against the best of at least three reference maneuvers per scene.
Metric definitions and per-horizon profiles are detailed in \Cref{sec:appx:benchmark_details:e2e}.

\paragraph{Results.}

\begin{figure}[t]
    \centering
    \begin{minipage}[t]{0.45\textwidth}
    \centering
    \includegraphics[width=\columnwidth]{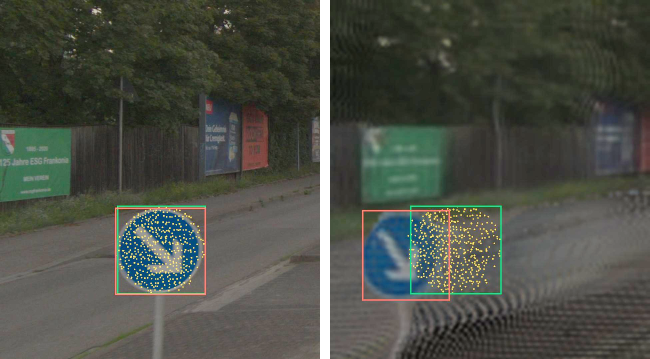}
    \caption{
    Example of geometric inconsistency in current NVS methods. The traffic sign aligns with the green ground-truth box in the source view (left), but is displaced from it after a \qty{1}{\meter} lateral viewpoint shift (right). Yellow dots show projected lidar points.}
    \label{fig:nvs_visualization}
    \end{minipage}
    \hfill
    \begin{minipage}[t]{0.52\textwidth}
    \centering
    \includegraphics[width=0.738\linewidth]{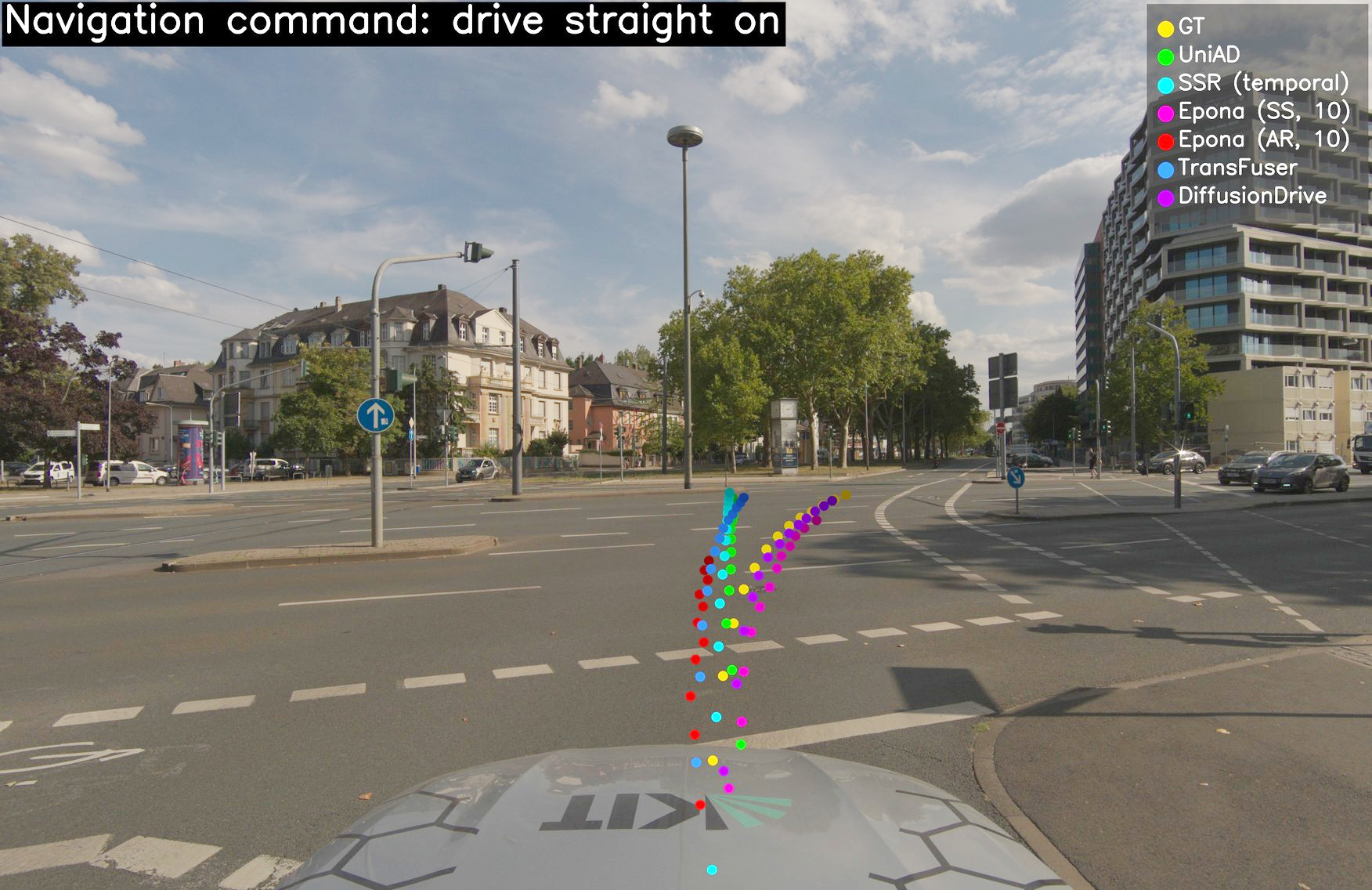}%
    \hfill
    \includegraphics[width=0.24\linewidth]{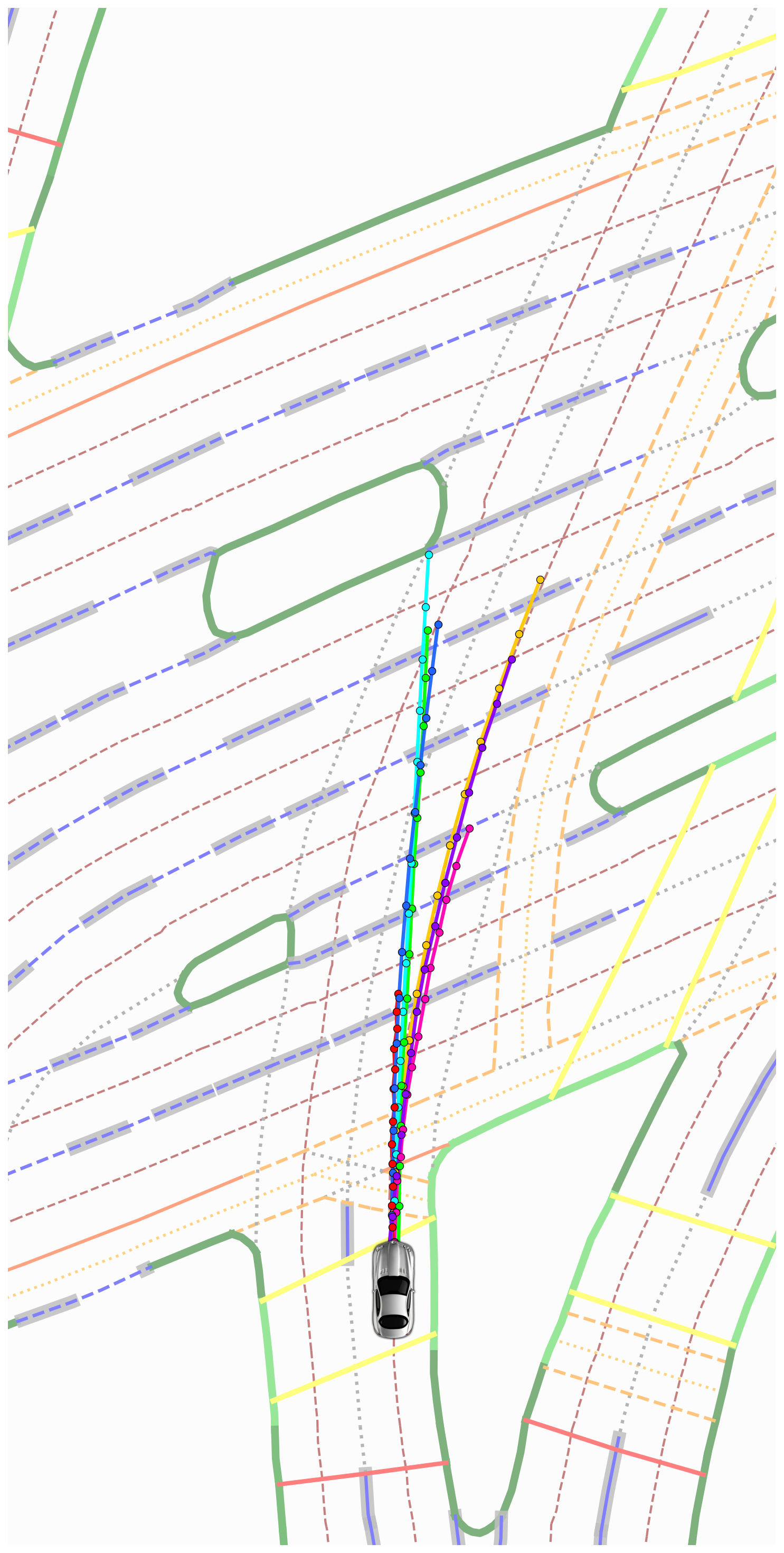}%
    \caption{Qualitative end-to-end prediction, showing the front-view camera image with a top-view of all model trajectories overlaid on the HD map and ground truth. Epona tracks the road curvature better than the over-committed navigation-conditioned models, see \Cref{fig:app:en2end_quali} for more examples.}
    \label{fig:en2end_quali}
    \end{minipage}
    \label{fig:nvs_e2e}
\end{figure}

\begin{table*}[t]
\centering
\caption{End-to-end results on the 483 e2e samples of the evaluation set with all metrics evaluated at the \SI{3}{\second} horizon. ADE and FDE follow~\cite{alahi2016social}; the map-grounded metrics map compliance, collision-free rate, and centerline distance leverage our HD maps together with a lidar-based occupancy layer.
ADE is additionally broken out by scene category (number of samples in brackets).
}
\label{tab:kitscenes_e2e_results}
\scriptsize
\renewcommand{\arraystretch}{1.08}
\setlength{\aboverulesep}{0pt}
\setlength{\belowrulesep}{0pt}
\setlength{\extrarowheight}{2pt}
\resizebox{\textwidth}{!}{%
\begin{tabular}{l|c|ccccccc|ccc}
\toprule
\multirow{2}{*}{Model}
&
FDE $\downarrow$
&
\multicolumn{7}{c|}{ADE $\downarrow$}
&
\multicolumn{3}{c}{Compliance / Tracking}
\\
\cmidrule{2-2}
\cmidrule{3-9}
\cmidrule{10-12}
& Avg.
& Avg. & \makecell{Selec.\\(3)} & \makecell{Constr.\\(3)} & \makecell{Overt.\\(47)} & \makecell{Inters.\\(153)} & \makecell{Night\\(63)} & \makecell{Nom.\\(212)}
& \makecell{Map.\\comp.}$\uparrow$ & \makecell{Coll.-\\free}$\uparrow$ & \makecell{CL\\dist.}$\downarrow$
\\
\midrule
UniAD~\cite{hu2023uniad}
& 4.63
& 2.32 & 2.56 & 1.38 & 2.13 & 1.85 & 3.37 & 2.42
& 61.1 & 89.2 & 0.62
\\
DMAD~\cite{shen2025dmad}
& 4.47
& 2.28 & 2.56 & 1.45 & 2.17 & 1.88 & 3.33 & 2.31
& 59.4 & 87.6 & 0.63
\\
SSR~\cite{li2025ssr} (non-temp.)
& 7.47
& 3.80 & 1.36 & 1.80 & 4.51 & 2.74 & 5.57 & 3.97
& 68.9 & 86.3 & 0.66
\\
SSR~\cite{li2025ssr} (temporal)
& 4.35
& 2.27 & 4.04 & 4.50 & 2.07 & 2.31 & 2.94 & 2.00
& 68.7 & 87.8 & 0.60
\\
\midrule
Epona~\cite{zhang2025epona} (AR, 10)
& 7.19
& 3.31 & 1.84 & 4.09 & 3.55 & 2.96 & 4.41 & 3.21
& \underline{82.0} & 91.5 & \textbf{0.39}
\\
Epona~\cite{zhang2025epona} (AR, 100)
& 5.62
& 2.60 & 2.07 & 3.30 & 2.99 & 2.24 & 3.55 & 2.50
& 73.7 & 93.0 & 0.43
\\
Epona~\cite{zhang2025epona} (SS, 10)
& 3.78
& 1.83 & 1.66 & 3.08 & 2.19 & 1.61 & 2.70 & 1.66
& 81.4 & 95.7 & \underline{0.40}
\\
Epona~\cite{zhang2025epona} (SS, 100)
& 3.82
& 1.83 & 1.75 & 2.77 & 2.20 & 1.61 & 2.68 & 1.65
& 78.5 & 95.7 & 0.41
\\
\midrule
TransFuser~\cite{chitta2023transfuser}
& \underline{2.24}
& 0.95 & \textbf{0.61} & \underline{0.92} & 1.13 & \underline{0.94} & \underline{1.03} & 0.90
& 78.5 & \underline{96.5} & 0.52
\\
LTF~\cite{chitta2023transfuser}
& \underline{2.24}
& \underline{0.94} & 0.69 & 1.07 & \underline{1.02} & 0.95 & 1.09 & \underline{0.88}
& \underline{82.0} & 95.7 & 0.53
\\
DiffusionDrive~\cite{liao2025diffusiondrive}
& \textbf{2.11}
& \textbf{0.87} & \underline{0.68} & \textbf{0.81} & \textbf{0.96} & \textbf{0.89} & \textbf{0.96} & \textbf{0.82}
& \textbf{83.9} & \textbf{97.5} & 0.48
\\
\midrule
\textit{Constant velocity (ref.)}
& \textit{3.50}
& \textit{1.39} & \textit{1.32} & \textit{3.61} & \textit{1.41} & \textit{1.88} & \textit{1.06} & \textit{1.10}
& \textit{75.2} & \textit{87.8} & \textit{0.41}
\\
\textit{Ground truth (ref.)}
& --- & --- & --- & --- & --- & --- & --- & ---
& \textit{91.5} & \textit{100.0} & \textit{0.30}
\\
\bottomrule
\end{tabular}%
}
\vspace{-3pt}
{\parbox{1.0\linewidth}{
    \vspace{1pt}
    \scriptsize
    Training data: nuScenes (UniAD, DMAD, SSR), nuPlan (Epona), NAVSIM (TransFuser, LTF, DiffusionDrive); inputs in \Cref{tab:app:e2e_inputs}. For SSR, \emph{non-temp.} uses only the current keyframe whereas \emph{temporal} also aggregates two past frames (\SI{1}{\second}). Epona uses single-step (SS) or autoregressive (AR) rollouts with 10 or 100 diffusion denoising steps. TransFuser and DiffusionDrive fuse three front cameras with lidar, LTF is the camera-only variant; all three receive the current ego velocity and acceleration as explicit planner inputs. \emph{Constant velocity} extrapolates the ego velocity at the keyframe; \emph{ground truth} evaluates the recorded trajectory and bounds the map-grounded metrics from above.
}}
\end{table*}

We zero-shot evaluate seven open-source baselines, ordered by training data: UniAD~\cite{hu2023uniad} and DMAD~\cite{shen2025dmad}, multi-task perception, prediction, and planning models trained with navigation commands on nuScenes; SSR~\cite{li2025ssr}, which plans directly with a self-supervised BEV regulariser; Epona~\cite{zhang2025epona}, an autoregressive front-view diffusion world model trained on nuPlan~\cite{caesar2021nuplan} without navigation commands; and three planners trained on NAVSIM~\cite{dauner2024navsim}, a nuPlan-derived benchmark:
TransFuser~\cite{chitta2023transfuser}, which fuses three front cameras with a lidar BEV, its camera-only variant Latent TransFuser (LTF)~\cite{chitta2023transfuser}, and DiffusionDrive~\cite{liao2025diffusiondrive}, which replaces the TransFuser planning head with a truncated diffusion decoder. 

\Cref{tab:kitscenes_e2e_results} reveals a substantial domain gap for the nuScenes-trained models:
Epona transfers better, consistent with its larger nuPlan pretraining corpus.
The NAVSIM-trained planners achieve by far the lowest displacement errors, and DiffusionDrive also leads in map compliance and collision-free rate.
Although the training corpora differ between the model families, we attribute a large part of this lead to the inputs rather than to better scene understanding (see  \Cref{tab:app:e2e_inputs}): the NAVSIM planners receive the current ego velocity and acceleration as explicit planning inputs, whereas Epona conditions on past ego poses.
UniAD, DMAD, and SSR also receive the ego vehicle state, but only indirectly within the architecture, \eg as an embedding added to the BEV queries, from which the planner has to recover the ego motion.

A constant-velocity extrapolation of the current ego velocity already outperforms every model without these explicit inputs in ADE.
Similar effects have been reported for open-loop evaluation on nuScenes by Zhai~\etal~\cite{zhai2023admlp} and Li~\etal~\cite{li2024egostatus}, indicating that short-horizon displacement errors are dominated by the ego state rather than by scene understanding.
Epona illustrates the difference: although it observes past poses, it is a monocular, command-free world model whose trajectory is optimised jointly with future-frame generation rather than for ego-motion regression. Its single-step variants beat the constant-velocity reference at intersections but trail it in nominal and night scenes, where extrapolation suffices.
Beyond the \SI{3}{\second} horizon used here, KITScenes Multimodal provides \SI{5}{\second} ground-truth trajectories and driving commands for both a 3 and \SI{5}{\second} window, so future models trained for longer horizons can be also evaluated at \SI{5}{\second}.
For the current models, which predict at most \SI{4}{\second}, we report extrapolated \SI{5}{\second} results in \Cref{tab:app:kitscenes_5s_results}.
\Cref{fig:en2end_quali} illustrates a qualitative example of end-to-end predictions.

\section{Limitations}
\label{sec:limitations}

\paragraph{Dynamic-Object Annotations.}
The current release does not include 3D bounding boxes, tracks, or 3D instance segmentations for dynamic agents.
These annotations will be added in a future release.

\paragraph{Dataset Scale.}
At \SI{11.8}{\hour} of current recorded data, KITScenes Multimodal is smaller in raw volume than recent large-scale sensor corpora such as nuPlan Sensor (${\approx}120$\,h) or Nvidia Physical AI AV~(${\approx}1700$\,h). 
However, these datasets target fundamentally different tasks and provide neither the same annotation types nor comparable sensor fidelity.
Progress in spatial machine learning is increasingly driven by two complementary regimes: large-scale pre-training, where data volume is central, and curated evaluation or fine-tuning data with benchmark protocols that reflect target deployment behavior.
Our dataset primarily supports the latter, offering sensor fidelity, annotation completeness, and benchmark breadth that are difficult to replicate at auto-labeled corpus scale.

\paragraph{Open-Loop End-to-End Evaluation.}
While the maps are validated end-to-end through closed-loop driving trials in Autoware~\cite{autoware} as shown in \Cref{sec:appx:autoware_trials}, our end-to-end benchmark evaluates open-loop trajectory prediction only. While the released artifacts enable closed-loop evaluation in the Autoware simulator, we leave such experiments to future work.

\section{Conclusion}

We presented KITScenes Multimodal, a European multi-modal driving dataset that pairs a state-of-the-art sensor suite with high-resolution synchronized global-shutter cameras, lidar reaching beyond \SI{400}{\meter}, and 4D imaging radar with the most complete public HD maps of any dataset, covering \SI{60}{\kilo\meter\squared} of area and validated by closed-loop autonomous-driving trials.
Across our four benchmarks, online HD map construction, long-range depth estimation, novel view synthesis, and end-to-end driving, current state-of-the-art methods leave systematic capability gaps that prior datasets cannot surface, from complete map prediction at full Lanelet2 fidelity, through long-range depth and geometrically consistent novel views, to map-grounded trajectory evaluation in cluttered European urban scenes.
By coupling deployment-grade maps with long-range, high-fidelity sensing, KITScenes Multimodal offers a controlled testbed for the spatial reasoning capabilities required on the path to L4 autonomy.

\section*{Acknowledgements}
The HD maps required more than 10,000 hours of annotation and manual review.
We sincerely thank the following student research assistants who contributed to this effort: Yuxuan Sun, Karan Lashkare, Tom Wetzel, Xuyi Shi, Jiarui Zhang, Vincent Clemens Scherer, Lasse Nils Niehues, Jana-Sophie Schwarzkopf, Mingxuan Dong, Chen Yiyang, Kaiqi Qu, Yong Peng, Yaraslau Krautsou, Victoria von Borstel, Juntao Wei, Ege Güner, Uğur Sesli, Özge Özmısır, Tom Eller, Junjie Luo, Xin Xu, Xiaoyu Yan, Qingquan Wei, Jiayi You, Qicai Tan, Omkar Jarande, Jonathan Glück, Chengshuang Yang, Özgür Karadag, Ningyuan Gao, and Chunlin Yin.

We also thank the student developers who supported us in building the tooling behind the scenes: Gleb Stepanov, Luca Fleckenstein, Cheng Zhu, Wenhao Guo, and Bence Balint.

The research leading to these results is partially funded by the German Federal Ministry for Economic Affairs and Energy within the project NXT GEN AI METHODS, and by the Shift2SDV project (Grant Agreement No. 101194245), co-funded by the European Union through the Chips Joint Undertaking and its members. The authors gratefully acknowledge the computing time provided on the high-performance computer HoreKa by the National High-Performance Computing Center at KIT (NHR@KIT). This center is jointly supported by the Federal Ministry of Education and Research and the Ministry of Science, Research and the Arts of Baden-Württemberg, as part of the National High-Performance Computing (NHR) joint funding program. HoreKa is partly funded by the German Research Foundation (DFG). The authors also gratefully acknowledge the support of the Large Scale Data Facility (LSDF) at the Karlsruhe Institute of Technology, funded by the Ministry of Science, Research and the Arts of Baden-Württemberg and by the Federal Ministry of Education and Research. Road network data for SD Maps were obtained from OpenStreetMap (© OpenStreetMap contributors; available under the ODbL, https://www.openstreetmap.org/copyright).

{

\small
\bibliographystyle{unsrtnat} 
\bibliography{kitscenes_perception}

@misc{szabadka2024jpegli,
  author       = {Szabadka, Zoltan and Bruse, Martin and Alakuijala, Jyrki},
  title        = {Introducing {Jpegli}: A New {JPEG} Coding Library},
  howpublished = {Google Open Source Blog},
  year         = {2024},
  month        = apr,
  day          = {3},
  url          = {https://opensource.googleblog.com/2024/04/introducing-jpegli-new-jpeg-coding-library.html},
  note         = {Accessed: 2026-05-01}
}

@article{immel2024lanelet2mlconverter,
  title         = {Generation of Training Data from HD Maps in the Lanelet2 Framework},
  author        = {Immel, Fabian and Fehler, Richard and Bieder, Frank and Stiller, Christoph},
  journal       = {arXiv preprint arXiv:2407.17409},
  year          = {2024},
  eprint        = {2407.17409},
  archiveprefix = {arXiv},
  primaryclass  = {cs.CV},
  url           = {https://arxiv.org/abs/2407.17409}
}

@inproceedings{immel2026sdtagnet,
 author = {Immel, Fabian and Pauls, Jan-Hendrik and Schwarzkopf, Richard and Bieder, Frank and Merkert, Jonas and Stiller, Christoph},
 booktitle = {Advances in Neural Information Processing Systems},
 doi = {10.52202/085713-1013},
 editor = {D. Belgrave and C. Zhang and H. Lin and R. Pascanu and P. Koniusz and M. Ghassemi and N. Chen},
 pages = {30175--30201},
 publisher = {Curran Associates, Inc.},
 title = {SDTagNet: Leveraging Text-Annotated Navigation Maps for Online HD Map Construction},
 url = {https://proceedings.neurips.cc/paper_files/paper/2025/file/2b3dd4ff044c3e1e0bbc485acd8a3229-Paper-Conference.pdf},
 volume = {38, Main Conference},
 year = {2025}
}

@inproceedings{wang2023openlanev2,
  title     = {Openlane-v2: A topology reasoning benchmark for unified 3d hd mapping},
  author    = {Wang, Huijie and Li, Tianyu and Li, Yang and Chen, Li and Sima, Chonghao and Liu, Zhenbo and Wang, Bangjun and Jia, Peijin and Wang, Yuting and Jiang, Shengyin and others},
  booktitle = {Thirty-seventh Conference on Neural Information Processing Systems Datasets and Benchmarks Track},
  year      = {2023}
}

@INPROCEEDINGS {unisim,
author = { Yang, Ze and Chen, Yun and Wang, Jingkang and Manivasagam, Sivabalan and Ma, Wei-Chiu and Yang, Anqi Joyce and Urtasun, Raquel },
booktitle = { 2023 IEEE/CVF Conference on Computer Vision and Pattern Recognition (CVPR) },
title = {{ UniSim: A Neural Closed-Loop Sensor Simulator }},
year = {2023},
volume = {},
ISSN = {},
pages = {1389-1399},
doi = {10.1109/CVPR52729.2023.00140},
url = {https://doi.ieeecomputersociety.org/10.1109/CVPR52729.2023.00140},
publisher = {IEEE Computer Society},
address = {Los Alamitos, CA, USA},
month =Jun}

@inproceedings{li2022hdmapnet,
  author    = {Li, Qi and Wang, Yue and Wang, Yilun and Zhao, Hang},
  booktitle = {2022 International Conference on Robotics and Automation (ICRA)},
  title     = {HDMapNet: An Online HD Map Construction and Evaluation Framework},
  year      = {2022},
  volume    = {},
  number    = {},
  pages     = {4628-4634},
  doi       = {10.1109/ICRA46639.2022.9812383}
}

@inproceedings{qiao2023bemapnet,
  author    = {Qiao, Limeng and Ding, Wenjie and Qiu, Xi and Zhang, Chi},
  title     = {End-to-End Vectorized HD-Map Construction With Piecewise Bezier Curve},
  booktitle = {Proceedings of the IEEE/CVF Conference on Computer Vision and Pattern Recognition (CVPR)},
  month     = {June},
  year      = {2023},
  pages     = {13218-13228}
}

@inproceedings{ding2023pivotnet,
  title     = {Pivotnet: Vectorized pivot learning for end-to-end hd map construction},
  author    = {Ding, Wenjie and Qiao, Limeng and Qiu, Xi and Zhang, Chi},
  booktitle = {Proceedings of the IEEE/CVF International Conference on Computer Vision},
  pages     = {3672--3682},
  year      = {2023}
}

@INPROCEEDINGS{caesar2021nuplan, 
  title={NuPlan: A closed-loop ML-based planning benchmark for autonomous vehicles},
  author={H. Caesar, J. Kabzan, K. Tan et al.},
  booktitle={CVPR ADP3 workshop},
  year=2021
}

@inproceedings{av2_trust_but_verify,
  author    = {Lambert, John and Hays, James},
  booktitle = {Proceedings of the Neural Information Processing Systems Track on Datasets and Benchmarks},
  editor    = {J. Vanschoren and S. Yeung},
  pages     = {},
  publisher = {Curran},
  title     = {Trust, but Verify: Cross-Modality Fusion for HD Map Change Detection},
  volume    = {1},
  year      = {2021}
}

@misc{ghilotti2026truckdrive,
      title={TruckDrive: Long-Range Autonomous Highway Driving Dataset}, 
      author={Filippo Ghilotti and Edoardo Palladin and Samuel Brucker and Adam Sigal and Mario Bijelic and Felix Heide},
      year={2026},
      eprint={2603.02413},
      archivePrefix={arXiv},
      primaryClass={cs.CV},
      url={https://arxiv.org/abs/2603.02413}, 
}

@inproceedings{liu2023vectormapnet,
  title     = {{V}ector{M}ap{N}et: End-to-end Vectorized {HD} Map Learning},
  author    = {Liu, Yicheng and Yuan, Tianyuan and Wang, Yue and Wang, Yilun and Zhao, Hang},
  booktitle = {Proceedings of the 40th International Conference on Machine Learning},
  pages     = {22352--22369},
  year      = {2023},
  editor    = {Krause, Andreas and Brunskill, Emma and Cho, Kyunghyun and Engelhardt, Barbara and Sabato, Sivan and Scarlett, Jonathan},
  volume    = {202},
  series    = {Proceedings of Machine Learning Research},
  month     = {23--29 Jul},
  publisher = {PMLR}
}

@inproceedings{mao2021once,
  author    = {Mao, Jiageng and Minzhe, Niu and Jiang, ChenHan and liang, hanxue and Chen, Jingheng and Liang, Xiaodan and Li, Yamin and Ye, Chaoqiang and Zhang, Wei and Li, Zhenguo and Yu, Jie and XU, Chunjing and Xu, Hang},
  booktitle = {Proceedings of the Neural Information Processing Systems Track on Datasets and Benchmarks},
  editor    = {J. Vanschoren and S. Yeung},
  pages     = {},
  title     = {One Million Scenes for Autonomous Driving: ONCE Dataset},
  url       = {https://datasets-benchmarks-proceedings.neurips.cc/paper_files/paper/2021/file/67c6a1e7ce56d3d6fa748ab6d9af3fd7-Paper-round1.pdf},
  volume    = {1},
  year      = {2021}
}

@article{huang2018apolloscape,
  author    = {Huang, Xinyu and Wang, Peng and Cheng, Xinjing and Zhou, Dingfu and Geng, Qichuan and Yang, Ruigang},
  journal   = { IEEE Transactions on Pattern Analysis \& Machine Intelligence },
  title     = {{ The ApolloScape Open Dataset for Autonomous Driving and Its Application }},
  year      = {2020},
  volume    = {42},
  number    = {10},
  issn      = {1939-3539},
  pages     = {2702-2719},
  doi       = {10.1109/TPAMI.2019.2926463},
  url       = {https://doi.ieeecomputersociety.org/10.1109/TPAMI.2019.2926463},
  publisher = {IEEE Computer Society},
  address   = {Los Alamitos, CA, USA},
  month     = oct
}

@article{Liao2022PAMI,
  title   = {{KITTI}-360: A Novel Dataset and Benchmarks for Urban Scene Understanding in 2D and 3D},
  author  = {Yiyi Liao and Jun Xie and Andreas Geiger},
  journal = {Pattern Analysis and Machine Intelligence (PAMI)},
  year    = {2022}
}

@article{maptrv2,
  author   = {Liao, Bencheng
              and Chen, Shaoyu
              and Zhang, Yunchi
              and Jiang, Bo
              and Zhang, Qian
              and Liu, Wenyu
              and Huang, Chang
              and Wang, Xinggang},
  title    = {MapTRv2: An End-to-End Framework for Online Vectorized HD Map Construction},
  journal  = {International Journal of Computer Vision},
  year     = {2024},
  month    = {Oct},
  day      = {06},
  issn     = {1573-1405},
  doi      = {10.1007/s11263-024-02235-z},
  url      = {https://doi.org/10.1007/s11263-024-02235-z}
}

@misc{bruse2024userspreferjpeglisamesized,
  title         = {Users prefer Jpegli over same-sized libjpeg-turbo or MozJPEG},
  author        = {Martin Bruse and Luca Versari and Zoltan Szabadka and Jyrki Alakuijala},
  year          = {2024},
  eprint        = {2403.18589},
  archiveprefix = {arXiv},
  primaryclass  = {eess.IV},
  url           = {https://arxiv.org/abs/2403.18589}
}

@inproceedings{ettinger2021waymo_motion,
  author    = {Ettinger, Scott and Cheng, Shuyang and Caine, Benjamin and Liu, Chenxi and Zhao, Hang and Pradhan, Sabeek and Chai, Yuning and Sapp, Ben and Qi, Charles R. and Zhou, Yin and Yang, Zoey and Chouard, Aur\'elien and Sun, Pei and Ngiam, Jiquan and Vasudevan, Vijay and McCauley, Alexander and Shlens, Jonathon and Anguelov, Dragomir},
  title     = {Large Scale Interactive Motion Forecasting for Autonomous Driving: The Waymo Open Motion Dataset},
  booktitle = {Proceedings of the IEEE/CVF International Conference on Computer Vision (ICCV)},
  month     = {October},
  year      = {2021},
  pages     = {9710-9719}
}

@inproceedings{sun2020waymo_perception,
  author    = {Sun, Pei and Kretzschmar, Henrik and Dotiwalla, Xerxes and Chouard, Aurelien and Patnaik, Vijaysai and Tsui, Paul and Guo, James and Zhou, Yin and Chai, Yuning and Caine, Benjamin and Vasudevan, Vijay and Han, Wei and Ngiam, Jiquan and Zhao, Hang and Timofeev, Aleksei and Ettinger, Scott and Krivokon, Maxim and Gao, Amy and Joshi, Aditya and Zhang, Yu and Shlens, Jonathon and Chen, Zhifeng and Anguelov, Dragomir},
  title     = {Scalability in Perception for Autonomous Driving: Waymo Open Dataset},
  booktitle = {Proceedings of the IEEE/CVF Conference on Computer Vision and Pattern Recognition (CVPR)},
  month     = {June},
  year      = {2020}
}

@article{geiger2012kitti,
  author    = {Andreas Geiger and Philip Lenz and Raquel Urtasun and Christoph Stiller},
  title     = {Vision meets robotics: The KITTI dataset},
  journal = {The International Journal of Robotics Research},
  volume = {32},
  number = {11},
  pages = {1231-1237},
  year = {2013},
  doi = {10.1177/0278364913491297}
}

@inproceedings{caesar2020nuscenes,
  author    = {Caesar, Holger and Bankiti, Varun and Lang, Alex H. and Vora, Sourabh and Liong, Venice Erin and Xu, Qiang and Krishnan, Anush and Pan, Yu and Baldan, Giancarlo and Beijbom, Oscar},
  title     = {nuScenes: A Multimodal Dataset for Autonomous Driving},
  booktitle = {Proceedings of the IEEE/CVF Conference on Computer Vision and Pattern Recognition (CVPR)},
  month     = {June},
  year      = {2020}
}

@inproceedings{alibeigi2023zenseact,
  title     = {Zenseact Open Dataset: A large-scale and diverse multimodal dataset for autonomous driving},
  author    = {Alibeigi, Mina and Ljungbergh, William and Tonderski, Adam and Hess, Georg and Lilja, Adam and Lindstrom, Carl and Motorniuk, Daria and Fu, Junsheng and Widahl, Jenny and Petersson, Christoffer},
  booktitle = {Proceedings of the IEEE/CVF International Conference on Computer Vision},
  year      = {2023}
}

@inproceedings{fent2024truckscenes,
  title     = {MAN TruckScenes: A multimodal dataset for autonomous trucking in diverse conditions},
  author    = {Fent, Felix and Kuttenreich, Fabian and Ruch, Florian and Rizwin, Farija and Juergens, Stefan and Lechermann, Lorenz and Nissler, Christian and Perl, Andrea and Voll, Ulrich and Yan, Min and Lienkamp, Markus},
  booktitle = {Advances in Neural Information Processing Systems},
  editor    = {A. Globerson and L. Mackey and D. Belgrave and A. Fan and U. Paquet and J. Tomczak and C. Zhang},
  pages     = {62062--62082},
  publisher = {Curran Associates, Inc.},
  url       = {https://proceedings.neurips.cc/paper_files/paper/2024/file/71ac06f0f8450e7d49063c7bfb3257c2-Paper-Datasets_and_Benchmarks_Track.pdf},
  volume    = {37},
  year      = {2024}
}

@misc{autoware,
  title        = {{Autoware}},
  author       = {{Autoware Foundation}},
  howpublished = {\url{https://github.com/autowarefoundation/autoware}},
  note         = {Accessed: 2026-05-02}
}

@inproceedings{yang2024emernerf,
title={EmerNe{RF}: Emergent Spatial-Temporal Scene Decomposition via Self-Supervision},
author={Jiawei Yang and Boris Ivanovic and Or Litany and Xinshuo Weng and Seung Wook Kim and Boyi Li and Tong Che and Danfei Xu and Sanja Fidler and Marco Pavone and Yue Wang},
booktitle={The Twelfth International Conference on Learning Representations},
year={2024},
url={https://openreview.net/forum?id=ycv2z8TYur}
}

@InProceedings{wu2023mars,
author="Wu, Zirui
and Liu, Tianyu
and Luo, Liyi
and Zhong, Zhide
and Chen, Jianteng
and Xiao, Hongmin
and Hou, Chao
and Lou, Haozhe
and Chen, Yuantao
and Yang, Runyi
and Huang, Yuxin
and Ye, Xiaoyu
and Yan, Zike
and Shi, Yongliang
and Liao, Yiyi
and Zhao, Hao",
editor="Fang, Lu
and Pei, Jian
and Zhai, Guangtao
and Wang, Ruiping",
title="MARS: An Instance-Aware, Modular and Realistic Simulator for Autonomous Driving",
booktitle="Artificial Intelligence",
year="2024",
publisher="Springer Nature Singapore",
address="Singapore",
pages="3--15",
isbn="978-981-99-8850-1"
}

@article{jiang2023vad,
  title={VAD: Vectorized Scene Representation for Efficient Autonomous Driving},
  author={Jiang, Bo and Chen, Shaoyu and Xu, Qing and Liao, Bencheng and Chen, Jiajie and Zhou, Helong and Zhang, Qian and Liu, Wenyu and Huang, Chang and Wang, Xinggang},
  journal={ICCV},
  year={2023}
}

@techreport{brighterai2022dnat,
  author      = {{brighter AI Technologies}},
  title       = {Face Off: Privacy v Progress --- How Deep Natural Anonymization Protects Privacy in the Age of Machine Learning},
  institution = {brighter AI Technologies GmbH},
  type        = {White Paper},
  year        = {2022},
  address     = {Berlin, Germany},
  url         = {https://ac-landing-pages-user-uploads-production.s3.amazonaws.com/0000122471/803bb7a7-de73-4596-9548-6d1ca3a80e32.pdf},
  urldate     = {2026-05-02}
}

@inproceedings{guizilini2020ddad,
  author = {Vitor Guizilini and Rares Ambrus and Sudeep Pillai and Allan Raventos and Adrien Gaidon},
  title = {3D Packing for Self-Supervised Monocular Depth Estimation},
  booktitle = {IEEE Conference on Computer Vision and Pattern Recognition (CVPR)},
  primaryClass = {cs.CV},
  year = {2020},
}

@inproceedings{Zhao20243DRef,
  author    = {Zhao, Xiting and Schwertfeger, Sören},
  booktitle = {2024 International Conference on 3D Vision (3DV)},
  title     = {3DRef: 3D Dataset and Benchmark for Reflection Detection in RGB and Lidar Data},
  year      = {2024},
  pages     = {225-234},
  doi       = {10.1109/3DV62453.2024.00009}
}

@software{rawtherapee,
  author  = {{RawTherapee Development Team}},
  title   = {{RawTherapee}: A Powerful Cross-Platform Raw Photo Processing Program},
  version = {5.12},
  date    = {2025-05-28},
  url     = {https://github.com/RawTherapee/RawTherapee},
  urldate = {2026-05-02},
  note    = {Includes the AMaZE demosaicing algorithm and raw-domain chromatic aberration correction by E.~J.~Martinec}
}

@inproceedings{poggenhans2018lanelet2,
  author    = {Poggenhans, Fabian and Pauls, Jan-Hendrik and Janosovits, Johannes and Orf, Stefan and Naumann, Maximilian and Kuhnt, Florian and Mayr, Matthias},
  booktitle = {2018 21st International Conference on Intelligent Transportation Systems (ITSC)},
  title     = {Lanelet2: A high-definition map framework for the future of automated driving},
  year      = {2018},
  pages     = {1672-1679},
  doi       = {10.1109/ITSC.2018.8569929}
}

@inproceedings{pauls2021automatic,
  author    = {Pauls, Jan-Hendrik and Schmidt, Benjamin and Stiller, Christoph},
  booktitle = {2021 IEEE International Conference on Robotics and Automation (ICRA)},
  title     = {Automatic Mapping of Tailored Landmark Representations for Automated Driving and Map Learning},
  year      = {2021},
  pages     = {6725-6731},
  doi       = {10.1109/ICRA48506.2021.9561432}
}

@inproceedings{carnot2026gtsign,
  author    = {Carnot, Miriam Louise and Fastermann, Erik and Kunze, Jonas and Peukert, Eric and Ludwig, André and Franczyk, Bogdan},
  title     = {GTSIGN-220: A Crowd-Sourced, StVO-Aligned Benchmark for Fine-Grained German Traffic Sign Recognition},
  booktitle = {Intelligent Vehicles Symposium (IV)},
  year      = {2026}
}

@inproceedings{wilson2021argoverse2,
  author    = {Benjamin Wilson and William Qi and Tanmay Agarwal and John Lambert and Jagjeet Singh and Siddhesh Khandelwal and Bowen Pan and Ratnesh Kumar and Andrew Hartnett and Jhony Kaesemodel Pontes and Deva Ramanan and Peter Carr and James Hays},
  title     = {Argoverse 2: Next Generation Datasets for Self-driving Perception and Forecasting},
  booktitle = {Proceedings of the Neural Information Processing Systems Track on Datasets and Benchmarks (NeurIPS Datasets and Benchmarks 2021)},
  year      = {2021}
}

@inproceedings{strauss2014calibrating,
  author    = {Strauß, Tobias and Ziegler, Julius and Beck, Johannes},
  booktitle = {17th International IEEE Conference on Intelligent Transportation Systems (ITSC)},
  title     = {Calibrating multiple cameras with non-overlapping views using coded checkerboard targets},
  year      = {2014},
  pages     = {2623-2628},
  doi       = {10.1109/ITSC.2014.6958110}
}

@inproceedings{beck2018generalized,
  author    = {Beck, Johannes and Stiller, Christoph},
  booktitle = {2018 IEEE Intelligent Vehicles Symposium (IV)},
  title     = {Generalized B-spline Camera Model},
  year      = {2018},
  pages     = {2137-2142},
  doi       = {10.1109/IVS.2018.8500466}
}

@inproceedings{carion2026sam3,
  title     = {{SAM} 3: Segment Anything with Concepts},
  author    = {Nicolas Carion and Laura Gustafson and Yuan-Ting Hu and Shoubhik Debnath and Ronghang Hu and Didac Suris Coll-Vinent and Chaitanya Ryali and Kalyan Vasudev Alwala and Haitham Khedr and Andrew Huang and Jie Lei and Tengyu Ma and Baishan Guo and Arpit Kalla and Markus Marks and Joseph Greer and Meng Wang and Peize Sun and Roman R{\"a}dle and Triantafyllos Afouras and Effrosyni Mavroudi and Katherine Xu and Tsung-Han Wu and Yu Zhou and Liliane Momeni and RISHI HAZRA and Shuangrui Ding and Sagar Vaze and Francois Porcher and Feng Li and Siyuan Li and Aishwarya Kamath and Ho Kei Cheng and Piotr Dollar and Nikhila Ravi and Kate Saenko and Pengchuan Zhang and Christoph Feichtenhofer},
  booktitle = {The Fourteenth International Conference on Learning Representations},
  year      = {2026},
  url       = {https://openreview.net/forum?id=r35clVtGzw}
}

@misc{mapillary2026,
  author       = {{Mapillary}},
  title        = {Mapillary},
  year         = {2026},
  howpublished = {\url{https://www.mapillary.com/app}},
  note         = {Street-level imagery platform. Accessed: 2026-05-04}
}

@misc{ganesan2026unidacuniversalmetricdepth,
  title         = {UniDAC: Universal Metric Depth Estimation for Any Camera},
  author        = {Girish Chandar Ganesan and Yuliang Guo and Liu Ren and Xiaoming Liu},
  year          = {2026},
  eprint        = {2603.27105},
  archiveprefix = {arXiv},
  primaryclass  = {cs.CV},
  url           = {https://arxiv.org/abs/2603.27105}
}

@inproceedings{keetha2026mapanything,
  title        = {{MapAnything}: Universal Feed-Forward Metric {3D} Reconstruction},
  author       = {Nikhil Keetha and Norman M\"{u}ller and Johannes Sch\"{o}nberger and Lorenzo Porzi and Yuchen Zhang and Tobias Fischer and Arno Knapitsch and Duncan Zauss and Ethan Weber and Nelson Antunes and Jonathon Luiten and Manuel Lopez-Antequera and Samuel Rota Bul\`{o} and Christian Richardt and Deva Ramanan and Sebastian Scherer and Peter Kontschieder},
  booktitle    = {International Conference on 3D Vision (3DV)},
  year         = {2026},
  organization = {IEEE}
}

@article{depthanything3,
  title   = {Depth Anything 3: Recovering the visual space from any views},
  author  = {Haotong Lin and Sili Chen and Jun Hao Liew and Donny Y. Chen and Zhenyu Li and Guang Shi and Jiashi Feng and Bingyi Kang},
  journal = {arXiv preprint arXiv:2511.10647},
  year    = {2025}
}

@misc{josm,
  author       = {{JOSM}},
  title        = {{J}ava {O}pen{S}treet{M}ap {E}ditor},
  howpublished = {\url{https://josm.openstreetmap.de/}},
  year         = {2026},
  note         = {Accessed: 01.05.2026}
}

@misc{openstreetmap,
  author       = {{OpenStreetMap contributors}},
  title        = {{OpenStreetMap}},
  year         = {2026},
  howpublished = {\url{https://www.openstreetmap.org}},
  note         = {Available under the Open Database License (ODbL), \url{https://www.openstreetmap.org/copyright}. Data retrieved 2026-11-06}
}

@inproceedings{lilja2024localization,
  title     = {Localization is all you evaluate: Data leakage in online mapping datasets and how to fix it},
  author    = {Lilja, Adam and Fu, Junsheng and Stenborg, Erik and Hammarstrand, Lars},
  booktitle = {Proceedings of the IEEE/CVF Conference on Computer Vision and Pattern Recognition},
  pages     = {22150--22159},
  year      = {2024}
}

@inproceedings{yuan2024streammapnet,
  title     = {Streammapnet: Streaming mapping network for vectorized online hd map construction},
  author    = {Yuan, Tianyuan and Liu, Yicheng and Wang, Yue and Wang, Yilun and Zhao, Hang},
  booktitle = {Proceedings of the IEEE/CVF Winter Conference on Applications of Computer Vision},
  pages     = {7356--7365},
  year      = {2024}
}

@inproceedings{yu2026_recondrive,
  title     = {ReconDrive: Fast Feed-Forward 4D Gaussian Splatting for Autonomous Driving Scene Reconstruction},
  author    = {Haibao Yu and Kuntao Xiao and Jiahang Wang and Ruiyang Hao and Guoran Hu and Yuxin Huang and Haifang Qin and Bowen Jing and Yuntian Bo and Ping Luo},
  booktitle = {https://arxiv.org/abs/2603.07552},
  year      = {2026}
}

@misc{nvidia2025physicalai_av,
  author       = {{NVIDIA Corporation}},
  title        = {{PhysicalAI-Autonomous-Vehicles}},
  year         = {2025},
  month        = {oct},
  publisher    = {Hugging Face},
  howpublished = {\url{https://huggingface.co/datasets/nvidia/PhysicalAI-Autonomous-Vehicles}},
  note         = {Accessed 2026-05-06, released 2025-10-28}
}

@inproceedings{yan2024street,
  title        = {Street gaussians: Modeling dynamic urban scenes with gaussian splatting},
  author       = {Yan, Yunzhi and Lin, Haotong and Zhou, Chenxu and Wang, Weijie and Sun, Haiyang and Zhan, Kun and Lang, Xianpeng and Zhou, Xiaowei and Peng, Sida},
  booktitle    = {European Conference on Computer Vision},
  pages        = {156--173},
  year         = {2024},
  organization = {Springer}
}

@inproceedings{chen2025omnire,
  title     = {OmniRe: Omni Urban Scene Reconstruction},
  author    = {Ziyu Chen and Jiawei Yang and Jiahui Huang and Riccardo de Lutio and Janick Martinez Esturo and Boris Ivanovic and Or Litany and Zan Gojcic and Sanja Fidler and Marco Pavone and Li Song and Yue Wang},
  booktitle = {The Thirteenth International Conference on Learning Representations},
  year      = {2025}
}

@article{minderer2023scaling,
  title   = {Scaling open-vocabulary object detection},
  author  = {Minderer, Matthias and Gritsenko, Alexey and Houlsby, Neil},
  journal = {Advances in Neural Information Processing Systems},
  volume  = {36},
  pages   = {72983--73007},
  year    = {2023}
}

@inproceedings{liu2024mapqr,
  title     = {Leveraging Enhanced Queries of Point Sets for Vectorized Map Construction},
  author    = {Liu, Zihao and Zhang, Xiaoyu and Liu, Guangwei and Zhao, Ji and Xu, Ningyi},
  booktitle = {European Conference on Computer Vision},
  year      = {2024}
}

@misc{wagner2026longtail,
  title         = {LongTail Driving Scenarios with Reasoning Traces: The KITScenes LongTail Dataset},
  author        = {Royden Wagner and Omer Sahin Tas and Jaime Villa and Felix Hauser and Yinzhe Shen and Marlon Steiner and Dominik Strutz and Carlos Fernandez and Christian Kinzig and Guillermo S. Guitierrez-Cabello and Hendrik Königshof and Fabian Immel and Richard Schwarzkopf and Nils Alexander Rack and Kevin Rösch and Kaiwen Wang and Jan-Hendrik Pauls and Martin Lauer and Igor Gilitschenski and Holger Caesar and Christoph Stiller},
  year          = {2026},
  eprint        = {2603.23607},
  archiveprefix = {arXiv},
  primaryclass  = {cs.CV},
  url           = {https://arxiv.org/abs/2603.23607}
}

@inproceedings{alahi2016social,
  title     = {Social LSTM: Human Trajectory Prediction in Crowded Spaces},
  author    = {Alahi, Alexandre and Goel, Kratarth and Ramanathan, Vignesh and Robicquet, Alexandre and Fei-Fei, Li and Savarese, Silvio},
  booktitle = {Proceedings of the IEEE Conference on Computer Vision and Pattern Recognition (CVPR)},
  year      = {2016}
}

@inproceedings{chen2024maptracker,
  title        = {Maptracker: Tracking with strided memory fusion for consistent vector hd mapping},
  author       = {Chen, Jiacheng and Wu, Yuefan and Tan, Jiaqi and Ma, Hang and Furukawa, Yasutaka},
  booktitle    = {European Conference on Computer Vision},
  pages        = {90--107},
  year         = {2024},
  organization = {Springer}
}

@inproceedings{zhang2025epona,
  author = {Zhang, Kaiwen and Tang, Zhenyu and Hu, Xiaotao and Pan, Xingang and Guo, Xiaoyang and Liu, Yuan and Huang,
  Jingwei and Yuan, Li and Zhang, Qian and Long, Xiao-Xiao and Cao, Xun and Yin, Wei},
  title = {Epona: Autoregressive Diffusion World Model for Autonomous Driving},
  booktitle = {Proceedings of the IEEE/CVF International Conference on Computer Vision (ICCV)},
  year = {2025}
}

@inproceedings{li2025ssr,
  title={Navigation-Guided Sparse Scene Representation for End-to-End Autonomous Driving},
  author={Peidong Li and Dixiao Cui},
  booktitle={International Conference on Learning Representations (ICLR)},
  year={2025}
}

@inproceedings{hu2023uniad,
  title={Planning-oriented autonomous driving},
  author={Hu, Yihan and Yang, Jiazhi and Chen, Li and Li, Keyu and Sima, Chonghao and Zhu, Xizhou and Chai, Siqi and Du, Senyao and Lin, Tianwei and Wang, Wenhai and others},
  booktitle={Proceedings of the IEEE/CVF conference on computer vision and pattern recognition},
  pages={17853--17862},
  year={2023}
}

@article{shen2025dmad,
  title={Divide and Merge: Motion and Semantic Learning in End-to-End Autonomous Driving},
  author={Shen, Yinzhe and Tas, Omer {\c{S}}ahin and Wang, Kaiwen and Wagner, Royden and Stiller, Christoph},
  journal={Transactions on Machine Learning Research},
  volume={2025},
  number={11},
  year={2025}
}

@article{ROS2,
    author = {Steven Macenski  and Tully Foote  and Brian Gerkey  and Chris Lalancette  and William Woodall },
    title = {Robot Operating System 2: Design, architecture, and uses in the wild},
    journal = {Science Robotics},
    volume = {7},
    number = {66},
    pages = {eabm6074},
    year = {2022},
    doi = {10.1126/scirobotics.abm6074},
    URL = {https://www.science.org/doi/abs/10.1126/scirobotics.abm6074}
}

@inproceedings{guadagnino2025kiss,
  title={KISS-SLAM: A simple, robust, and accurate 3D LiDAR SLAM system with enhanced generalization capabilities},
  author={Guadagnino, Tiziano and Mersch, Benedikt and Gupta, Saurabh and Vizzo, Ignacio and Grisetti, Giorgio and Stachniss, Cyrill},
  booktitle={2025 IEEE/RSJ International Conference on Intelligent Robots and Systems (IROS)},
  pages={5363--5370},
  year={2025},
  organization={IEEE}
}

@inproceedings{ni2025recondreamer,
  title={Recondreamer: Crafting world models for driving scene reconstruction via online restoration},
  author={Ni, Chaojun and Zhao, Guosheng and Wang, Xiaofeng and Zhu, Zheng and Qin, Wenkang and Huang, Guan and Liu, Chen and Chen, Yuyin and Wang, Yida and Zhang, Xueyang and others},
  booktitle={Proceedings of the Computer Vision and Pattern Recognition Conference},
  pages={1559--1569},
  year={2025}
}

@article{liao2022maptr,
  title={Maptr: Structured modeling and learning for online vectorized hd map construction},
  author={Liao, Bencheng and Chen, Shaoyu and Wang, Xinggang and Cheng, Tianheng and Zhang, Qian and Liu, Wenyu and Huang, Chang},
  journal={arXiv preprint arXiv:2208.14437},
  year={2022}
}

@inproceedings{wang2024stream_sqd_mapnet,
  title={Stream query denoising for vectorized hd-map construction},
  author={Wang, Shuo and Jia, Fan and Mao, Weixin and Liu, Yingfei and Zhao, Yucheng and Chen, Zehui and Wang, Tiancai and Zhang, Chi and Zhang, Xiangyu and Zhao, Feng},
  booktitle={European Conference on Computer Vision},
  pages={203--220},
  year={2024},
  organization={Springer}
}

@article{shi2024globalmapnet,
  title={Globalmapnet: An online framework for vectorized global hd map construction},
  author={Shi, Anqi and Cai, Yuze and Chen, Xiangyu and Pu, Jian and Fu, Zeyu and Lu, Hong},
  journal={arXiv preprint arXiv:2409.10063},
  year={2024}
}

@inproceedings{zhang2024enhancing_HR_mapnet,
  title={Enhancing vectorized map perception with historical rasterized maps},
  author={Zhang, Xiaoyu and Liu, Guangwei and Liu, Zihao and Xu, Ningyi and Liu, Yunhui and Zhao, Ji},
  booktitle={European Conference on Computer Vision},
  pages={422--439},
  year={2024},
  organization={Springer}
}

@inproceedings{zhang2025mapexpert,
  title={Mapexpert: Online hd map construction with simple and efficient sparse map element expert},
  author={Zhang, Dapeng and Chen, Dayu and Zhi, Peng and Chen, Yinda and Yuan, Zhenlong and Li, Chenyang and Zhou, Rui and Zhou, Qingguo and others},
  booktitle={Proceedings of the AAAI Conference on Artificial Intelligence},
  volume={39},
  number={14},
  pages={14745--14753},
  year={2025}
}

@article{yang2025histrackmap,
  title={Histrackmap: Global vectorized high-definition map construction via history map tracking},
  author={Yang, Jing and Yang, Sen and Tan, Xiao and Wang, Hanli},
  journal={arXiv preprint arXiv:2503.07168},
  year={2025}
}

@article{erdougan2025mapping_skeptic,
  title={Mapping like a Skeptic: Probabilistic BEV Projection for Online HD Mapping},
  author={Erdo{\u{g}}an, Fatih and Bar{\i}n, Merve Rabia and G{\"u}ney, Fatma},
  journal={arXiv preprint arXiv:2508.21689},
  year={2025}
}

@inproceedings{anyview,
  title={AnyView: Synthesizing Any Novel View in Dynamic Scenes},
  author={Van Hoorick, Basile and Chen, Dian and Iwase, Shun and Tokmakov, Pavel and Irshad, Muhammad Zubair and Vasiljevic, Igor and Gupta, Swati and Cheng, Fangzhou and Zakharov, Sergey and Guizilini, Vitor Campagnolo},
  booktitle={European Conference on Computer Vision (ECCV)},
  year={2026}
}

@inproceedings{jiang2025defom,
  author    = {Jiang, Hualie and Lou, Zhiqiang and Ding, Laiyan and Xu, Rui and Tan, Minglang and Jiang, Wenjie and Huang, Rui},
  title     = {DEFOM-Stereo: Depth Foundation Model Based Stereo Matching},
  booktitle = {Proceedings of the IEEE/CVF Conference on Computer Vision and Pattern Recognition (CVPR)},
  pages     = {21857--21867},
  year      = {2025}
}

@inproceedings{wen2025foundationstereo,
  author    = {Wen, Bowen and Trepte, Matthew and Aribido, Joseph and Kautz, Jan and Gallo, Orazio and Birchfield, Stan},
  title     = {FoundationStereo: Zero-Shot Stereo Matching},
  booktitle = {Proceedings of the IEEE/CVF Conference on Computer Vision and Pattern Recognition (CVPR)},
  pages     = {5249--5260},
  year      = {2025}
}

@article{chitta2023transfuser,
  title   = {{TransFuser}: Imitation with Transformer-Based Sensor Fusion for Autonomous Driving},
  author  = {Chitta, Kashyap and Prakash, Aditya and Jaeger, Bernhard and Yu, Zehao and Renz, Katrin and Geiger, Andreas},
  journal = {IEEE Transactions on Pattern Analysis and Machine Intelligence},
  year    = {2023}
}

@inproceedings{liao2025diffusiondrive,
  title     = {{DiffusionDrive}: Truncated Diffusion Model for End-to-End Autonomous Driving},
  author    = {Liao, Bencheng and Chen, Shaoyu and Yin, Haoran and Jiang, Bo and Wang, Cheng and Yan, Sixu and Zhang, Xinbang and Li, Xiangyu and Zhang, Ying and Zhang, Qian and Wang, Xinggang},
  booktitle = {Proceedings of the IEEE/CVF Conference on Computer Vision and Pattern Recognition (CVPR)},
  year      = {2025}
}

@inproceedings{dauner2024navsim,
  title     = {{NAVSIM}: Data-Driven Non-Reactive Autonomous Vehicle Simulation and Benchmarking},
  author    = {Dauner, Daniel and Hallgarten, Marcel and Li, Tianyu and Weng, Xinshuo and Huang, Zhiyu and Yang, Zetong and Li, Hongyang and Gilitschenski, Igor and Ivanovic, Boris and Pavone, Marco and Geiger, Andreas and Chitta, Kashyap},
  booktitle = {Advances in Neural Information Processing Systems (NeurIPS), Datasets and Benchmarks Track},
  year      = {2024}
}

@inproceedings{li2024egostatus,
  title     = {Is Ego Status All You Need for Open-Loop End-to-End Autonomous Driving?},
  author    = {Li, Zhiqi and Yu, Zhiding and Lan, Shiyi and Li, Jiahan and Kautz, Jan and Lu, Tong and Alvarez, Jose M.},
  booktitle = {Proceedings of the IEEE/CVF Conference on Computer Vision and Pattern Recognition (CVPR)},
  year      = {2024}
}

@article{zhai2023admlp,
  title   = {Rethinking the Open-Loop Evaluation of End-to-End Autonomous Driving in {nuScenes}},
  author  = {Zhai, Jiang-Tian and Feng, Ze and Du, Jinhao and Mao, Yongqiang and Liu, Jiang-Jiang and Tan, Zichang and Zhang, Yifu and Ye, Xiaoqing and Wang, Jingdong},
  journal = {arXiv preprint arXiv:2305.10430},
  year    = {2023}
}

@article{hirschmuller2008stereo,
  author={Hirschmuller, Heiko},
  journal={IEEE Transactions on Pattern Analysis and Machine Intelligence}, 
  title={Stereo Processing by Semiglobal Matching and Mutual Information}, 
  year={2008},
  volume={30},
  number={2},
  pages={328-341},
  doi={10.1109/TPAMI.2007.1166}
}

@misc{openscene2023,
  title={OpenScene: The Largest Up-to-Date 3D Occupancy Prediction Benchmark in Autonomous Driving},
  author={OpenScene Contributors},
  howpublished={\url{https://github.com/OpenDriveLab/OpenScene}},
  year={2023}
}
}

\newpage
\appendix

\section{Details on the Sensor Setup}
\label{sec:appx:sensor_setup_details}

\begin{figure}
  \centering
  \includegraphics[width=\linewidth]{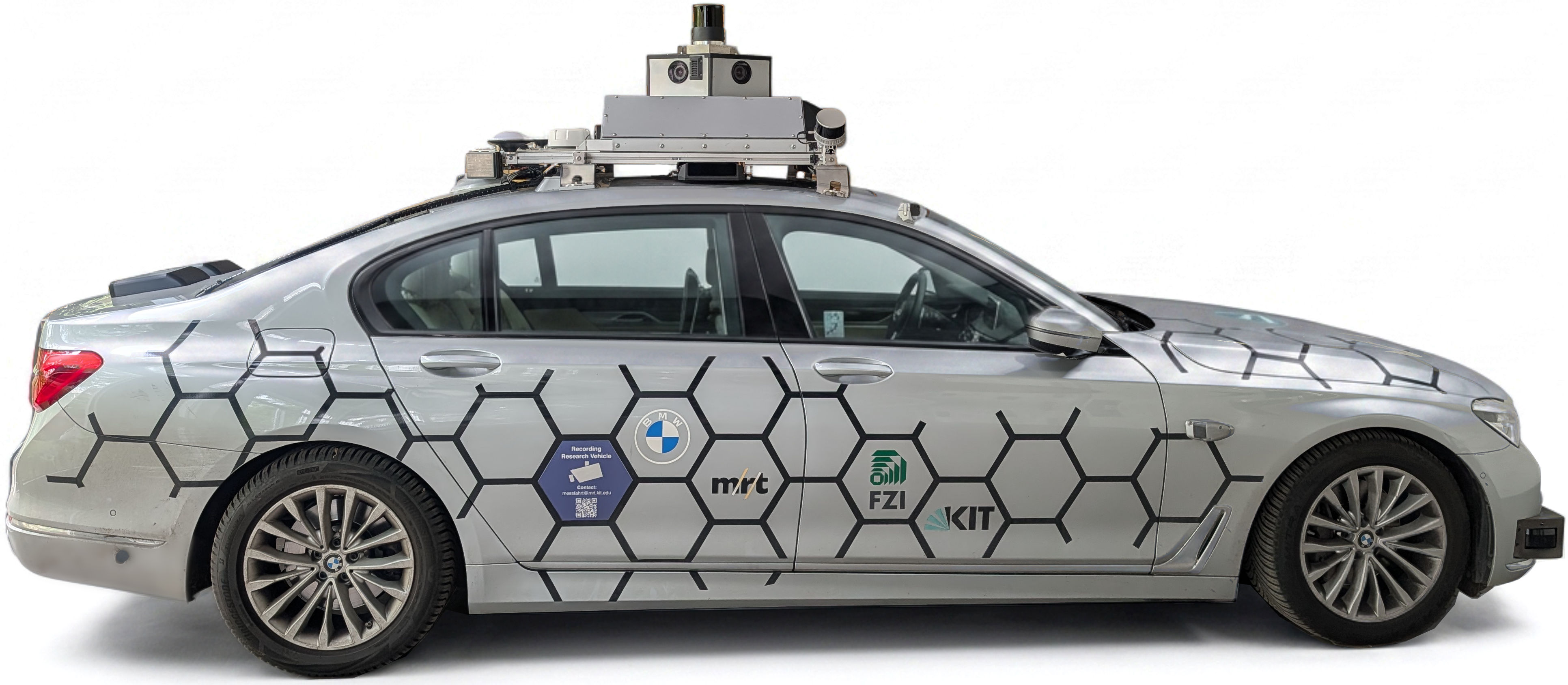}
  \caption{The KITScenes recording vehicle, a BMW 740 Li (G12), with the sensor setup as roofmount.}
  \label{fig:joy_kitscenes}
\end{figure}

\Cref{tab:camera_setup,tab:lidar_setup,tab:radar_setup,tab:gnss_setup} describe our sensor setup in detail, with a real-world picture of it shown in \Cref{fig:joy_kitscenes}.

\begin{table}[H]
  \centering
  \caption{\textbf{Camera setup.} All cameras are manufactured by Lucid Vision Labs and use low-distortion Fujinon CF8ZA-1S-23M lenses with \SI{23}{\mega\pixel} maximum resolution.}
  \label{tab:camera_setup}
  \small
  \begin{tabular}{@{}lccc@{}}
    \toprule
                  & Surround                                    & Stereo (tilted)                                & Hi-Res/Long-range \\
    \midrule
    Count         & $6$                                         & $1$ pair                                     & $1$ \\
    Camera        & ATL071S-CC                            & ATL071S-CC + ATL071S-MC                       & ATP162S-CC \\
    Sensor        & Sony IMX420, $1.1''$                & Sony IMX420, $1.1''$                 & Sony IMX542, $1.1''$ \\
    Resolution    & $3200\times2200$ (\SI{7.1}{\mega\pixel})    & $3200\times2200$ (\SI{7.1}{\mega\pixel})     & $5320\times3032$ (\SI{16.2}{\mega\pixel}) \\
    Pixel pitch   & \SI{4.5}{\micro\meter}                      & \SI{4.5}{\micro\meter}                       & \SI{4.5}{\micro\meter} \\
    FOV (H$\times$V) & \SI{87.1}{\degree}$\times$\SI{63.3}{\degree} & \SI{63.3}{\degree}$\times$\SI{86.9}{\degree} & \SI{88.4}{\degree}$\times$\SI{54.4}{\degree} \\
    \bottomrule
  \end{tabular}
\end{table}
\begin{table}[H]
  \centering
  \caption{\textbf{Lidar setup.} Per-unit specifications for the four lidar groups in the sensor suite.}
  \label{tab:lidar_setup}
  \small
  \begin{tabular}{@{}lcccc@{}}
    \toprule
                     & Top (--Dec.\ 2025) & Top (Jan.\ 2026--) & Corner (tilted) & Automotive \\
    \midrule
    Count            & $1$                                       & $1$                                       & $2$                                       & $4$ \\
    Model            & Velodyne VLS128-AP                        & Hesai OT128                               & Hesai XT32                                & Seyond Falcon K1 \\
    Channels         & $128$                                     & $128$                                     & $32$                                      & $150$ lines \\
    FOV (H$\times$V) & \SI{360}{\degree}$\times$\SI{40}{\degree} & \SI{360}{\degree}$\times$\SI{40}{\degree} & \SI{270}{\degree}$\times$\SI{31}{\degree} & \SI{120}{\degree}$\times$\SI{25}{\degree} \\
    Resolution (H$\times$V)            & \SI{0.2}{\degree}$\times$\SI{0.1}{\degree}   & \SI{0.1}{\degree}$\times$\SI{0.125}{\degree}   & \SI{0.18}{\degree}$\times$\SI{1.3}{\degree}  & \SI{0.18}{\degree}$\times$\SI{0.24}{\degree} \\
    Range (max)                        & \SI{245}{\meter}                             & \SI{230}{\meter}                               & \SI{120}{\meter}                             & \SI{500}{\meter} \\
    Range @ \SI{10}{\percent} refl.    & \SI{245}{\meter}                             & \SI{200}{\meter}                               & \SI{80}{\meter}                              & \SI{250}{\meter} \\
    Wavelength & \SI{905}{\nano\meter} & \SI{905}{\nano\meter} & \SI{905}{\nano\meter}  & \SI{1550}{\nano\meter} \\
    Effective Points/s & $2.19$\,M       & $6.91$\,M             & $864$\,k               & $900$\,k \\
    Returns          & strongest      & last + strongest       & last + strongest       & strongest \\
    \bottomrule
  \end{tabular}\\[2pt]
    {\footnotesize  The top lidar was improved in December 2025. FoV and effective points of corner lidars are intentionally limited.}
\end{table}
\begin{table}[H]
\centering
  \caption{\textbf{Radar setup.} Specifications of the three Continental ARS548 RDI 4D imaging radars.}
  \label{tab:radar_setup}
  \small
  \begin{tabular}{@{}lc@{}}
    \toprule
                                              & Long-range \\
    \midrule
    Count                                     & $3$ \\
    Model                                     & Continental ARS548 RDI \\
    Frequency band                            & \SIrange{76}{77}{\giga\hertz} \\
    FOV (H$\times$V)                          & \SI{120}{\degree}$\times$\SI{28}{\degree} \\
    Beam width (\SI{3}{\decibel}, H$\times$V) & \SI{1.2}{\degree}$\times$\SI{2.3}{\degree} \\
    Angular accuracy (H$\times$V)             & \SI{\pm 0.1}{\degree}$\times$\SI{\pm 0.1}{\degree} \\
    Range (max)                               & \SI{300}{\meter} \\
    Range resolution                          & \SI{0.22}{\meter} \\
    Velocity range                            & $-400$ to \SI{+200}{\kilo\meter\per\hour} \\
    Output                                    & 4D detections + RCS \\
    \bottomrule
  \end{tabular}
\end{table}
\begin{table}[H]
  \centering
  \caption{\textbf{GNSS and GNSS/INS setup.} We combine two receivers with independent antennas.}
  \label{tab:gnss_setup}
  \small
  \begin{tabular}{@{}lcc@{}}
    \toprule
                                          & GNSS                                              & GNSS/INS \\
    \midrule
    Model                                 & Septentrio mosaic-X5                              & Septentrio AsteRx SBi3 Pro+ \\
    Antennas                              & $1$                                        & $2$  \\
    Hardware channels                     & $448$                                             & $544$ \\
    RTK accuracy (H\,/\,V)                & \SI{0.6}{\centi\meter}\,/\,\SI{1.0}{\centi\meter} & \SI{0.6}{\centi\meter}\,/\,\SI{1.0}{\centi\meter} \\
    Standalone accuracy (H\,/\,V)         & \SI{1.2}{\meter}\,/\,\SI{1.9}{\meter}             & \SI{1.2}{\meter}\,/\,\SI{1.9}{\meter} \\
    Heading accuracy (RTK)                & ---                                               & \SI{0.2}{\degree} \\
    Pitch/roll accuracy (RTK)             & ---                                               & \SI{0.02}{\degree} \\
    Velocity accuracy                     & \SI{3}{\centi\meter\per\second}                   & \SI{2}{\centi\meter\per\second} (RTK) \\
    Position update rate                  & \SI{100}{\hertz}                                  & \SI{10}{\hertz} GNSS / \SI{50}{\hertz} integrated \\
    Integrated IMU                        & ---                                               & ADIS16500 (\SI{50}{\hertz}) \\
    \bottomrule
  \end{tabular}
\end{table}

\subsection{Sensor Data Processing and Privacy}
\label{appx:sensor_processing_privacy}

To enable both long-range perception and neural rendering applications, it is crucial to preserve the high raw image fidelity.
We record raw Bayer images and employ high-quality offline debayering pipeline using AMaZE with chromatic aberration correction~\cite{rawtherapee}.
Images are then compressed using JPEGLI~\cite{szabadka2024jpegli}, a JPEG-compatible codec, with $4\colon\!4\colon\!4$ chroma subsampling at Q95, yielding visually lossless image quality at manageable file sizes~\cite{bruse2024userspreferjpeglisamesized}.
To comply with European privacy regulations, all faces and license plates are anonymized using DNAT, a state-of-the-art inpainting method that preserves photometric realism better than conventional blurring approaches~\cite{brighterai2022dnat}.

For the \SI{360}{\degree} main lidar data, we deliberately preserve non-return information to serve as additional information for occupancy tasks.
Additionally, all Hesai lidars return dual echoes, which carry valuable information about reflective surfaces and in adverse-weather conditions~\cite{Zhao20243DRef}. The four Seyond lidars furthermore provide elongation information of each return.

\paragraph{Radar and GNSS.}
Three 4D imaging radars complement the lidar suite, providing Doppler velocity measurements and resilience under adverse weather conditions where lidar performance degrades.
A redundant combination of one GNSS receiver and one combined GNSS-INS unit provides a high-accuracy localization reference used for map validation and as a SLAM reference.

\paragraph{Localization / SLAM.}
To achieve reprojection level accuracy of georeferenced 6-DoF poses, we fuse the position data of the redundant RTK GNSS sensors into a modified version of KISS-SLAM~\cite{guadagnino2025kiss} which we intend to publish. 

\subsection{Ego Motion Compensation}

Compensating the ego motion is challenging, especially when long-range sensors yield rotationally sensitive measurements.
We found \SI{10}{\hertz} SLAM poses being too coarsely sampled to capture high-frequency pitch and roll motion and hence provide motion-compensated lidar and radar measurement using the IMU measurements at \SI{50}{\hertz}.

\subsection{Statistical Analysis}

\begin{figure}[h]
    \centering
    \caption{Lidar (top) and radar (bottom) distribution in both yaw (left) and pitch (right) direction. the radial scale shows points per scan for both Lidar and Radar on 500 train samples.}
    \includegraphics[width=\columnwidth]{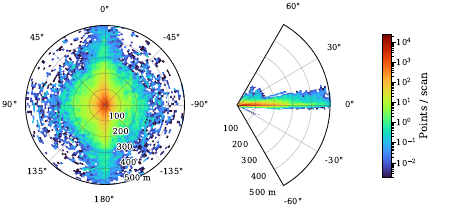}
    \includegraphics[width=\columnwidth]{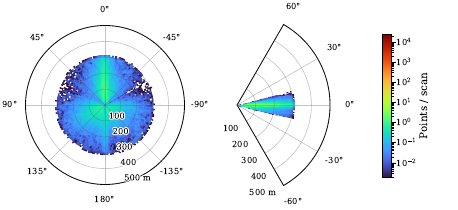}
    \label{fig:lidar_radr_distribution}
\end{figure}

In \Cref{fig:lidar_radr_distribution}, we provide a statistical analysis how lidar and radar detections are spatially distributed.
This highlights both the range and density of lidar sensors as well as the 4D resolution capabilities of our imaging radars.

Lidar intensities, return type, and elongation (for Seyond lidars) are not plotted but contained in the point clouds.
For radars, next to range, azimuth and elevation, we provide range rate (Doppler velocity), standard deviations for the four quantities, and radar cross section (RCS).

\FloatBarrier

\section{Calibration Details}
\label{sec:appx:calibration_details}

Camera calibration is significantly facilitated by using hardware-triggered global shutter cameras with low-distortion lenses.
Intrinsic and extrinsic camera calibration is performed using checkerboard targets~\cite{strauss2014calibrating} and a reference camera model~\cite{beck2018generalized}.
A pinhole model is fitted to this reference model at subpixel accuracy.

From the mechanical construction, translation and orientation are known all sensors up to few degrees and centimeters.
The remaining refinement hence focuses on the angular error that dominates at the long perception ranges that we tackle and which is best observed using far-range natural surroundings rather than close-by targets.

To avoid motion artifacts, we select one reference frame per standstill phase.
Lidar-to-lidar calibration is then formulated as joint ICP problem across all sensors and frames.
Using the same standstill frames, radar points are registered to the joint lidar point cloud using ICP.

Finally, we align the lidar and camera rigs by maximizing the reprojection of retro-reflective lidar points on semantically segmented~\cite{carion2026sam3} traffic signs and license plates using differentiable splatting.

\clearpage

\section{Additional Sample Data Visualization}

\begin{figure}[h]
    \centering
    \includegraphics[width=1\linewidth]{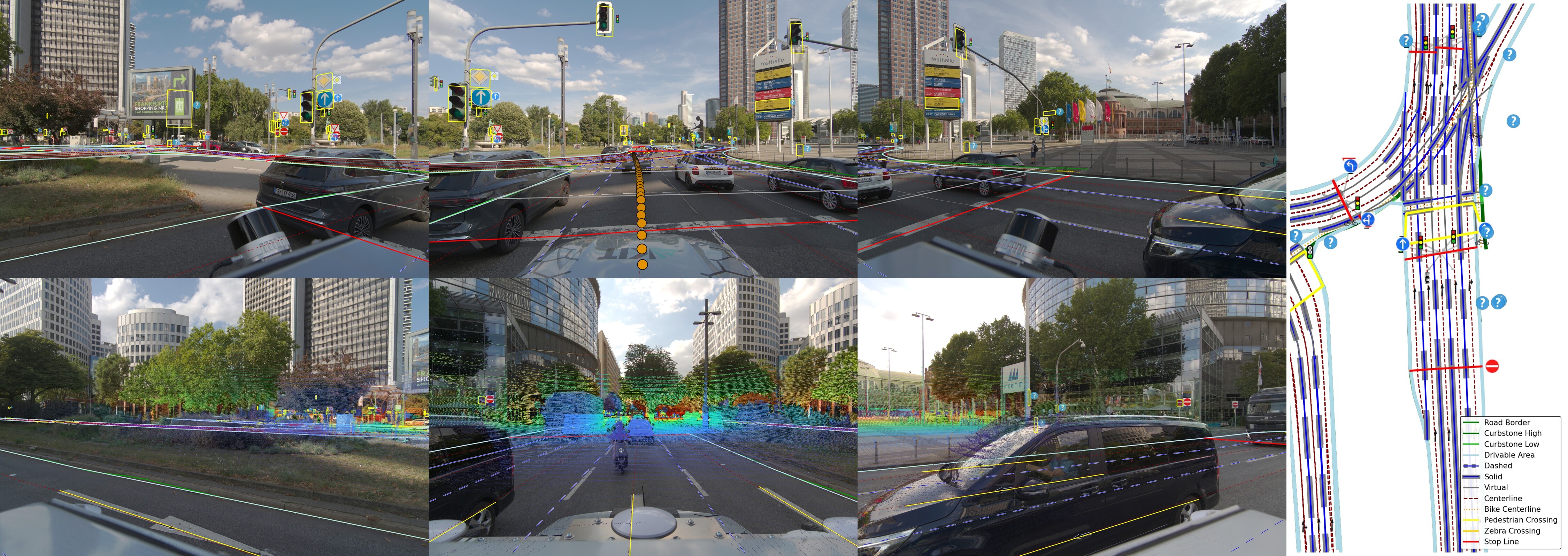}
    \hfill
    \vspace{1mm}
        \centering
        \includegraphics[width=1\linewidth]{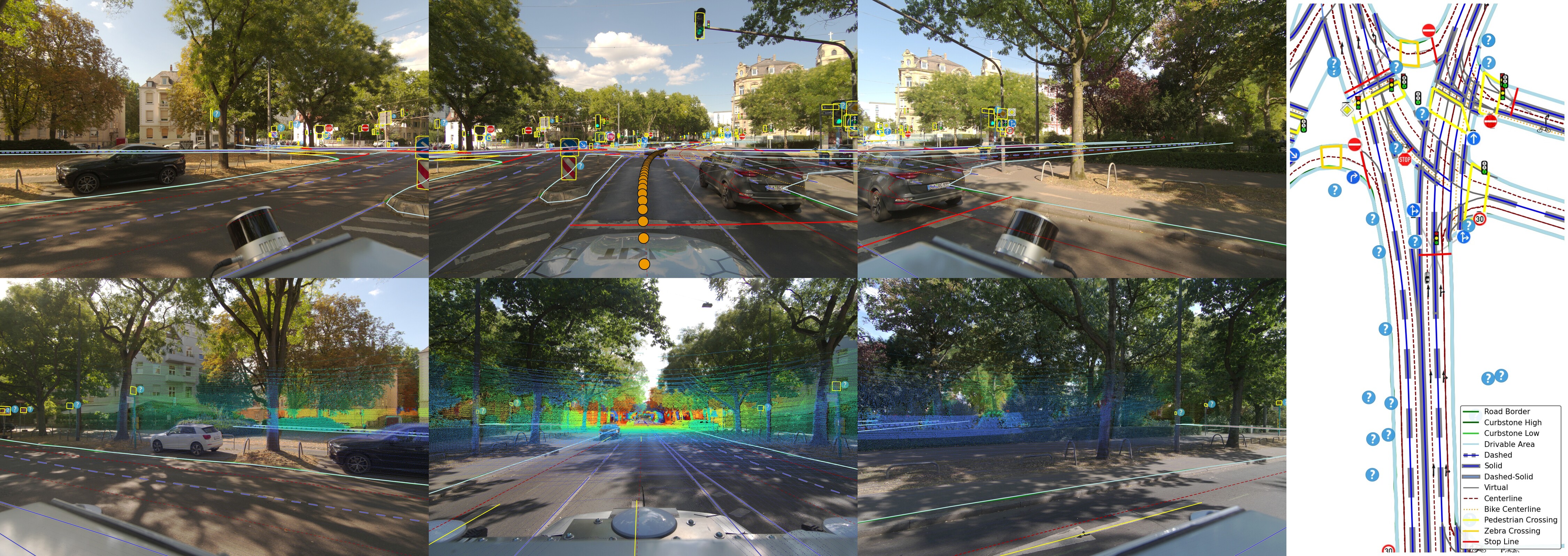}
        \caption{Example frames of a scenario in Frankfurt. The reprojected traffic lights and signs can be seen from multiple camera reprojections, highlighting the 3D nature of our HD map. The lidar data is reprojected with a transparency fade-out. The reprojected map element colors encode different class types. Icons representing class labels are transparently overlayed near the reprojected map element.   }
    \label{fig:more_teaser_samples}
\end{figure}

To give further insights into the sensor and annotation data we provide additional samples visualizations in~\Cref{fig:more_teaser_samples}.

\FloatBarrier

\section{Data Collection Routes and Conditions}

The recording vehicle, depicted in \Cref{fig:joy_kitscenes}, was human-driven by trained operators. 
All scenes are manually selected to ensure high annotation and localization quality while capturing diverse traffic scenarios and map layouts.    

\label{sec:appx:collection}

\section{Annotation Protocol and Quality Control}
\label{sec:appx:annotation}
\paragraph{Annotation Protocol and Quality Control.}
Annotation is performed by an in-house team of workers within \num{10000} total working hours (approx. \num{160} hours per \si{km^2}). Annotation tooling extends the Java OpenStreetMap Editor~\cite{josm} with tooling we developed for Lanelet2-native primitive creation, spline interpolation, routing-graph visualization, topology editing; this tooling and our class label presets will be released open-source.
Aerial imagery is sourced from municipal mapping authorities with up to \SI{6}{cm} ground sampling resolution, exceeding the resolution of public products; imagery from 2023-2024 is used and is validated frame-by-frame against the 2025/26 sensor recordings to detect map changes and adapt or exclude them for our dataset release.

Map creation is split into two complementary annotation passes.
Road-level content, like lane geometry, road markings, lane topology, crosswalks, as well as BEV traffic light and sign positions, are annotated from aerial imagery, which provides a geo-referenced, occlusion-free top-down view.
Additionally, we leverage crowdsourced streetlevel imagery~\cite{mapillary2026} to resolve ambiguities.
Elevated objects, like traffic lights, road signs, and poles, are additionally localized directly from geo-referenced onboard sensor data, yielding 3D shapes including orientation that are reprojection-accurate to the calibrated camera images~\cite{pauls2021automatic}.
Both annotation layers are fused into a unified map representation and manually reviewed for geometric and semantic correctness.
The resulting maps are further validated by automated structural-consistency checks, including Lanelet2-based topology checks and application tests in Autoware's planning simulation.

No region is annotated by a single person across all stages: annotators rotate between geometric drafting, attribute classification, and topology linking, providing implicit cross-validation and reducing systematic per-annotator artifacts.
Quality control combines (i)~Lanelet2 core-logic validators, (ii)~geometric and point/line integrity checks, (iii)~relational and topological completeness checks including a routing-graph orphan check, and (iv)~an in-house aerial-image polyline-attribute classifier that is run on every annotated polyline at QA time to flag outliers and likely tag errors against the source aerial imagery.

\section {Closed-loop autonomous driving map verification trials}
\label{sec:appx:autoware_trials}
\begin{figure}[h]
    \centering
    \includegraphics[width=\linewidth]{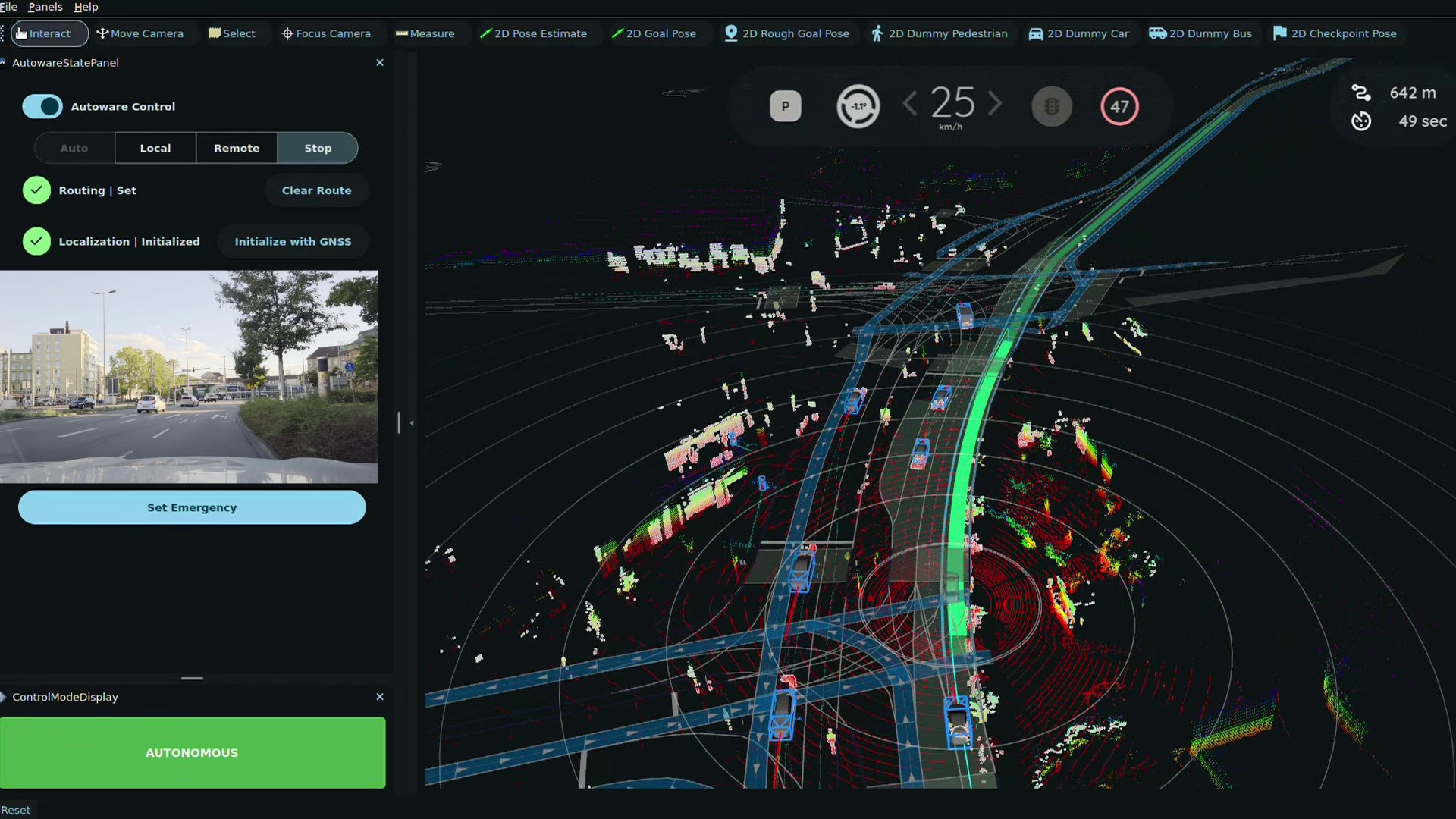}
    \caption{The open-source autonomous driving stack Autoware~\cite{autoware} driving closed-loop on one of the KITScenes Lanelet2~\cite{poggenhans2018lanelet2} HD maps}
    \label{fig:autoware}
\end{figure}

Following formal test-suite validation and simulation-based verification, closed-loop driving trials constitute the final stage of HD map qualification.
To facilitate adoption by the research community and to demonstrate the operational readiness of our maps, we validate their compatibility with Autoware~\cite{autoware}, an internationally adopted open-source autonomous driving stack that serves as the reference software platform for robotaxi deployments in Japan and beyond.
A representative closed-loop trial scene, with the Lanelet2 map visualized in RViz within a ROS~2~\cite{ROS2} environment, is depicted in \Cref{fig:autoware}. 

\clearpage

\section{Extended Dataset Statistics}
\label{sec:appx:statistics}

\begin{figure}[h]
  \centering
  \includegraphics[width=\linewidth]{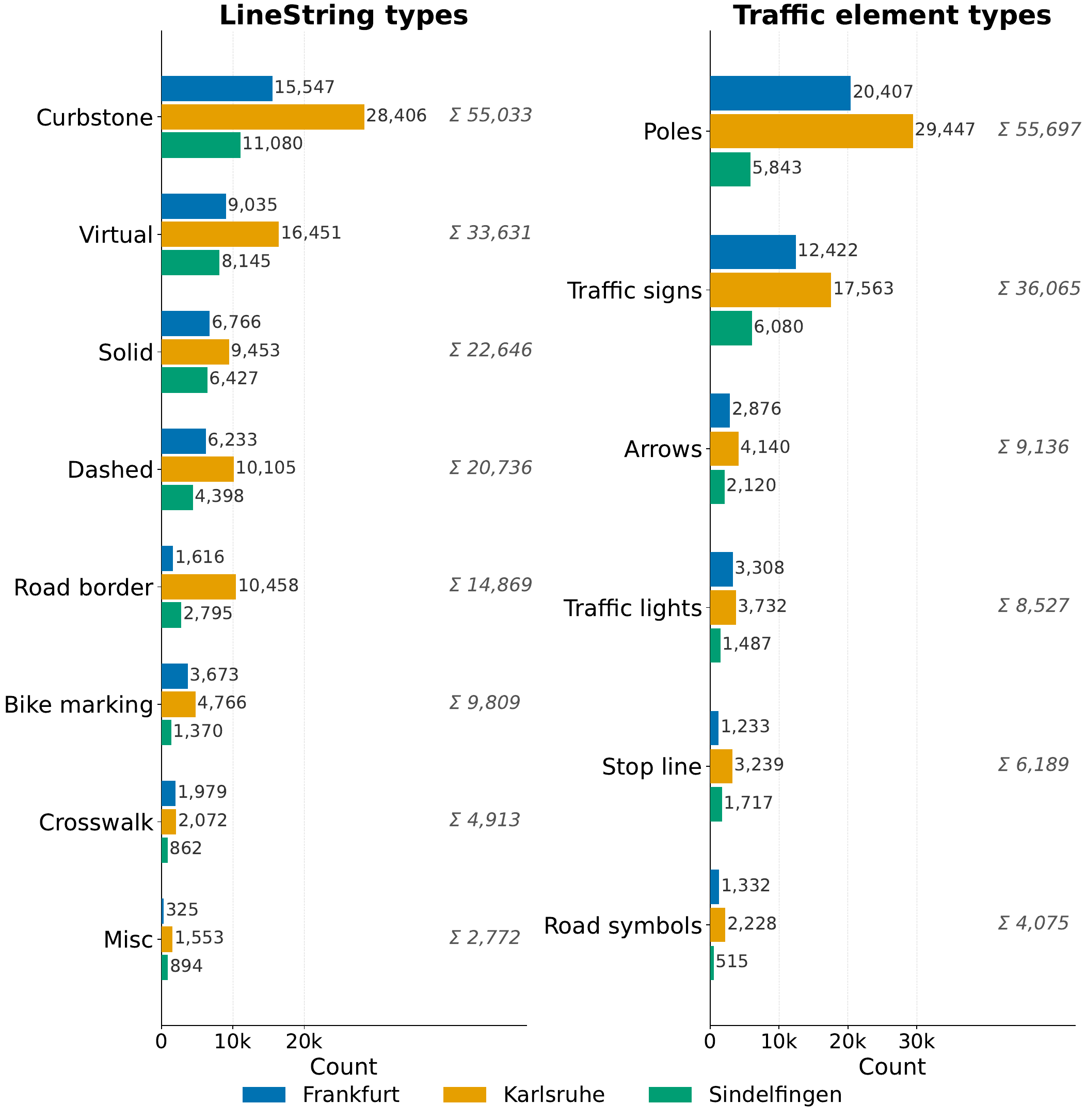}
  \caption{Binned label category statistics over Lanelet2 map elements. The left plot covers elements on the road surface as 3D polylines and the right side covers 3D traffic lights and signs as well as other lines usually not part of a lanelet border.}
  \label{fig:map_label_statistics}
\end{figure}

\paragraph{Label statistics.}
In \Cref{fig:map_label_statistics}, we depict the map element label distribution.

\begin{figure}[h]
  \centering
  \includegraphics[width=\linewidth]{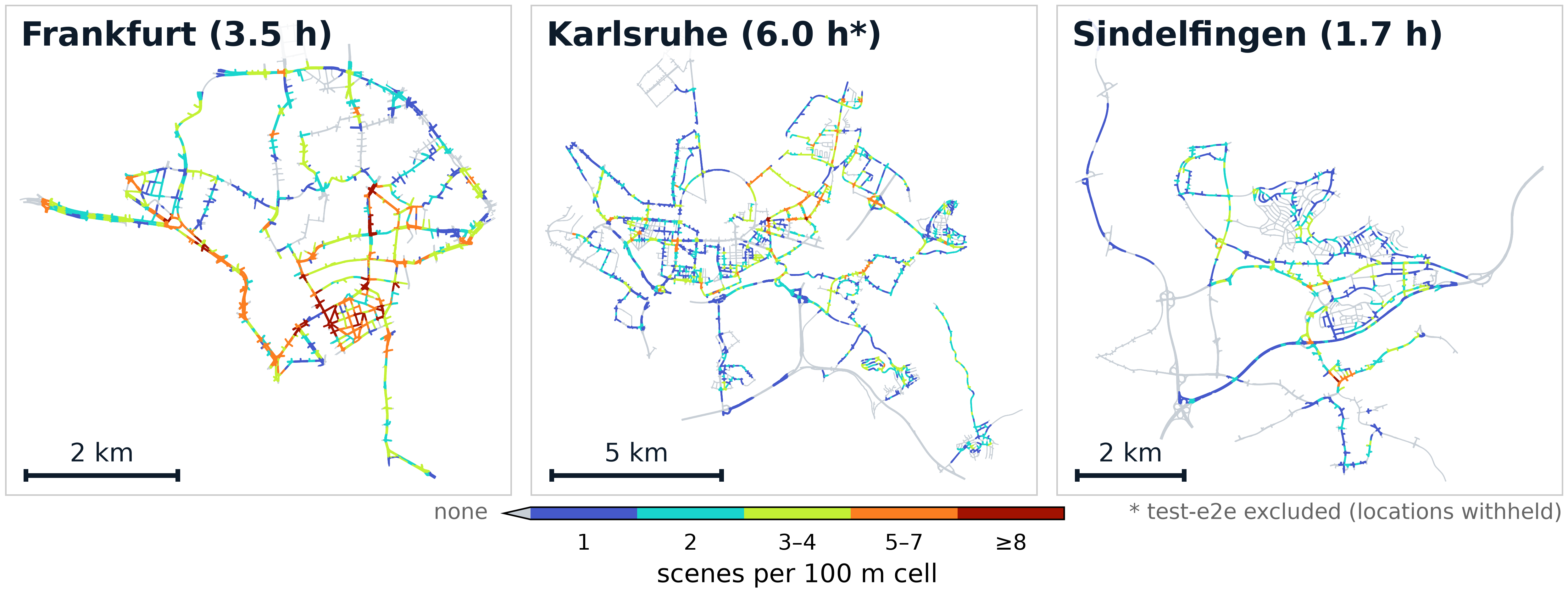}
  \caption{Repeated traversals: number of distinct scenes passing each \SI{100}{m} cell. Note that two consecutive scenes often lie within a 100m cell. }
  \label{fig:scene_coverage}
\end{figure}

\begin{figure}[h]
  \centering
  \includegraphics[width=\linewidth]{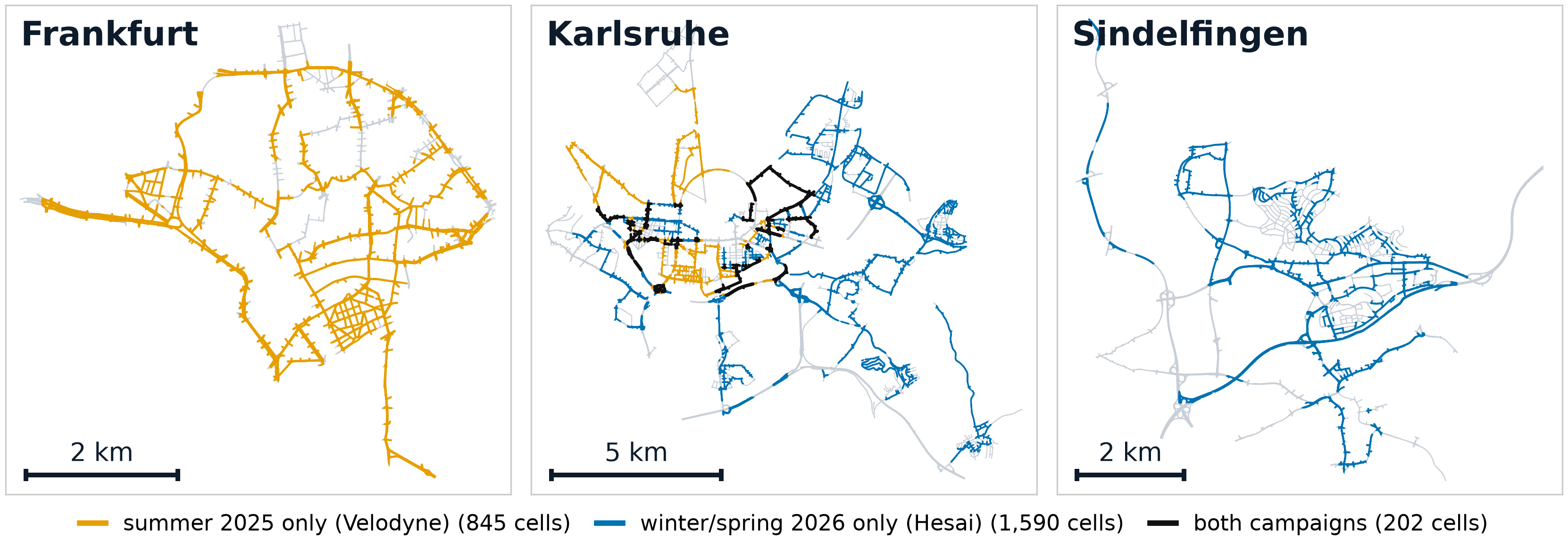}
  \caption{Recording campaigns: roads driven in summer 2025 (Velodyne lidar generation), in winter and spring 2026 (Hesai), or in both, per \SI{100}{m} cell. \texttt{test-e2e} scenarios are not included.}
  \label{fig:campaign_overlap}
\end{figure}

\paragraph{Repeated traversals.}
Many roads are driven repeatedly, which supports multi-traversal mapping and change detection.
\Cref{fig:scene_coverage} count the distinct scenes passing each \SI{100}{m} cell.
\SI{79}{\percent} of the driven cells in Frankfurt, \SI{56}{\percent} in Karlsruhe and \SI{47}{\percent} in Sindelfingen are covered by at least two scenes; \SI{50}{\percent}, \SI{27}{\percent} and \SI{11}{\percent} by at least two separate recordings (up to five, 43 recordings in total).
202 cells in central Karlsruhe were driven in both recording campaigns, summer 2025 with the Velodyne and winter 2026 with the Hesai lidar generation, so the same roads appear across seasons and sensor generations (\Cref{fig:campaign_overlap}).

\clearpage

\begin{figure}[h]
    \centering
    \includegraphics[width=\linewidth]{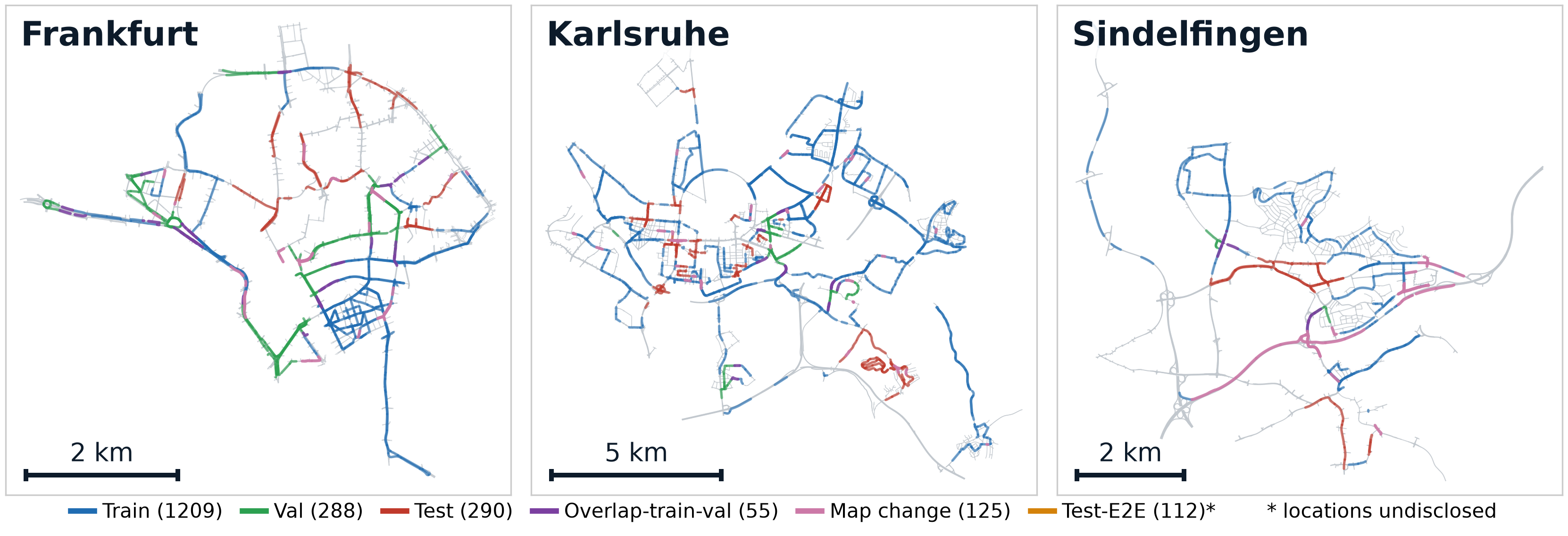}
    \caption{Visualization of the split definition for the KITScenes cities. The color indicates the split of a scenario; the counts are the number of scenarios over all cities. \texttt{test-e2e} scenarios are omitted.}
    \label{fig:splits_FRA_KA}
\end{figure}

\paragraph{Comparative Map Views}
In \Cref{fig:scene_coverage,fig:campaign_overlap,fig:splits_FRA_KA}, we provide a geo-spatial overview of traversal density, recording campaigns, and dataset splits.

\clearpage

\section{Dataset Splits}
\label{sec:appx:splits}

A persistent weakness of previous HD map construction method evaluation protocols is that training and
validation regions often overlap geographically, allowing models to implicitly memorize
map priors and inflate reported performance~\cite{lilja2024localization, yuan2024streammapnet}.

To close this loophole, we adopt a manually selected geographic split via specifically selected polygon regions of complex road layout areas for validation and test split. We greedily assign scenarios with poses in these polygon areas to the test and consecutively validation split and compute the distance between each pose of each scenario, allowing no overlap between scenario pairs including a test or test-e2e scenario up to \SI{50}{m} and between train and val scenarios up to \SI{40}{m} distance. The released city maps have the test regions cut out (a \SI{60}{m}~$\times$~\SI{60}{m} box around every pose of the test and test-e2e scenarios), so the released maps hold no map data around the test scenarios. Map-change splits hold scenes where the road changed after mapping; their map shows the old state.
The result is a strict geo-disjoint train/val/test boundary with no map overlap guaranteed
across all three splits, visualized in \Cref{fig:splits_FRA_KA}.
A unique aspect of our benchmark strategy is to withhold all map data in the test split. This is the first such held-out test set in the online HD map construction space and allows for truly trustworthy method comparisons in leaderboard challenges.

The additional test-e2e split provides only local-frame poses, no maps, and no future poses and sensor data after the keyframe of end-to-end predictions and is intended for held-out end-to-end driving evaluation.

A minor additional split called overlap-train-val is published comprising scenes with geographic overlap with the train and val sets. It joins our validation benchmark protocol for the long-range depth, novel view synthesis, and end-to-end benchmarks (Sections~\ref{sec:benchmarks:monocular_depth}--\ref{sec:benchmarks:e2e}), but is excluded from online HD map construction evaluation (Section~\ref{sec:benchmarks:online_hd_map_perception}), which requires strict train/val geo-separation.

\begin{figure}[h]
  \centering
  \includegraphics[width=\linewidth]{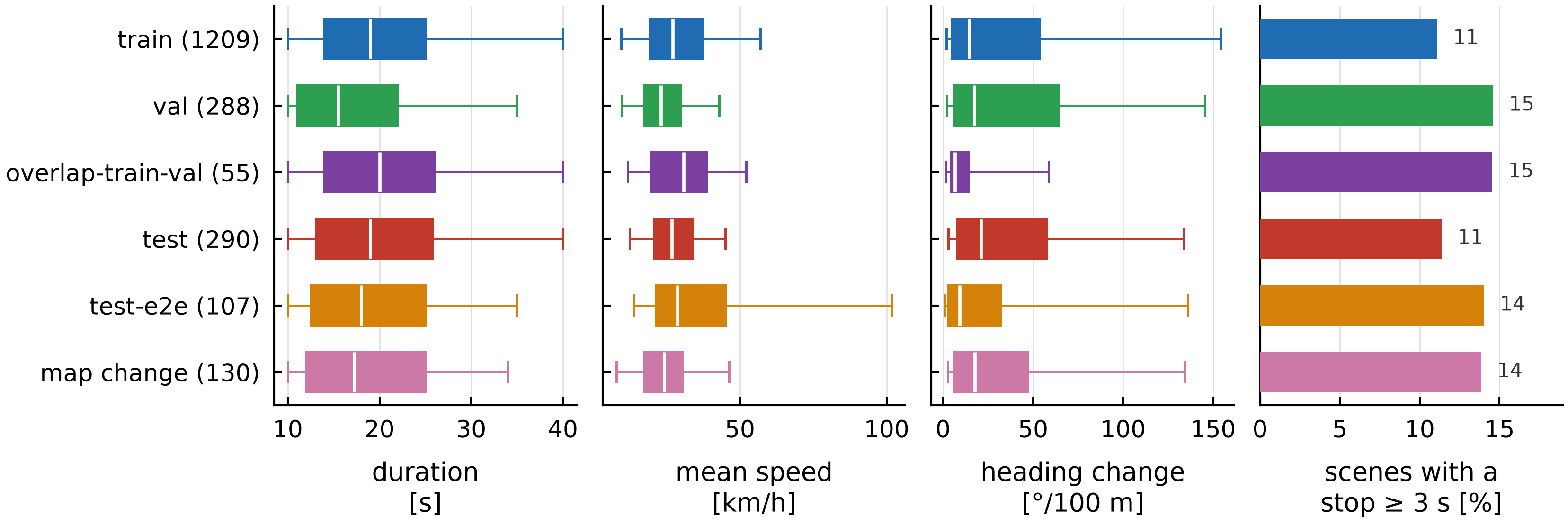}
  \caption{Scene-level statistics per split: duration, mean speed, heading change per \SI{100}{m} (boxes: quartiles, whiskers: 5--95\,\%), and the share of scenes that stand still for at least \SI{3}{\second}. Map-change splits are grouped.}
  \label{fig:scene_distributions}
\end{figure}

\Cref{fig:scene_distributions} compares the scenes of the splits: duration, speed, curvature and stops are distributed alike for train, val and test, so the geographic separation does not shift the driving conditions; test-e2e contains the fastest scenes.

\begin{figure}[t]
  \centering
  \includegraphics[width=\linewidth]{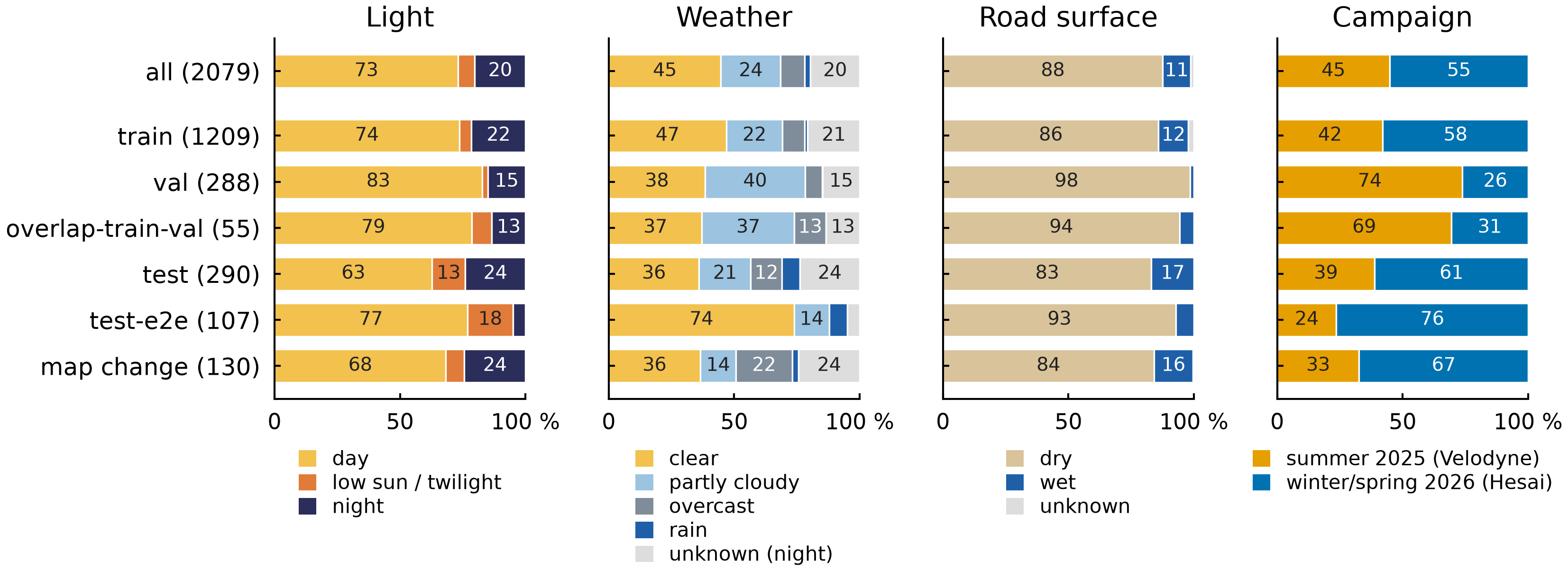}
  \caption{Recording conditions per split as shares of the recorded hours. Light from the sun elevation at the middle of each scene: day ($\geq\SI{6}{\degree}$), low sun or twilight (\SI{-6}{\degree} to \SI{6}{\degree}), night ($<\SI{-6}{\degree}$). Weather and road surface from an image classifier applied to one frame per \SI{10}{min} segment of each drive; unknown weather is almost entirely night without visible sky. Campaign: summer 2025 with the Velodyne and winter and spring 2026 with the Hesai lidar generation. Map-change splits are grouped.}
  \label{fig:recording_conditions}
\end{figure}

\Cref{fig:recording_conditions} compares the scene distribution of the splits with respect to ligthing, weather, and the condition of the road surface. Additionally the recording campaign distribution shows a bias towards the summer 2025 campaign in the val split, thus having dryer weather and road conditions.

\Cref{tab:splits_summary} provides additional statistics for the splits.

\begin{table}
  \centering
  \caption{%
    Default splits of the KITScenes Multimodal dataset (v2.0.0). A distance threshold of \SI{50}{m} to test and test-e2e scenarios and \SI{40}{m} between train and val scenarios guarantees geographically separate evaluation. Map-change splits hold scenes where the road changed after mapping; their map shows the old state.%
  }
  \label{tab:splits_summary}
  \small
  \setlength{\tabcolsep}{4pt}
  \begin{tabular}{@{}lrrrl@{}}
    \toprule
    \textbf{Split} & \textbf{Scenarios} & \textbf{Hours} & \textbf{Path~(km)} & \textbf{Released Data} \\
    \midrule
    train                          &  1209 &  7.00 &  204.0 & all \\
    val                            &   288 &  1.46 &   35.7 & all \\
    overlap-train-val              &    55 &  0.34 &    9.8 & all \\
    test                           &   290 &  1.70 &   46.7 & no maps \\
    test-e2e                       &   107 &  0.59 &   24.5 & no: maps, geo-pose, data after keyframe \\
    \midrule
    map-change-train               &    91 &  0.49 &   13.6 & all \\
    map-change-val                 &    15 &  0.08 &    1.8 & all \\
    map-change-overlap-train-val   &     2 &  0.01 &    0.3 & all \\
    map-change-test                &    17 &  0.09 &    2.0 & no maps \\
    map-change-test-e2e            &     5 &  0.03 &    0.7 & no: maps, geo-pose, data after keyframe \\
    \midrule
    \textbf{total} & \textbf{2079} & \textbf{11.79} & \textbf{339.1} & \\
    \bottomrule
  \end{tabular}
\end{table}

\FloatBarrier

\section{Per-Benchmark Details}
\label{sec:appx:benchmark_details}

\subsection{Online HD Map Construction}
\label{sec:appx:benchmark_details:online_hd_map_construction}

\paragraph{MapQR-Topo Architecture.}
The MapQR-Topo head proposed in \Cref{sec:benchmarks:online_hd_map_perception} consumes the map element tokens from the decoder and predicts pairwise relations between all predicted map elements.
Predicted relations are evaluated by matching each predicted map element to its nearest ground-truth counterpart via the Hungarian algorithm, with Chamfer distance as the cost function, and reporting a topology AP score ($\text{AP}_{Topo}$) computed over the predicted and ground-truth edges.
Contrary to the topology metric proposed by OpenLane-v2~\cite{wang2023openlanev2} ($\text{TOP}_{ll} + \text{TOP}_{lt}$), this metric design allows untangled evaluation of detection and topology prediction performance since we directly utilize the Hungarian matching algorithm already used for assigning predicted elements to GT elements. 
This avoids chamfer-distance thresholds for positive matching pairs, but instead computes the globally optimal one-to-one matching. Due to this change, even elements that fall out of the typical detection thresholds of 0.5,1.0 and 1.5 meters can be successfully evaluated with respect to their topology, when connections are predicted in agreement with the ground truth.

\begin{figure}[h]
  \centering
  \includegraphics[width=\linewidth]{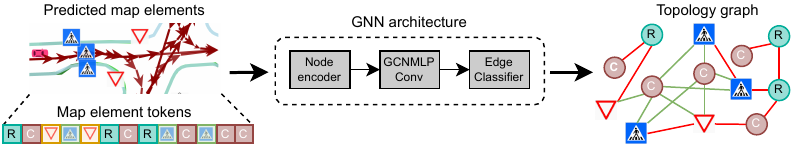}
  \caption{Schematic overview of the topology prediction with a GNN for the map elements road border (R), centerline (C) and traffic signs. Green links indicate positive link predictions, while red edges show sampled negatives, included to counteract the class imbalance introduced by the predominance of missing links.}
  \label{fig:gnn_link_pred_schema}
\end{figure}

\begin{figure}[h]
  \centering
  \begin{subfigure}{0.49\textwidth}
    \centering
    \includegraphics[width=\textwidth]{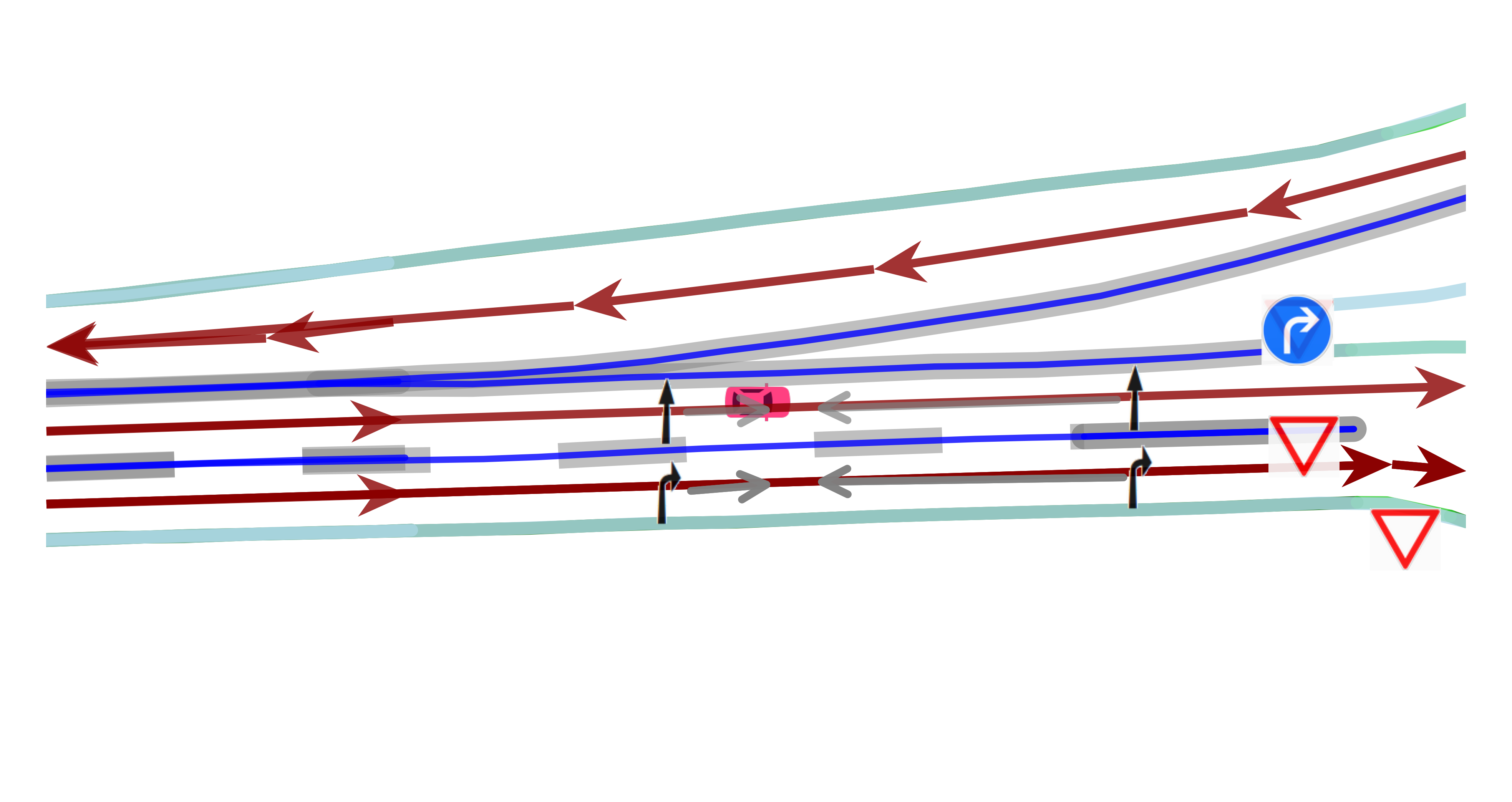}
    \caption{Ground Truth Map Scene 1}
    \label{fig:img1}
  \end{subfigure}
  \hfill
  \begin{subfigure}{0.49\textwidth}
    \centering
    \includegraphics[width=\textwidth]{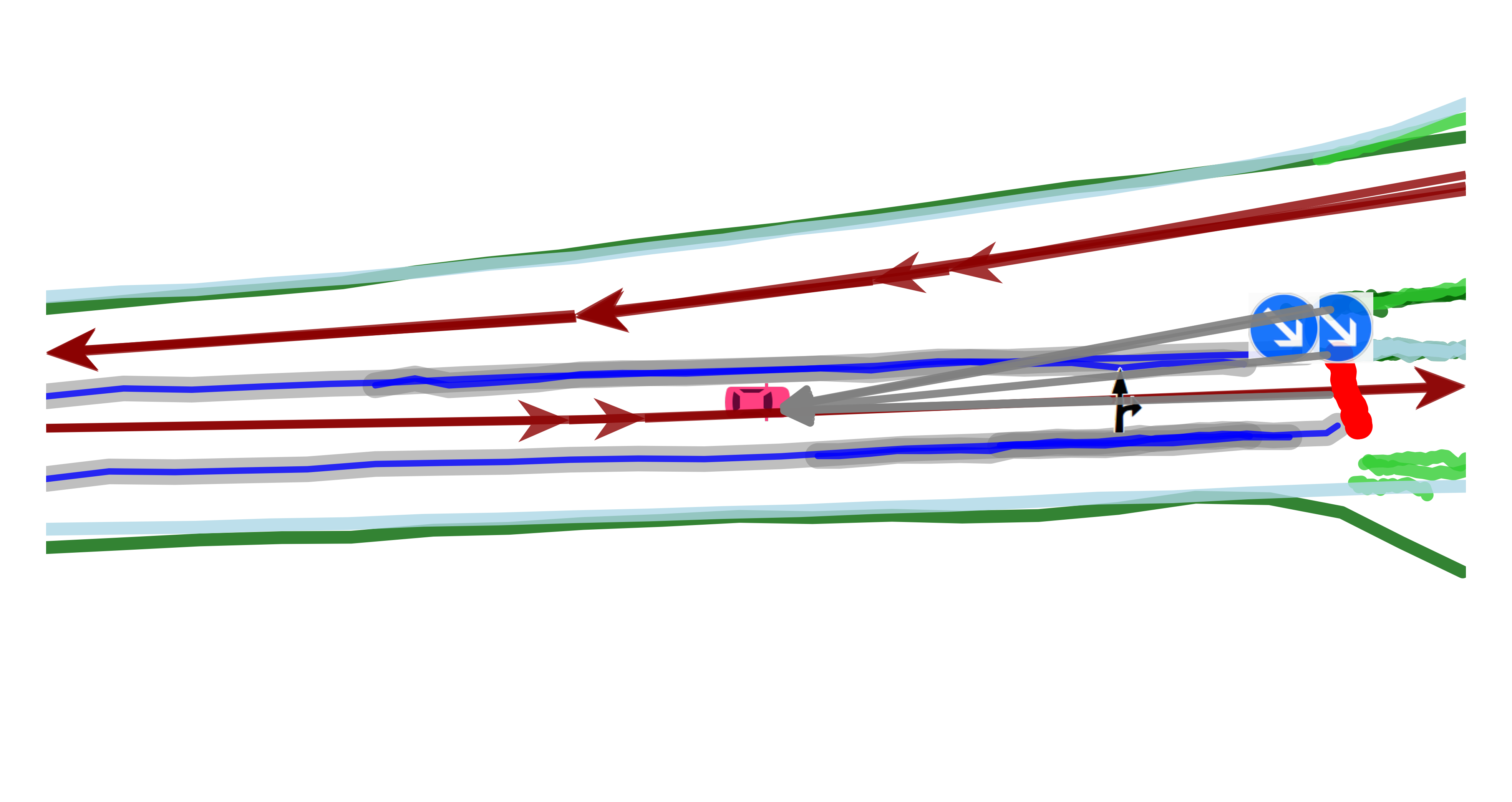}
    \caption{MapQR-Topo Prediction Scene 1}
    \label{fig:img2}
  \end{subfigure}

  \vspace{0.5em}

  \begin{subfigure}{0.49\textwidth}
    \centering
    \includegraphics[width=\textwidth]{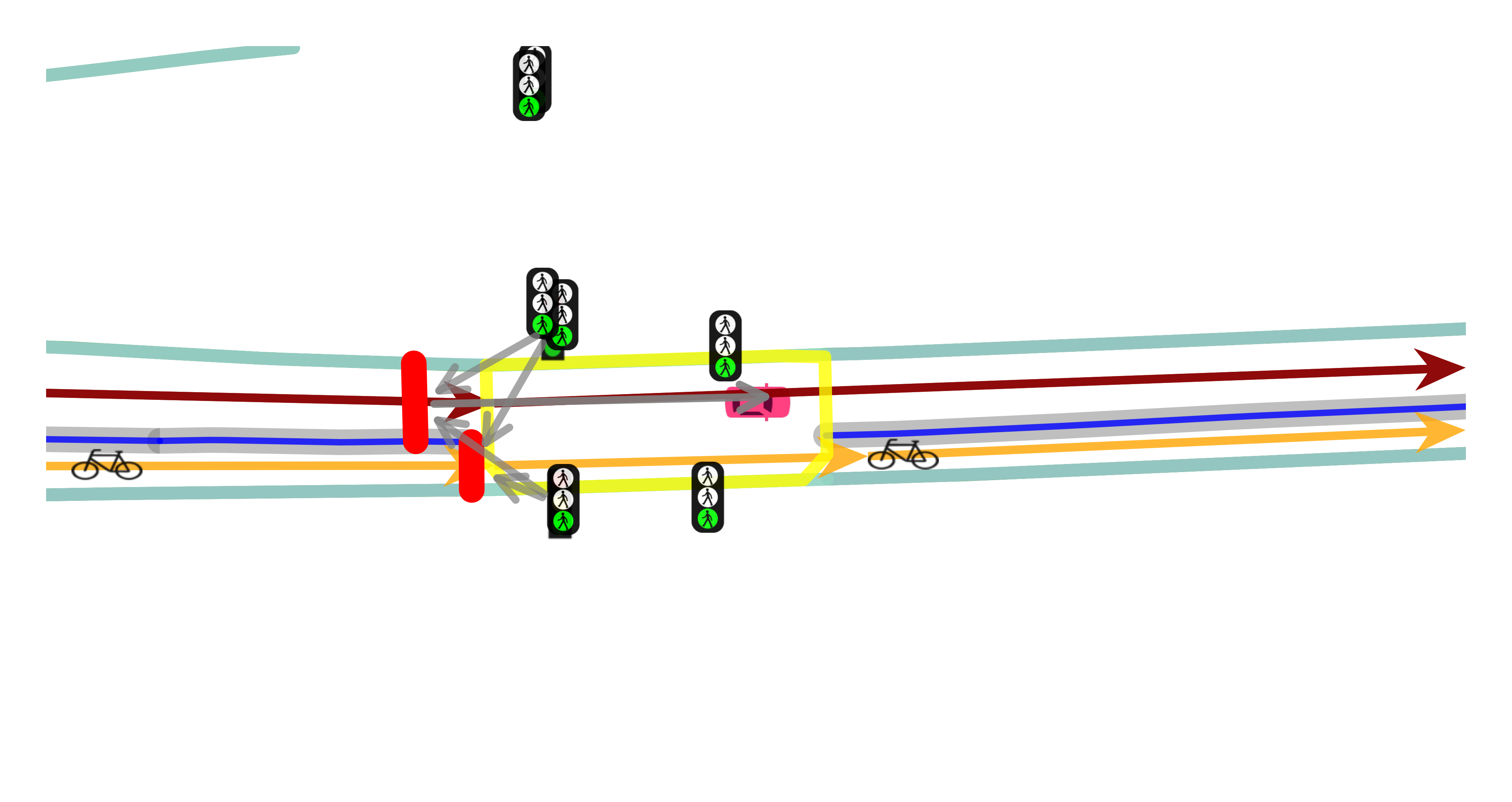}
    \caption{Ground Truth Map Scene 2}
    \label{fig:img3}
  \end{subfigure}
  \hfill
  \begin{subfigure}{0.49\textwidth}
    \centering
    \includegraphics[width=\textwidth]{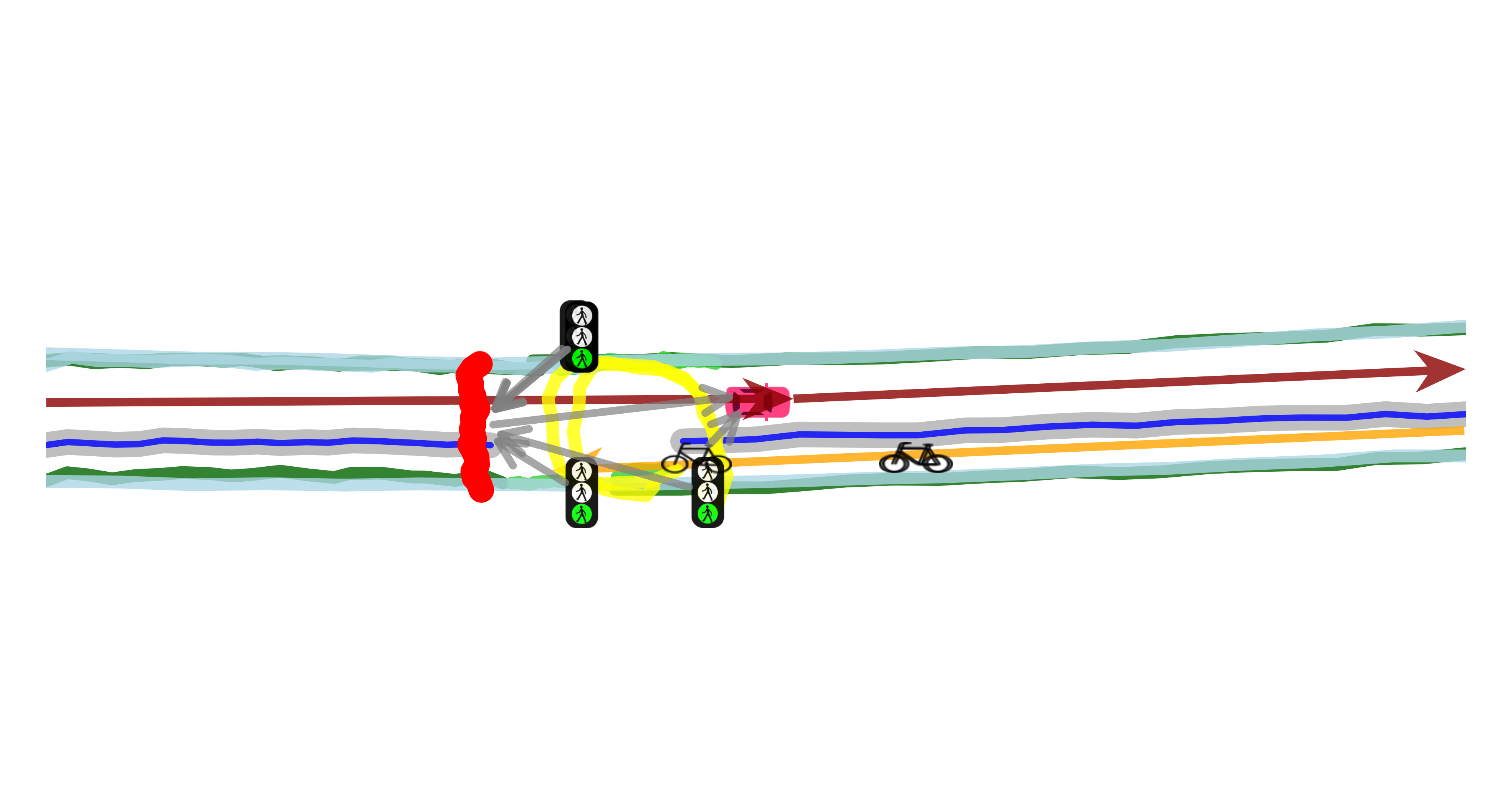}
    \caption{MapQR-Topo Prediction Scene 2}
    \label{fig:img4}
  \end{subfigure}

  \caption{Ground truth HD maps (left) and MapQR-Topo predictions (right) for two scenes. Grey edges denote topology links.}
  \label{fig:mapqr_pred}
\end{figure}

An overview of the adapted architecture is shown in \Cref{fig:gnn_link_pred_schema} and a qualitative example of the predictions in \Cref{fig:mapqr_pred}.

\paragraph{Setup.}
We employ the split described in \Cref{sec:appx:splits}, ensuring evaluation on previously unseen map areas.
An included converter~\cite{immel2024lanelet2mlconverter} translates Lanelet2 maps to and from the polyline instance graph representation used by state-of-the-art map perception models~\cite{maptrv2}, making it straightforward to apply and evaluate existing methods on our benchmark. For our benchmark, we use a subset of 120 out of 220 available traffic sign classes, retaining only the most common and semantically relevant signs. All methods are trained for 6 epochs with a batch size of 16 in line with standard evaluation settings on comparable dataset sizes~\cite{maptrv2}.

\paragraph{Metrics.}
We follow the standard protocol of~\cite{maptrv2}, reporting Average Precision (AP) with Chamfer distance thresholds of \SI{0.5}{\meter}, \SI{1.0}{\meter}, and \SI{1.5}{\meter}, averaged across the map element classes with ground truth in the evaluated frames.
To preserve readability, we group map element classes into six categories: Lane Markings (LM), Lane Centerlines (LC), Road Infrastructure (RI), Traffic Lights (TL), Traffic Signs (TS), and Road Markings (RM); the full assignment is given in~\Cref{tab:map_elem_to_category}.
For topology evaluation, we match each predicted map element to its nearest ground-truth counterpart via the Hungarian algorithm with Chamfer distance as the cost function, and report a topology AP score ($\text{AP}_{Topo}$) computed over the predicted and ground-truth edges.

\begin{table}[h]
  \centering
  \small
  \caption{Assignment of the map element classes to the six aggregated reporting categories used in \Cref{tab:detection_results}.}
  \renewcommand{\arraystretch}{1.2}
  \resizebox{0.8\textwidth}{!}{%
  \begin{tabular}{p{4cm} p{4cm} p{4cm}}
    \toprule
    \textbf{Lane Markings (LM)} & \textbf{Lane Centerlines (LC)} & \textbf{Road Infrastructure (RI)} \\
    \midrule
    \parbox[t]{2.8cm}{
      road\_border \\ dashed \\ solid \\ solid\_solid \\
      solid\_dashed \\ dashed\_solid \\ bike\_marking\_dashed \\
      bike\_marking\_solid \\ pedestrian\_crossing \\
      zebra\_crossing \\ te\_stop\_line
    }
    &
    \parbox[t]{2.3cm}{
      centerline \\
      bike\_centerline
    }
    &
    \parbox[t]{2.3cm}{
      curbstone\_high \\ curbstone\_low \\ fence \\
      building \\ wall \\ guard\_rail \\
      drivable\_area \\ divider
    }
    \\
    \bottomrule
  \end{tabular}%
  }

  \vspace{1em}

  \resizebox{0.8\textwidth}{!}{%
  \begin{tabular}{p{4cm} p{4cm} p{4cm}}
    \toprule
    \textbf{Traffic Lights (TL)} & \textbf{Traffic Signs (TS)} & \textbf{Road Markings (RM)} \\
    \midrule
    \parbox[t]{1.8cm}{
      te\_tl\_car \\ te\_tl\_bike \\
      te\_tl\_pedestrian \\ te\_tl\_misc
    }
    &
    \parbox[t]{3.2cm}{
      te\_ts\_stop \\ te\_ts\_yield \\ te\_ts\_no\_entry \\
      te\_ts\_right\_of\_way \\ te\_ts\_priority\_road \\
      te\_ts\_one\_way\_street \\ te\_ts\_roundabout \\
      te\_ts\_speed\_limit \\ te\_ts\_pedestrian\_crossing \\
      te\_ts\_turn\_right \\ te\_ts\_turn\_left \\
      te\_ts\_go\_straight \\ te\_ts\_go\_straight\_or\_right \\
      te\_ts\_go\_straight\_or\_left \\ te\_ts\_turn\_left\_or\_right \\
      te\_ts\_pass\_right \\ te\_ts\_pass\_left \\ te\_ts\_crossbuck
    }
    &
    \parbox[t]{3.5cm}{
      te\_arrow\_go\_straight \\ te\_arrow\_turn\_left \\
      te\_arrow\_turn\_right \\ te\_arrow\_go\_straight\_or\_left \\
      te\_arrow\_go\_straight\_or\_right \\
      te\_arrow\_turn\_left\_or\_right \\
      te\_bike\_symbol \\ te\_bus\_symbol \\
      te\_symbol30 \\ te\_symbol50 \\ te\_symbol70
    }
    \\
    \bottomrule
  \end{tabular}%
  }
  \label{tab:map_elem_to_category}
\end{table}

\paragraph{More Evaluation Details.}
The online HD map construction models are trained on the train split of v2.0.0 with a \SI{60}{m}~$\times$~\SI{30}{m} ego-centred range and evaluated after the last (6th) epoch on every fourth frame of the val split (\num{13123} frames).
Eight of its 54 classes have no ground truth in these frames (fence, building, divider, bike\_marking\_dashed, bike\_marking\_solid, te\_tl\_misc, te\_symbol50, te\_symbol70) and are left out of the averages, as the standard COCO mAP detection protocol prescribes.
The labels of all methods are generated with the Lanelet2 ML converter~\cite{immel2024lanelet2mlconverter} version 1.2.3+kitscenes3 (released as Lanelet2 fork at \url{https://github.com/KITScenes/Lanelet2}).
This version does not assign the bike\_marking and divider classes: bicycle lane markings are labeled as the generic dashed, solid or unknown lane markings.
A Lanelet2 ML converter version that handles these bike\_marking subclasses separately will follow in the next Lanelet2 release on \url{https://github.com/KITScenes/Lanelet2}.

\FloatBarrier

\subsection{Long-range Depth Estimation}
\label{sec:appx:benchmark_details:depth}

\paragraph{Setup.}
Monocular methods use RGB images from our front-facing \qty{16.2}{\mega\pixel} high-resolution long-range camera, while stereo methods use rectified image pairs from the tilted \qty{7.1}{\mega\pixel} forward-facing stereo setup (see~\Cref{sec:dataset:sensors}). We used the best self-reported pretrained weights for \cite{ganesan2026unidacuniversalmetricdepth, depthanything3, keetha2026mapanything, jiang2025defom, wen2025foundationstereo}. For SGBM~\cite{hirschmuller2008stereo}, we densify the semi-dense disparity with an weighted least squares (WLS) filter. 
Outputs are upscaled to camera resolution where necessary.
Ground-truth depth maps are produced by fusing motion-compensated lidar point clouds (all sensors from \Cref{sec:dataset:sensors}) over a \qty{\pm 1}{\second} temporal window. 
The merged cloud is projected onto the camera plane at $2\times$ super-resolution using point splatting, and outliers are removed via MAD-based consistency rejection and edge erosion at depth discontinuities. 
The super-resolved map is then block-median downsampled to native resolution. 
Only pixels with valid depth values are used for evaluation. 
Frames are sampled at \qty{0.1}{\hertz} (one per \qty{10}{\second}), each placed at the temporal midpoint of its \qty{10}{\second} window.

\begin{table}[h]
  \centering
  \caption{Monocular depth estimation - 0--100\,m}
  \label{tab:depth-allmetrics-full-0-100m}
  \resizebox{\textwidth}{!}{%
  \setlength{\tabcolsep}{3pt}
  \renewcommand{\arraystretch}{0.85}
  \begin{tabular}{@{}l c c c c c c c c c c c c@{}}
    \toprule
    Method & Abs.Rel $\downarrow$ & Sq.Rel $\downarrow$ & MAE\,[m] $\downarrow$ & RMSE\,[m] $\downarrow$ & IMAE $\downarrow$ & IRMSE $\downarrow$ & LogMAE $\downarrow$ & LogRMSE $\downarrow$ & SILog $\downarrow$ & $\delta_{1}$ $\uparrow$ & $\delta_{2}$ $\uparrow$ & $\delta_{3}$ $\uparrow$ \\
    \midrule
  UniDAC & 0.393 & 0.222 & 6.441 & 9.089 & 0.043 & 0.050 & 0.373 & 0.412 & \textbf{0.216} & 23.42 & 83.87 & 93.80 \\
  DA3 & 0.273 & 0.226 & 5.774 & 10.408 & 0.029 & 0.046 & 0.286 & 0.364 & 0.261 & 51.13 & 81.16 & 92.51 \\
  MapAny & \textbf{0.153} & \textbf{0.099} & \textbf{3.525} & \textbf{7.557} & \textbf{0.016} & \textbf{0.036} & \textbf{0.157} & \textbf{0.251} & 0.230 & \textbf{81.06} & \textbf{91.22} & \textbf{96.45} \\
    \bottomrule
  \end{tabular}%
  }
  
  \centering
  \caption{Monocular depth estimation - 100--200\,m}
  \label{tab:depth-allmetrics-full-100-200m}
  \resizebox{\textwidth}{!}{%
  \setlength{\tabcolsep}{3pt}
  \renewcommand{\arraystretch}{0.85}
  \begin{tabular}{@{}l c c c c c c c c c c c c@{}}
    \toprule
    Method & Abs.Rel $\downarrow$ & Sq.Rel $\downarrow$ & MAE\,[m] $\downarrow$ & RMSE\,[m] $\downarrow$ & IMAE $\downarrow$ & IRMSE $\downarrow$ & LogMAE $\downarrow$ & LogRMSE $\downarrow$ & SILog $\downarrow$ & $\delta_{1}$ $\uparrow$ & $\delta_{2}$ $\uparrow$ & $\delta_{3}$ $\uparrow$ \\
    \midrule
  UniDAC & \textbf{0.327} & \textbf{0.157} & \textbf{44.274} & \textbf{49.893} & 0.011 & 0.012 & \textbf{0.492} & \textbf{0.547} & \textbf{0.224} & \textbf{35.49} & \textbf{60.00} & \textbf{76.70} \\
  DA3 & 0.481 & 0.276 & 64.657 & 68.947 & \textbf{0.011} & \textbf{0.012} & 0.750 & 0.793 & 0.229 & 11.44 & 28.65 & 48.82 \\
  MapAny & 0.516 & 0.309 & 69.591 & 74.948 & 0.012 & 0.014 & 0.822 & 0.889 & 0.292 & 10.73 & 20.48 & 37.22 \\
    \bottomrule
  \end{tabular}%
  }
  
  \centering
  \caption{Monocular depth estimation - $>200$\,m}
  \label{tab:depth-allmetrics-full-200mplus}
  \resizebox{\textwidth}{!}{%
  \setlength{\tabcolsep}{3pt}
  \renewcommand{\arraystretch}{0.85}
  \begin{tabular}{@{}l c c c c c c c c c c c c@{}}
    \toprule
    Method & Abs.Rel $\downarrow$ & Sq.Rel $\downarrow$ & MAE\,[m] $\downarrow$ & RMSE\,[m] $\downarrow$ & IMAE $\downarrow$ & IRMSE $\downarrow$ & LogMAE $\downarrow$ & LogRMSE $\downarrow$ & SILog $\downarrow$ & $\delta_{1}$ $\uparrow$ & $\delta_{2}$ $\uparrow$ & $\delta_{3}$ $\uparrow$ \\
    \midrule
  UniDAC & \textbf{0.567} & \textbf{0.348} & \textbf{149.764} & \textbf{157.268} & \textbf{0.012} & \textbf{0.012} & \textbf{0.950} & \textbf{0.983} & 0.218 & \textbf{1.22} & \textbf{10.65} & \textbf{30.12} \\
  DA3 & 0.708 & 0.518 & 185.439 & 191.202 & 0.012 & 0.013 & 1.329 & 1.348 & \textbf{0.198} & 0.56 & 2.00 & 5.77 \\
  MapAny & 0.778 & 0.614 & 202.947 & 208.399 & 0.017 & 0.018 & 1.595 & 1.618 & 0.240 & 0.00 & 0.85 & 1.29 \\
    \bottomrule
  \end{tabular}%
  }

  \centering
  \caption{Monocular depth estimation - Overall}
  \label{tab:depth-allmetrics-full-overall}
  \resizebox{\textwidth}{!}{%
  \setlength{\tabcolsep}{3pt}
  \renewcommand{\arraystretch}{0.85}
  \begin{tabular}{@{}l c c c c c c c c c c c c@{}}
    \toprule
    Method & Abs.Rel $\downarrow$ & Sq.Rel $\downarrow$ & MAE\,[m] $\downarrow$ & RMSE\,[m] $\downarrow$ & IMAE $\downarrow$ & IRMSE $\downarrow$ & LogMAE $\downarrow$ & LogRMSE $\downarrow$ & SILog $\downarrow$ & $\delta_{1}$ $\uparrow$ & $\delta_{2}$ $\uparrow$ & $\delta_{3}$ $\uparrow$ \\
    \midrule
  UniDAC & 0.392 & 0.221 & 7.322 & \textbf{12.726} & 0.042 & 0.049 & 0.376 & 0.418 & \textbf{0.239} & 23.57 & 83.35 & 93.40 \\
  DA3 & 0.277 & 0.228 & 7.060 & 15.352 & 0.029 & 0.046 & 0.295 & 0.380 & 0.278 & 50.41 & 80.17 & 91.65 \\
  MapAny & \textbf{0.160} & \textbf{0.103} & \textbf{4.963} & 14.439 & \textbf{0.016} & \textbf{0.035} & \textbf{0.170} & \textbf{0.284} & 0.262 & \textbf{79.79} & \textbf{89.92} & \textbf{95.33} \\
    \bottomrule
  \end{tabular}%
  }
\end{table}
\begin{figure}[!thp]
    \centering
    \includegraphics[width=\textwidth]{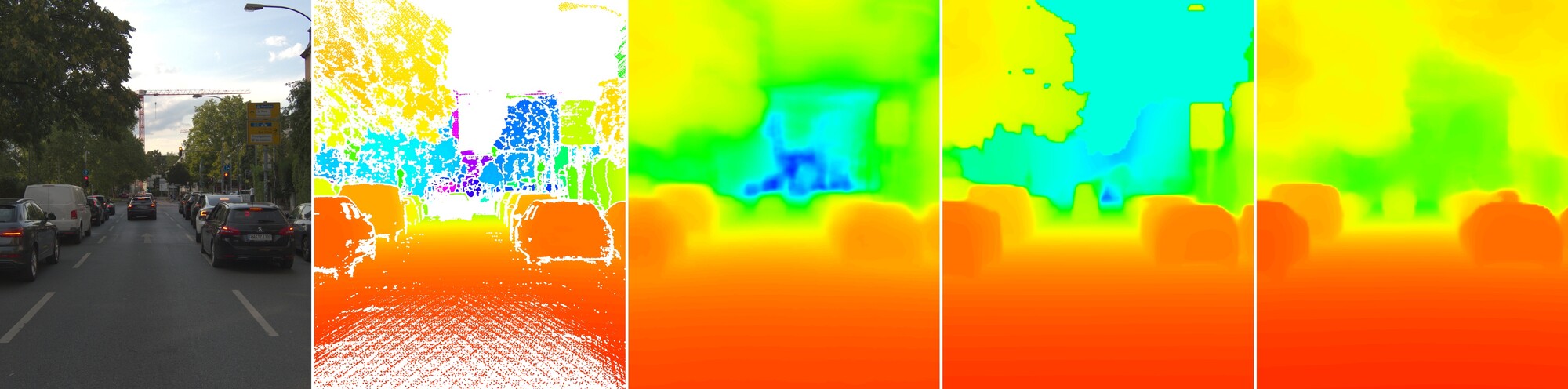}
    \includegraphics[width=\textwidth]{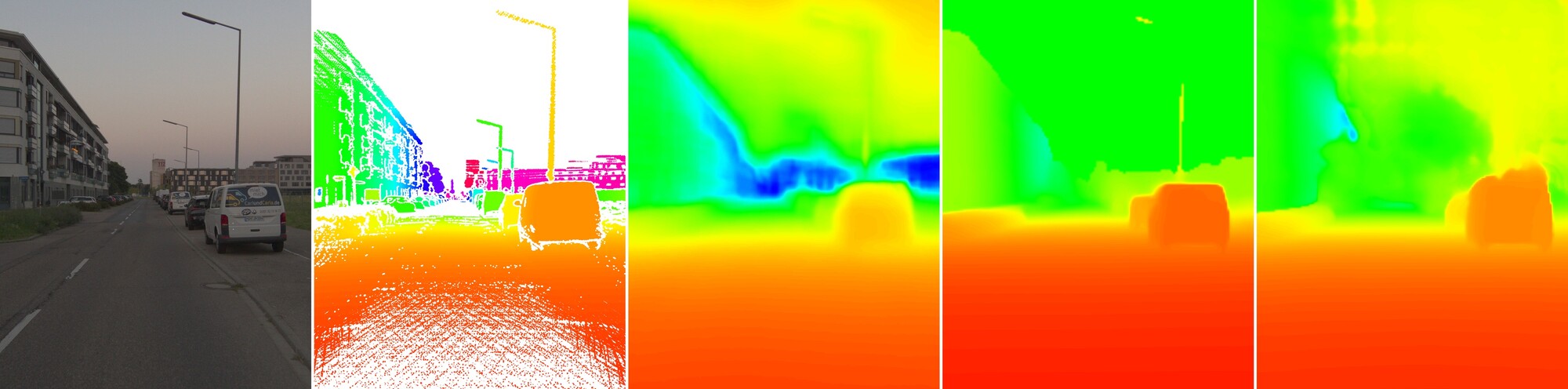}
    \includegraphics[width=\textwidth]{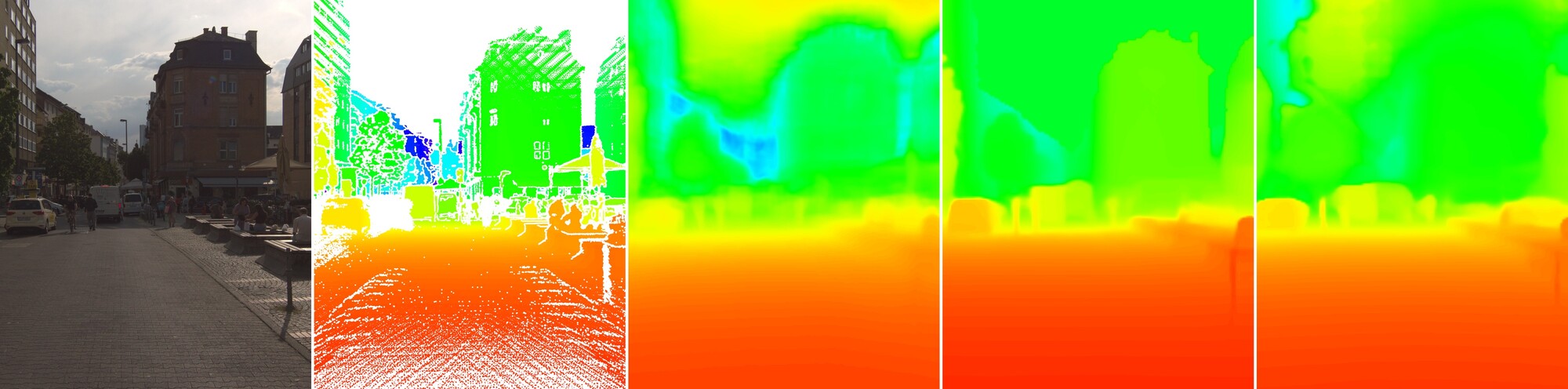}
    \includegraphics[width=\textwidth]{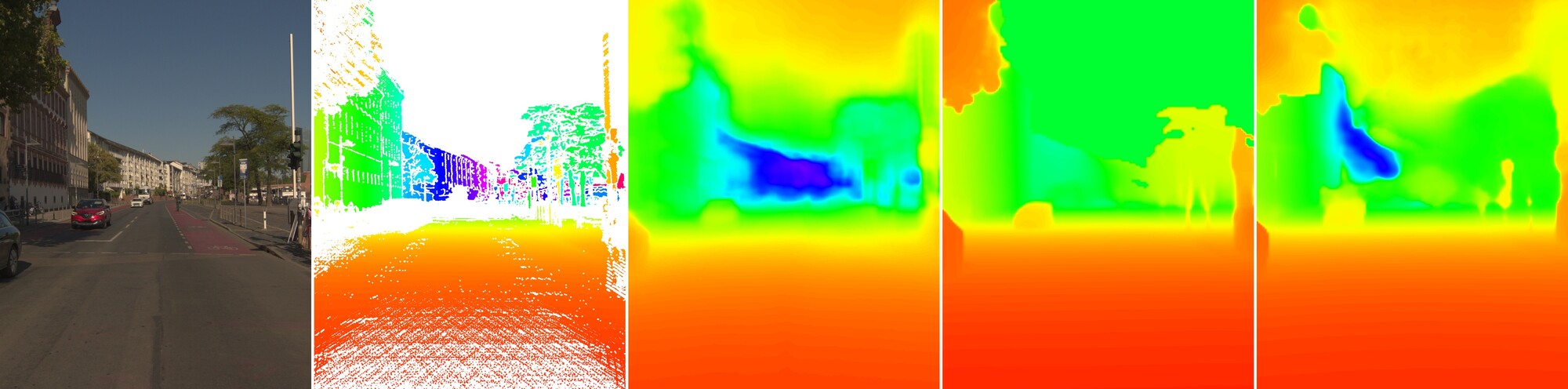}
    \includegraphics[width=\textwidth]{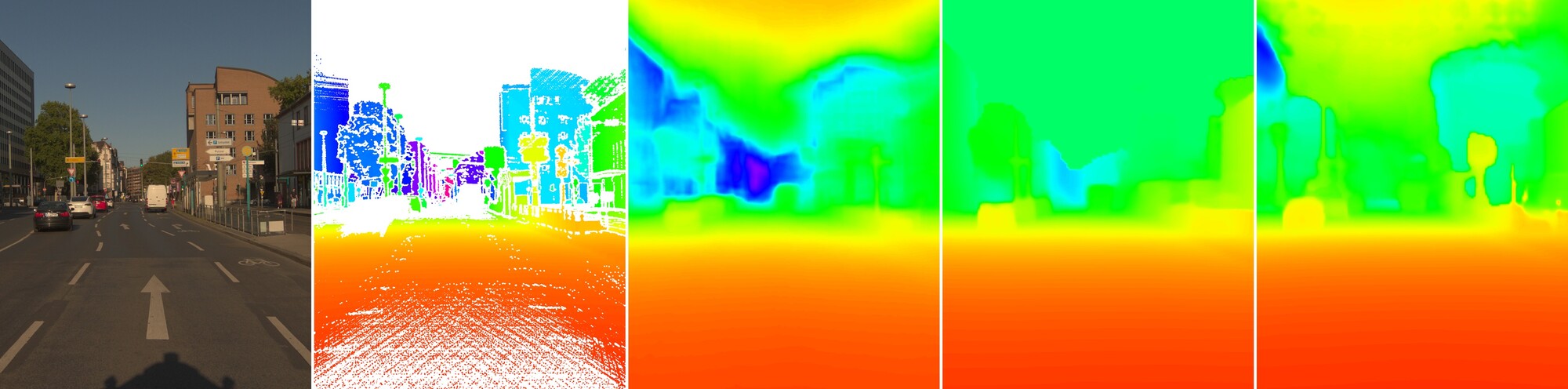}
    \includegraphics[width=\textwidth]{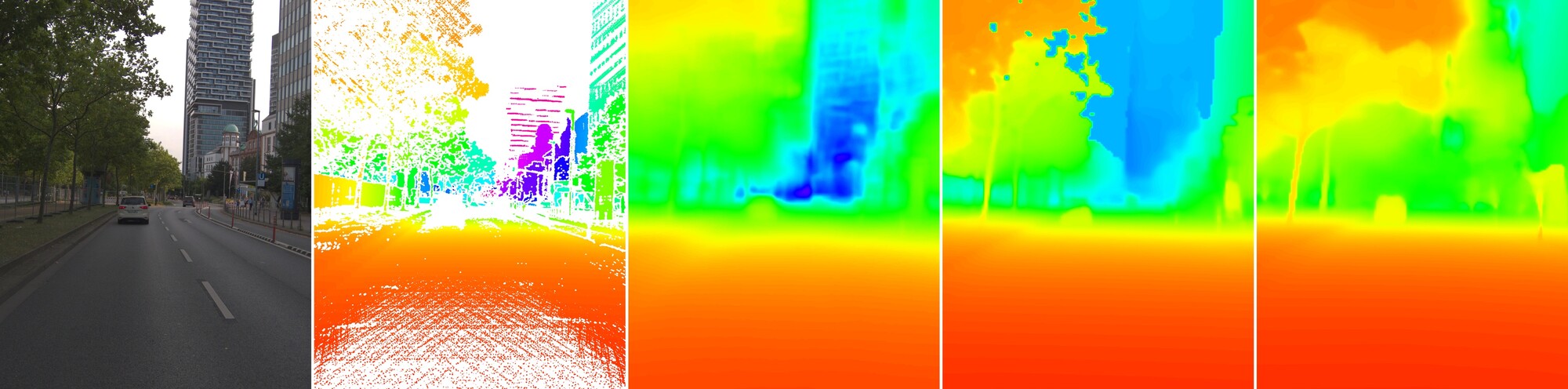}

    \newcolumntype{C}[1]{>{\centering\arraybackslash}p{#1}}
    {\small
    \begin{tabular}{C{0.17\textwidth}C{0.17\textwidth}C{0.17\textwidth}C{0.17\textwidth}C{0.17\textwidth}}
        RGB & KITScenes Lidar & UniDAC & Depth Anything 3 & MapAnything
    \end{tabular}
    }
    \vspace{-8pt}
    \caption{Qualitative comparison of monocular depth estimation methods. The corresponding non-linear depth scale is introduced in \Cref{fig:depth_distribution}.}
    \label{fig:qualitative_depth2}
\end{figure}

\begin{table}[h]
  \centering
  \caption{Stereo depth estimation - 0--100\,m}
  \label{tab:stereo-allmetrics-full-0-100m}
  \setlength{\tabcolsep}{6pt}
  \renewcommand{\arraystretch}{0.85}
  \begin{tabular}{@{}l c c c c c c@{}}
    \toprule
    Method & EPE\,[px] $\downarrow$ & BP-1\,[\%] $\downarrow$ & BP-2\,[\%] $\downarrow$ & D1\,[\%] $\downarrow$ & Abs.Rel $\downarrow$ & $\delta_{1}$ $\uparrow$ \\
    \midrule
  SGBM (WLS)~\cite{hirschmuller2008stereo} & 27.582 & 72.80 & 59.51 & 40.85 & 0.243 & 79.24 \\
  DEFOM-Stereo~\cite{jiang2025defom} & 4.790 & 44.74 & 25.42 & 12.41 & 0.155 & 93.11 \\
  FoundationStereo~\cite{wen2025foundationstereo} & \textbf{2.199} & \textbf{38.09} & \textbf{18.59} & \textbf{6.52} & \textbf{0.131} & \textbf{96.02} \\
    \bottomrule
  \end{tabular}

  \vspace{1em}
  \centering
  \caption{Stereo depth estimation - 100--200\,m}
  \label{tab:stereo-allmetrics-full-100-200m}
  \setlength{\tabcolsep}{6pt}
  \renewcommand{\arraystretch}{0.85}
  \begin{tabular}{@{}l c c c c c c@{}}
    \toprule
    Method & EPE\,[px] $\downarrow$ & BP-1\,[\%] $\downarrow$ & BP-2\,[\%] $\downarrow$ & D1\,[\%] $\downarrow$ & Abs.Rel $\downarrow$ & $\delta_{1}$ $\uparrow$ \\
    \midrule
  SGBM (WLS)~\cite{hirschmuller2008stereo} & 18.008 & 42.24 & 31.04 & 25.41 & 0.266 & 65.24 \\
  DEFOM-Stereo~\cite{jiang2025defom} & 2.805 & 28.29 & 19.99 & 16.96 & 0.192 & 76.77 \\
  FoundationStereo~\cite{wen2025foundationstereo} & \textbf{1.555} & \textbf{20.93} & \textbf{14.38} & \textbf{11.77} & \textbf{0.166} & \textbf{83.14} \\
    \bottomrule
  \end{tabular}

  \vspace{1em}
  \centering
  \caption{Stereo depth estimation - $>$200\,m}
  \label{tab:stereo-allmetrics-full-200mplus}
  \setlength{\tabcolsep}{6pt}
  \renewcommand{\arraystretch}{0.85}
  \begin{tabular}{@{}l c c c c c c@{}}
    \toprule
    Method & EPE\,[px] $\downarrow$ & BP-1\,[\%] $\downarrow$ & BP-2\,[\%] $\downarrow$ & D1\,[\%] $\downarrow$ & Abs.Rel $\downarrow$ & $\delta_{1}$ $\uparrow$ \\
    \midrule
  SGBM (WLS)~\cite{hirschmuller2008stereo} & 12.784 & 35.36 & 24.14 & 18.62 & 0.276 & 58.42 \\
  DEFOM-Stereo~\cite{jiang2025defom} & 1.875 & 21.97 & 15.54 & 12.54 & 0.206 & 70.57 \\
  FoundationStereo~\cite{wen2025foundationstereo} & \textbf{1.182} & \textbf{16.55} & \textbf{10.07} & \textbf{7.41} & \textbf{0.179} & \textbf{75.76} \\
    \bottomrule
  \end{tabular}

  \vspace{1em}
  \centering
  \caption{Stereo depth estimation - Overall}
  \label{tab:stereo-allmetrics-full-overall}
  \setlength{\tabcolsep}{6pt}
  \renewcommand{\arraystretch}{0.85}
  \begin{tabular}{@{}l c c c c c c@{}}
    \toprule
    Method & EPE\,[px] $\downarrow$ & BP-1\,[\%] $\downarrow$ & BP-2\,[\%] $\downarrow$ & D1\,[\%] $\downarrow$ & Abs.Rel $\downarrow$ & $\delta_{1}$ $\uparrow$ \\
    \midrule
  SGBM (WLS)~\cite{hirschmuller2008stereo} & 27.416 & 72.28 & 59.03 & 40.58 & 0.243 & 79.00 \\
  DEFOM-Stereo~\cite{jiang2025defom} & 4.756 & 44.46 & 25.32 & 12.48 & 0.156 & 92.83 \\
  FoundationStereo~\cite{wen2025foundationstereo} & \textbf{2.188} & \textbf{37.81} & \textbf{18.51} & \textbf{6.60} & \textbf{0.132} & \textbf{95.80} \\
    \bottomrule
  \end{tabular}
\end{table}

\clearpage

\begin{figure}[H]
    \centering
    \includegraphics[width=0.49\textwidth]{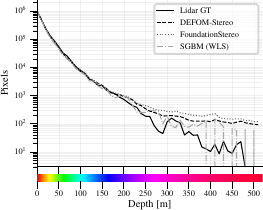}
    \caption{Depth distributions of pixels with valid lidar depth for the stereo methods.}
    \label{fig:stereo_depth_distribution}
\end{figure}

\subsection{Novel View Synthesis}
\label{sec:appx:benchmark_details:nvs}

\paragraph{Setup.}
We evaluate the \texttt{val} and \texttt{overlap-train-val} scenarios.
Each clip spans 41 consecutive frames centered at frame $49+100k$ of a scenario, yielding 543 clips from 342 scenarios, and is scored at eight target frames spaced \SI{0.5}{\second} apart.
Lateral views translate the front ring camera along the ego $y$ axis by $\Delta y\in\{\pm1,\pm2,\pm3\}$\,m, keeping its orientation and intrinsics.
Ground-truth traffic signs are projected from the scenario's Lanelet2 HD map into each shifted viewpoint.
To ensure a fair comparison, we apply lidar-based occlusion filtering: a sign is considered visible only if the density of lidar points (laterally shifted by the same offset) does not indicate a foreground occlusion.
Detections are obtained with OWLv2~\cite{minderer2023scaling} at a confidence threshold of 0.15 and matched against the projected GT bounding boxes using an IoU threshold of 0.5.

\paragraph{Evaluation Images and Model Inputs.}
All views are evaluated at $280{\times}498$. Target views and model inputs are cropped to exclude the ego hood, with intrinsics adjusted accordingly; inputs for the held-out camera use a taller crop to cover its field of view.
ReconDrive is a feed-forward 3D Gaussian Splatting method, whereas AnyView is a video diffusion model and alone observes the target instant.
For each target frame $t$, ReconDrive reconstructs the scene from the six ring cameras at $t{-}2$ and $t{+}3$ at $280{\times}518$.
AnyView generates the complete target video from all 41 frames of the front ring camera at $304{\times}576$, using 35 sampling steps and its translation scale for driving datasets ($1/16$).
Our benchmark therefore compares NVS methods across paradigms under a common evaluation protocol, rather than under equal inputs.

\paragraph{Metrics.}
We report traffic sign recall, defined as the ratio of detected to visible GT signs, with empty frames contributing zero counts.
Scored signs must lie fully inside the evaluation image and have a shortest side of at least \SI{8}{\pixel}.
This standard keeps 8046 of the 72737 projected real-photograph signs and 5734 of the 11216 detected ones, so it measures large, fully visible signs rather than distant ones.

\begin{figure*}[!thp]
  \centering
  \begin{subfigure}{0.24\textwidth}
    \includegraphics[width=\textwidth]{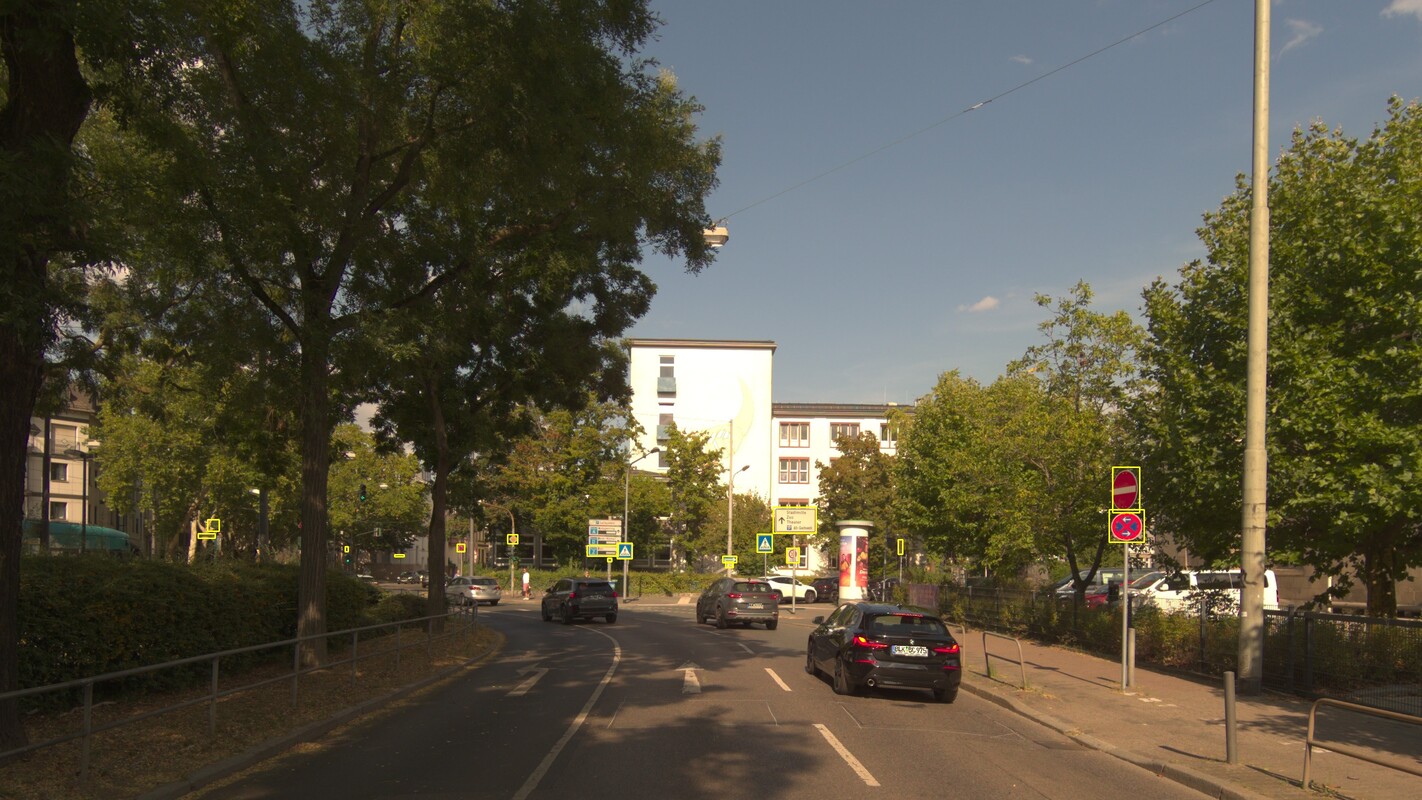}
    \caption{Photo}
  \end{subfigure}
  \hfill
  \begin{subfigure}{0.24\textwidth}
    \includegraphics[width=\textwidth]{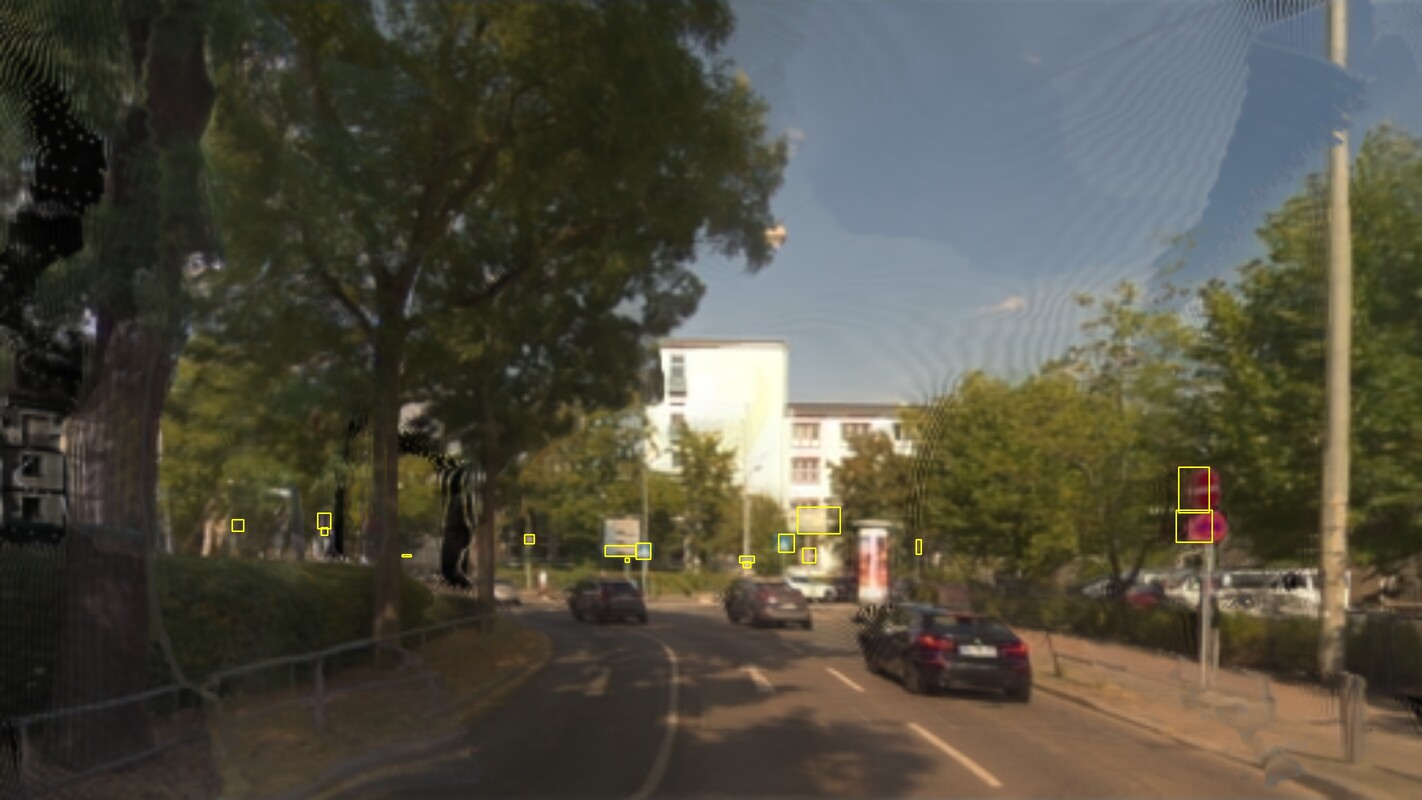}
    \caption{Left \SI{1}{\meter}}
  \end{subfigure}
  \hfill
  \begin{subfigure}{0.24\textwidth}
    \includegraphics[width=\textwidth]{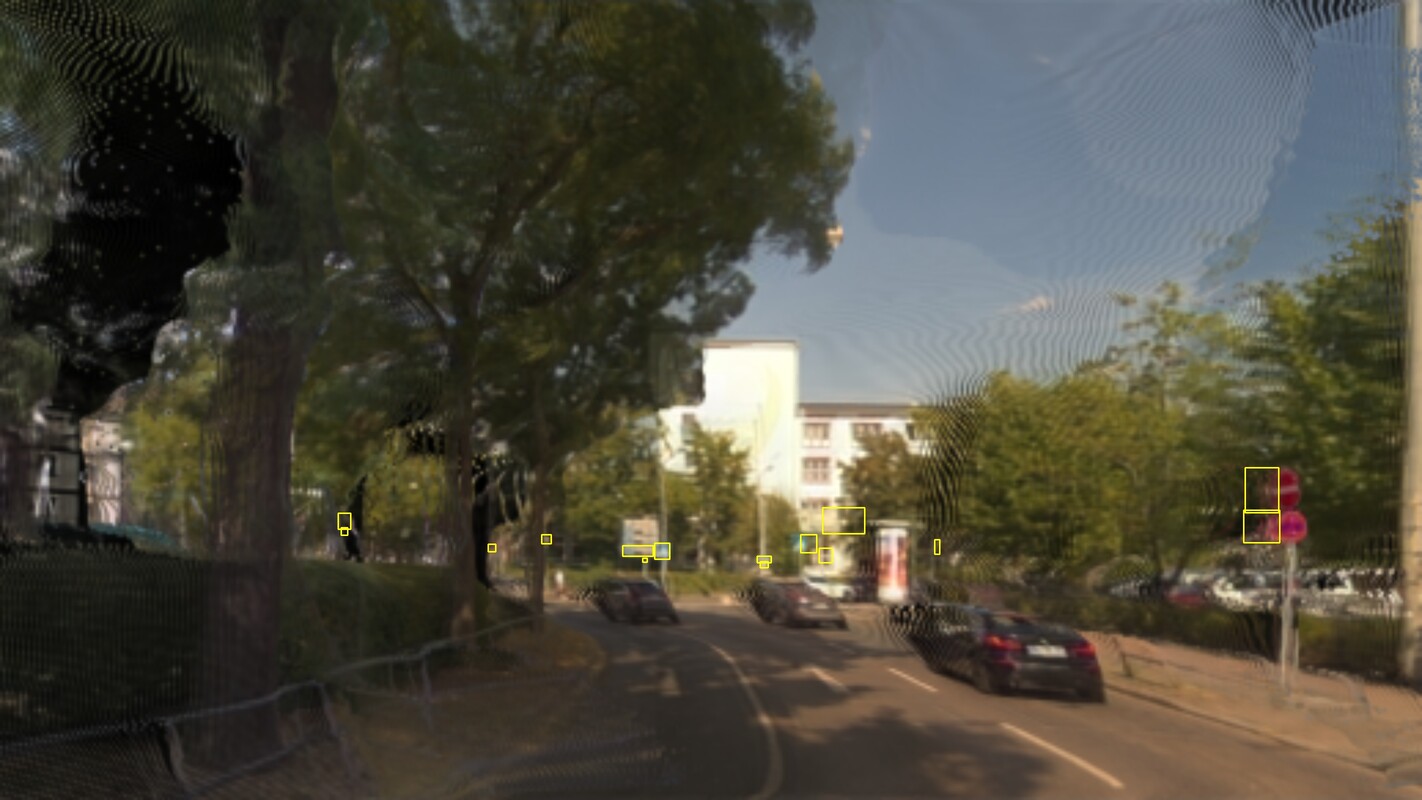}
    \caption{Left \SI{2}{\meter}}
  \end{subfigure}
  \hfill
  \begin{subfigure}{0.24\textwidth}
    \includegraphics[width=\textwidth]{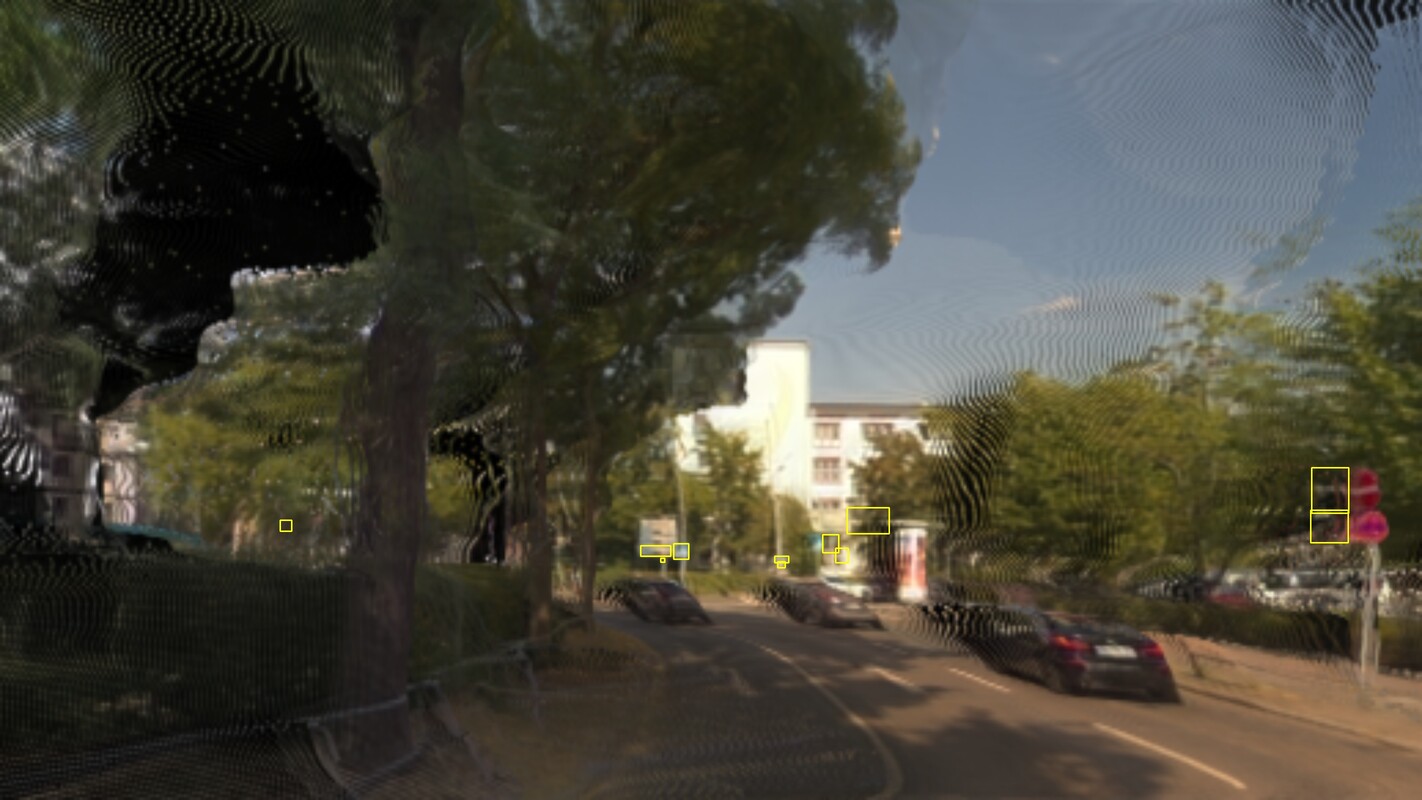}
    \caption{Left \SI{3}{\meter}}
  \end{subfigure}

  \vspace{4pt}

  \begin{subfigure}{0.24\textwidth}
    \includegraphics[width=\textwidth]{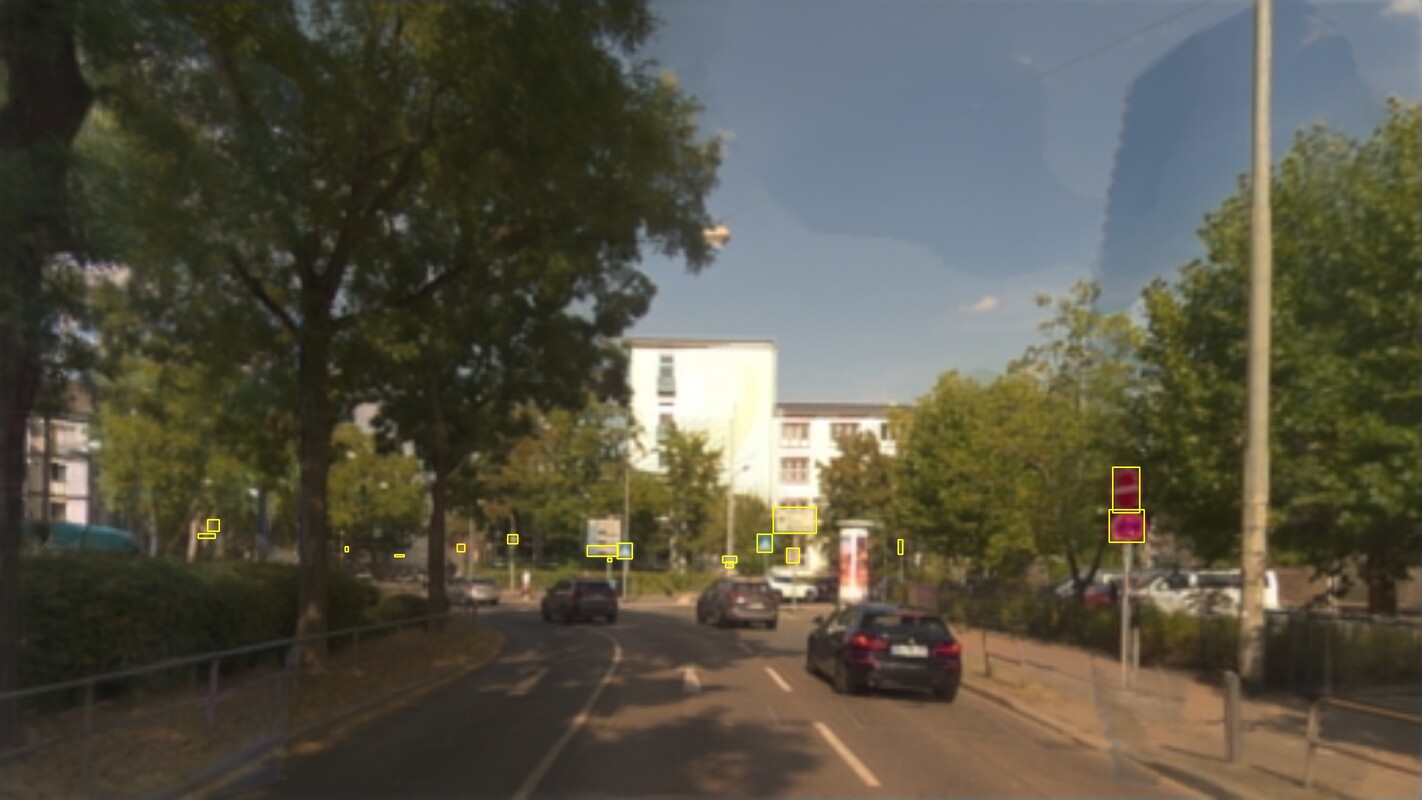}
    \caption{Reconstruction}
    \label{fig:qualitative_lateral_nvs_recon1}
  \end{subfigure}
  \hfill
  \begin{subfigure}{0.24\textwidth}
    \includegraphics[width=\textwidth]{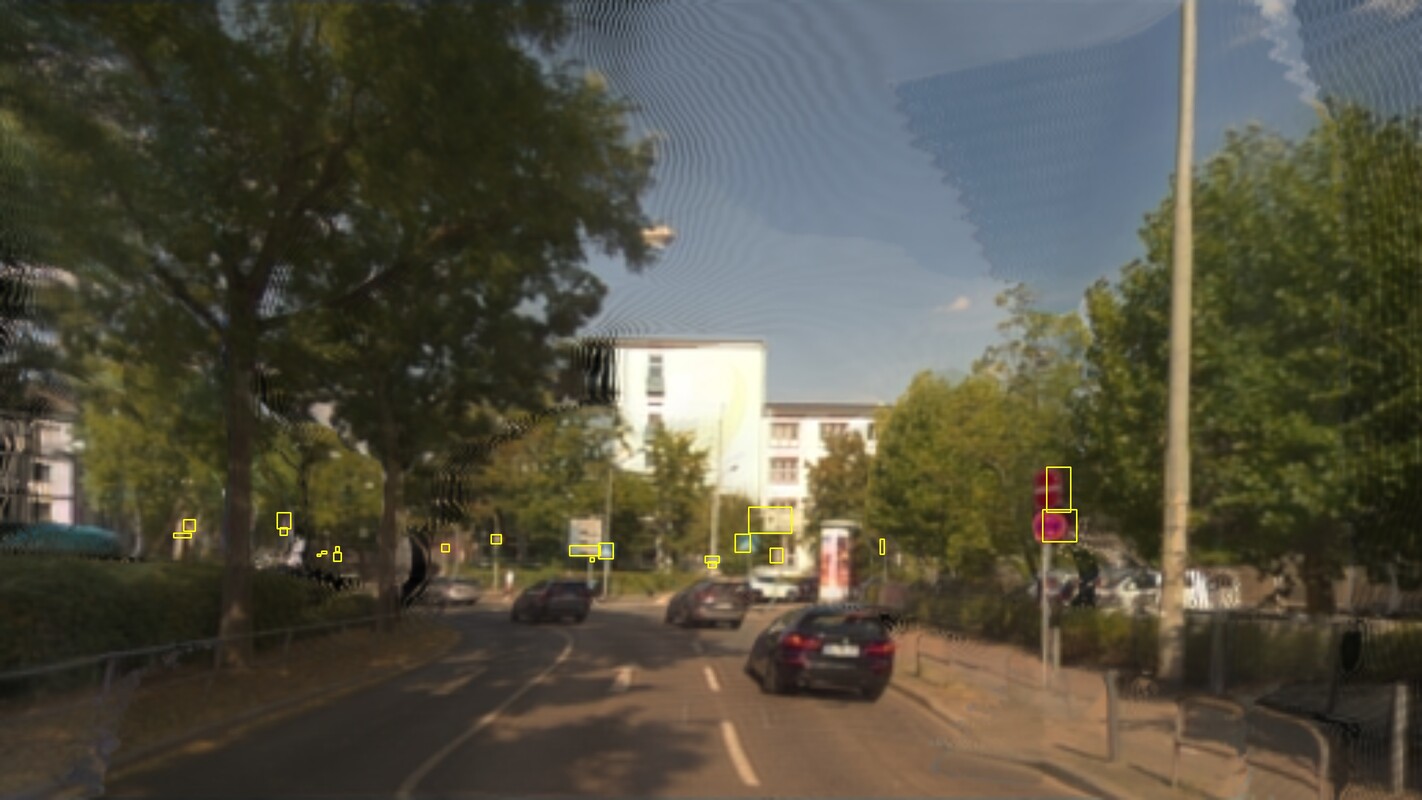}
    \caption{Right \SI{1}{\meter}}
  \end{subfigure}
  \hfill
  \begin{subfigure}{0.24\textwidth}
    \includegraphics[width=\textwidth]{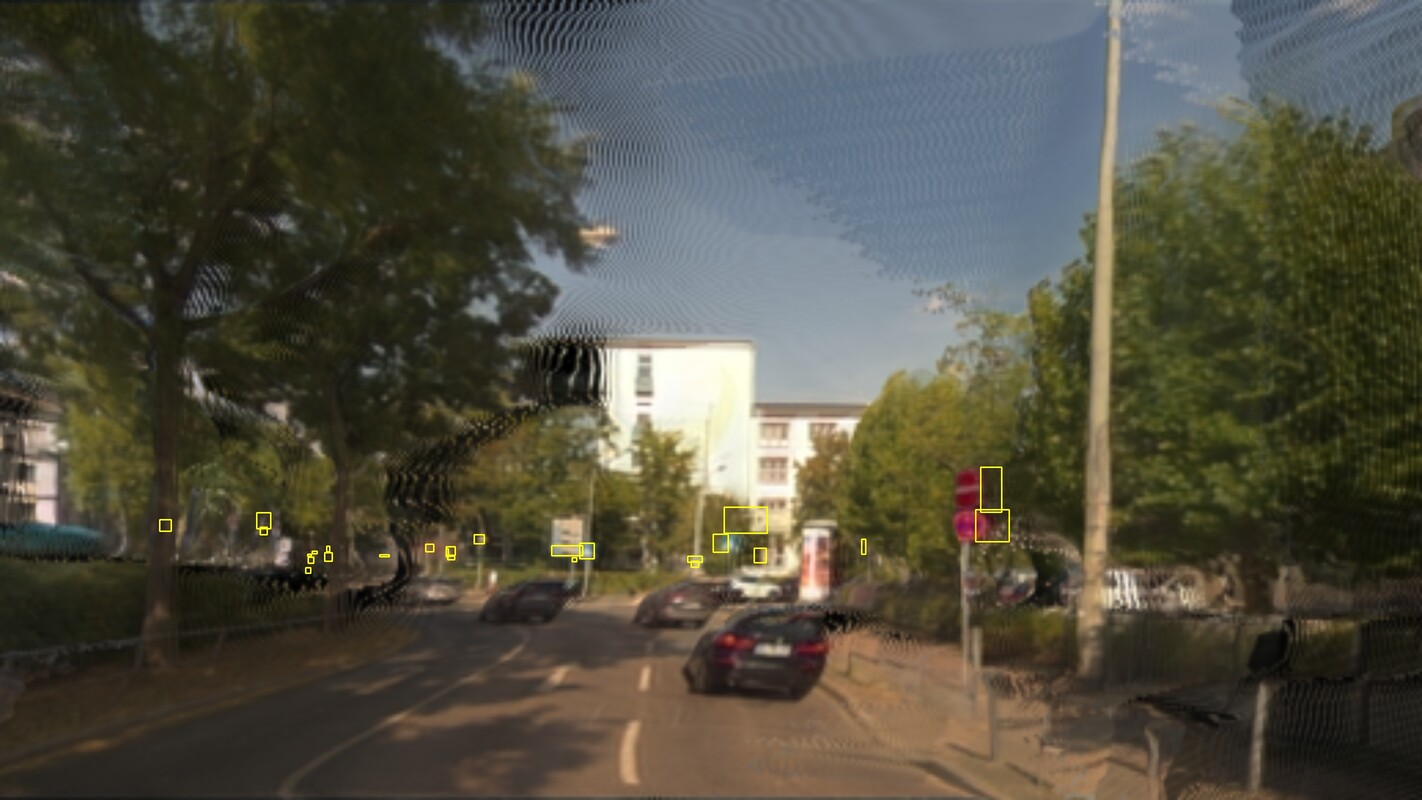}
    \caption{Right \SI{2}{\meter}}
  \end{subfigure}
  \hfill
  \begin{subfigure}{0.24\textwidth}
    \includegraphics[width=\textwidth]{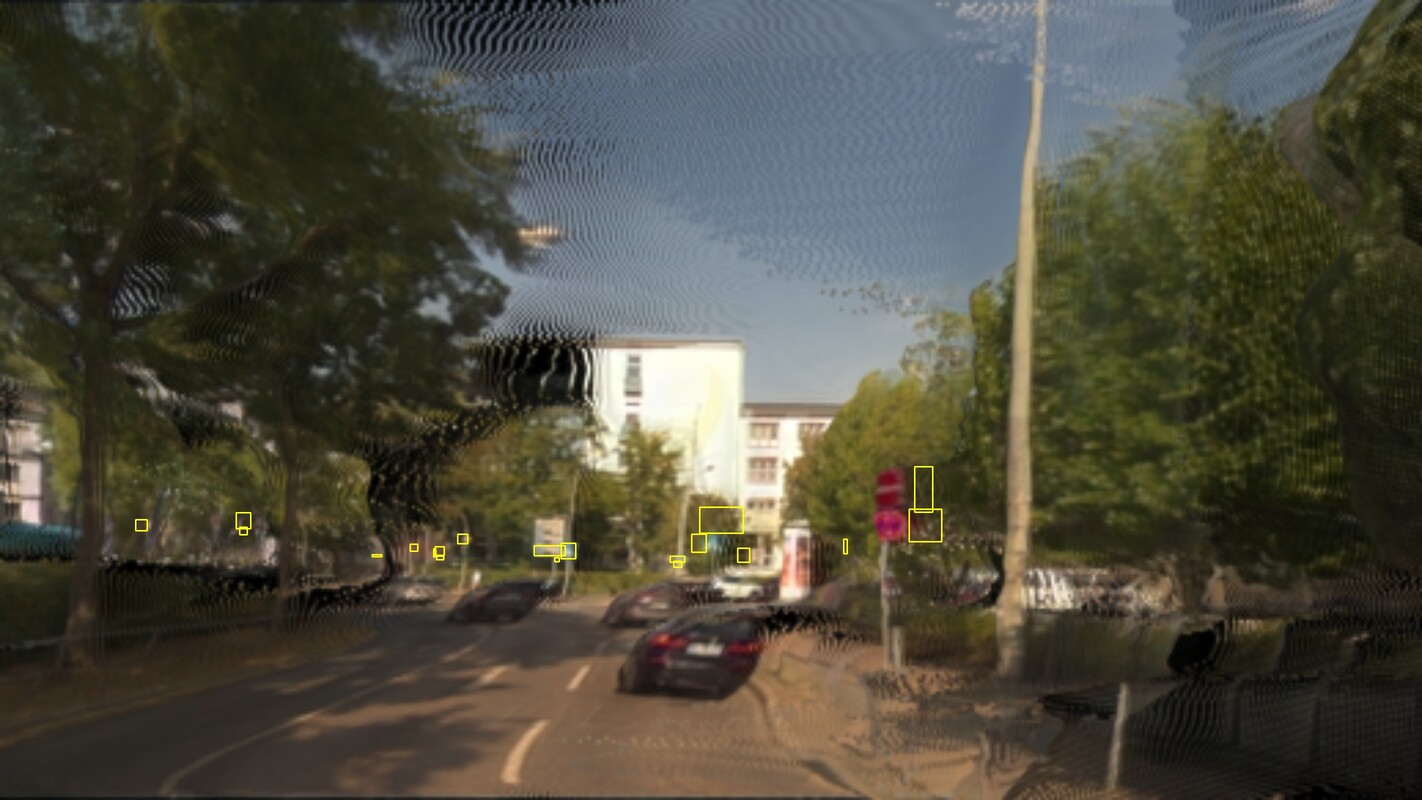}
    \caption{Right \SI{3}{\meter}}
  \end{subfigure}
  \vspace{4pt}
  \noindent\rule{\linewidth}{0.4pt}
  \vspace{4pt}
  \begin{subfigure}{0.24\textwidth}
    \includegraphics[width=\textwidth]{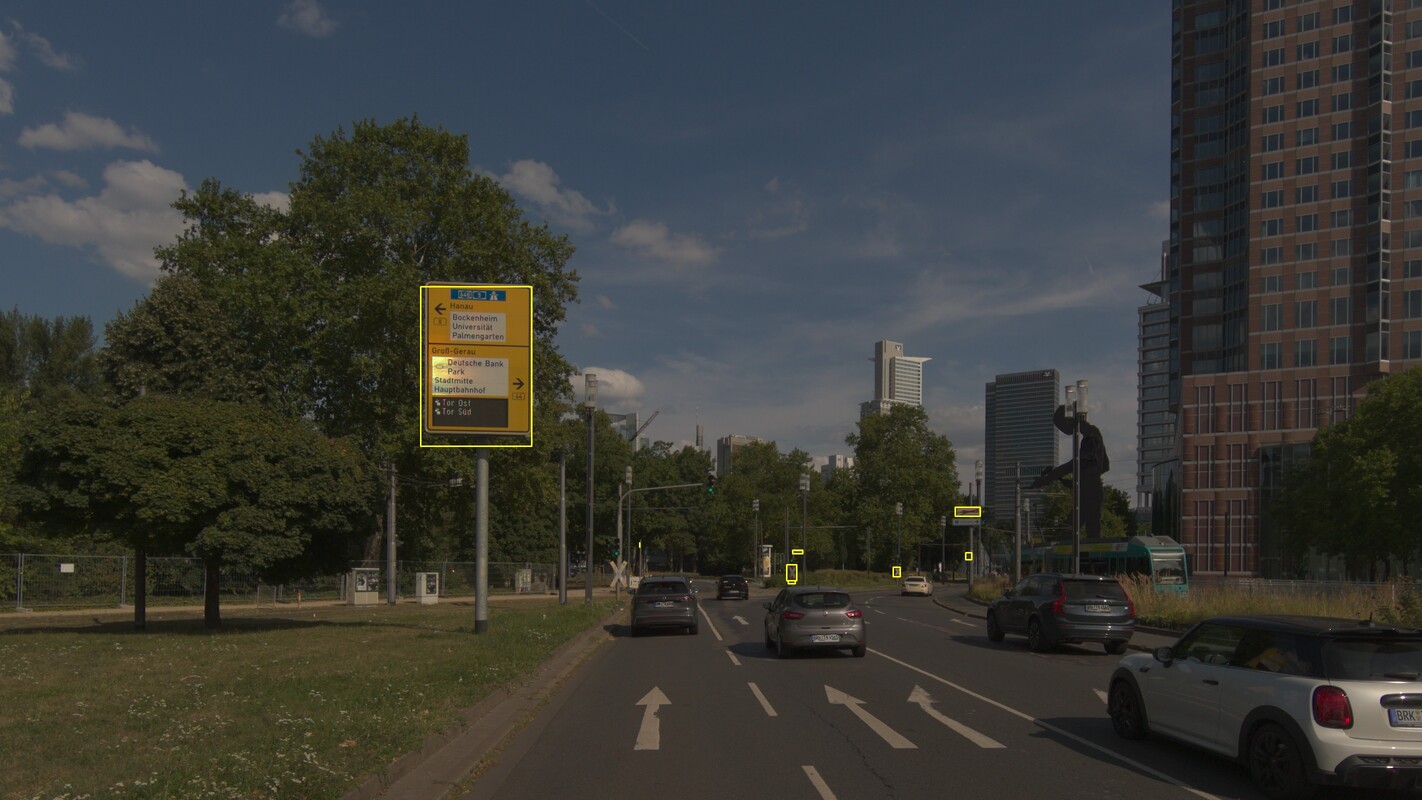}
    \caption{Photo}
  \end{subfigure}
  \hfill
  \begin{subfigure}{0.24\textwidth}
    \includegraphics[width=\textwidth]{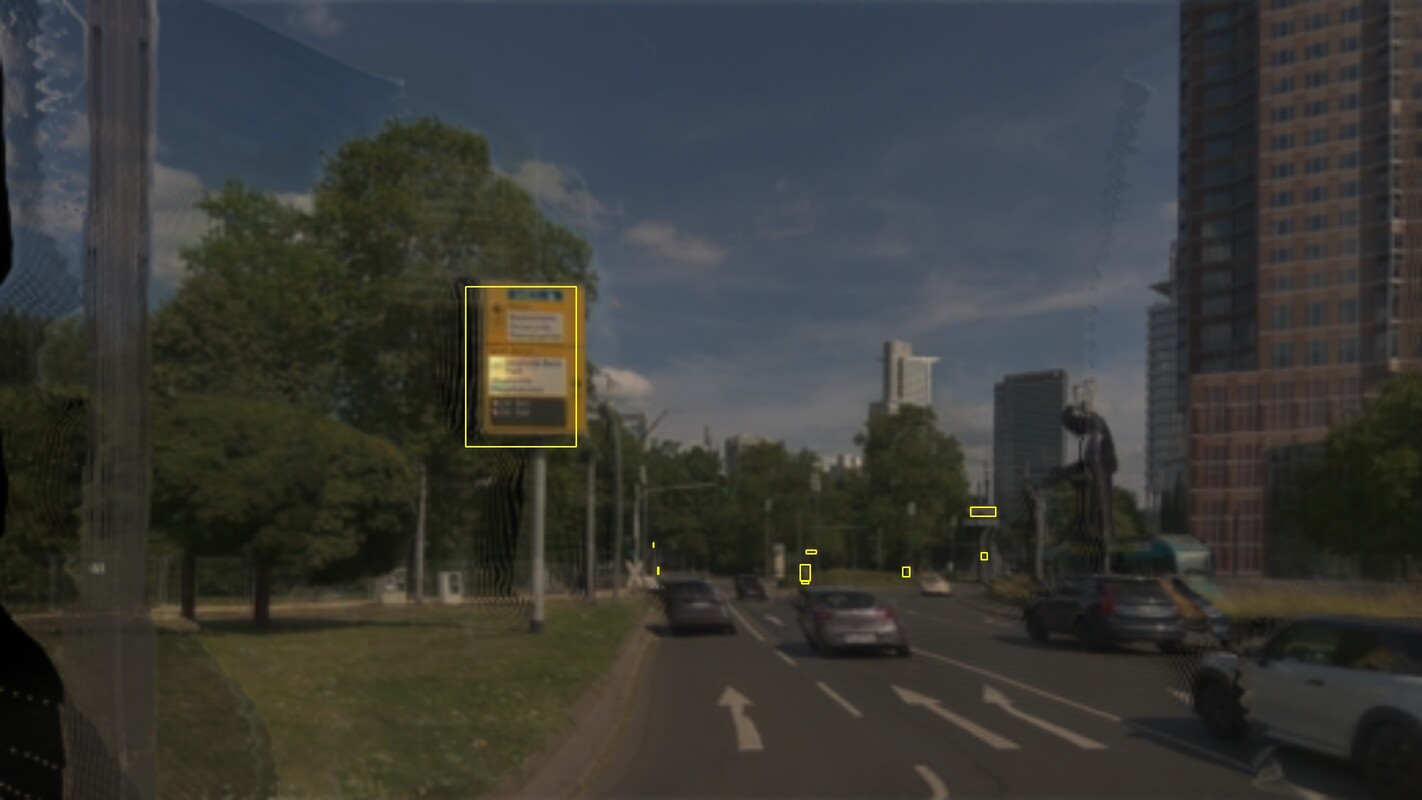}
    \caption{Left \SI{1}{\meter}}
  \end{subfigure}
  \hfill
  \begin{subfigure}{0.24\textwidth}
    \includegraphics[width=\textwidth]{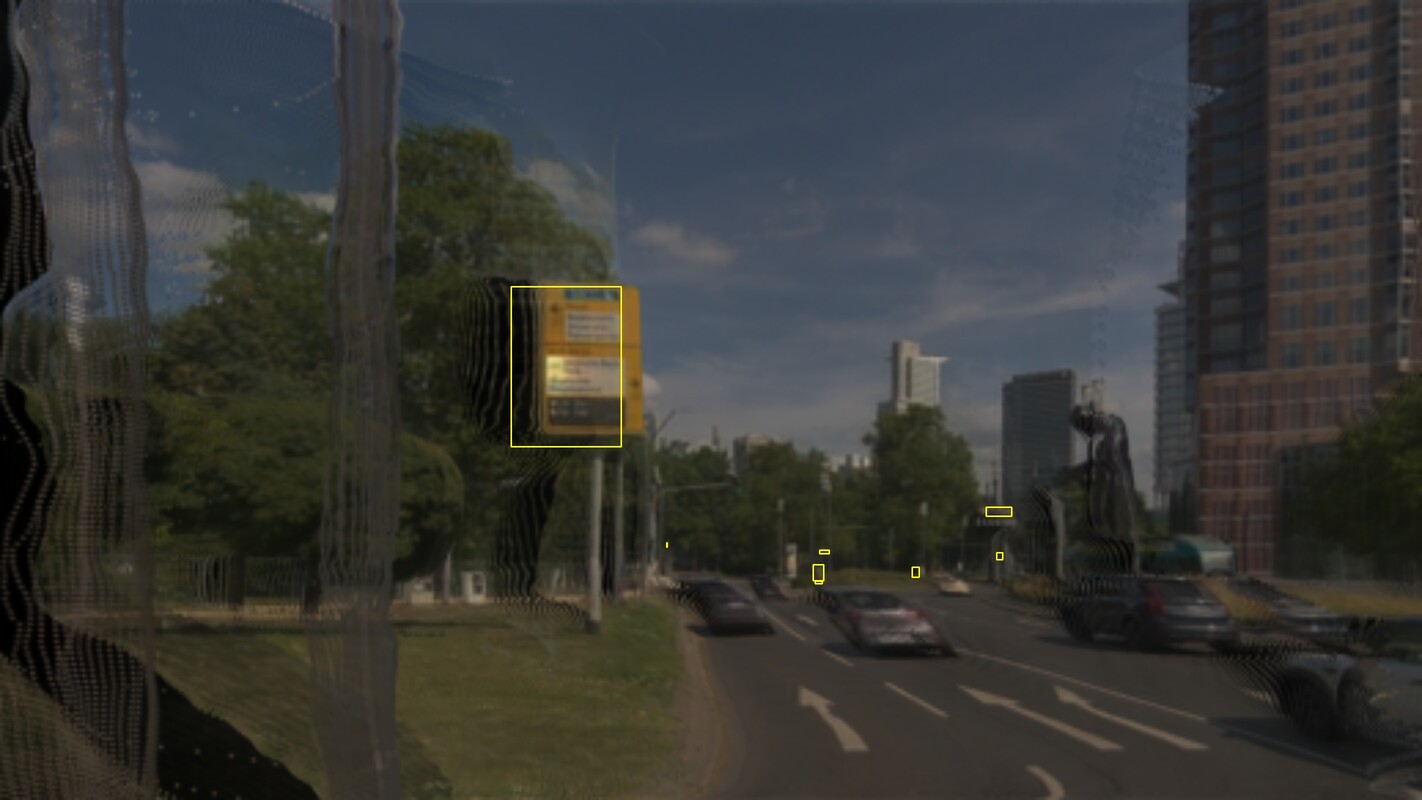}
    \caption{Left \SI{2}{\meter}}
  \end{subfigure}
  \hfill
  \begin{subfigure}{0.24\textwidth}
    \includegraphics[width=\textwidth]{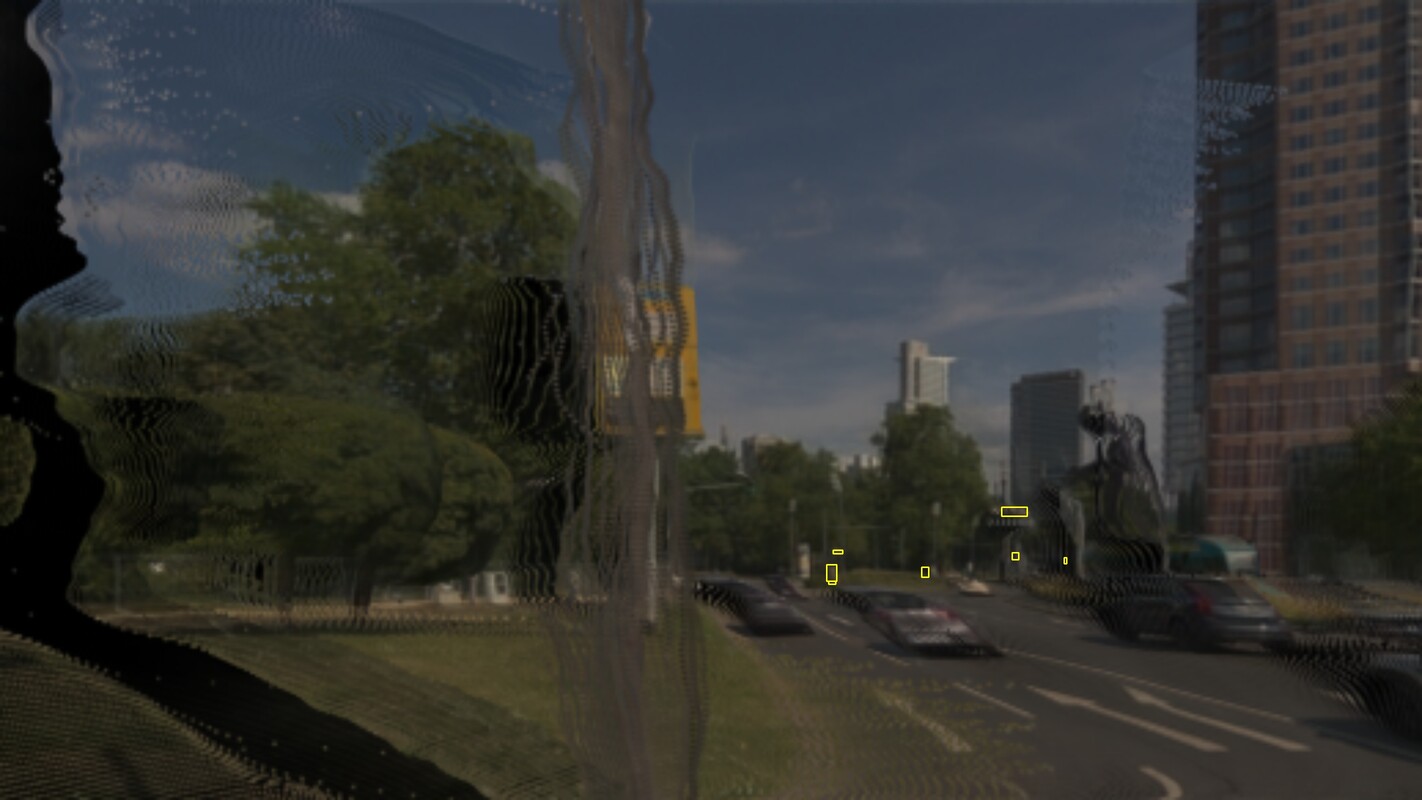}
    \caption{Left \SI{3}{\meter}}
  \end{subfigure}

  \vspace{4pt}

  \begin{subfigure}{0.24\textwidth}
    \includegraphics[width=\textwidth]{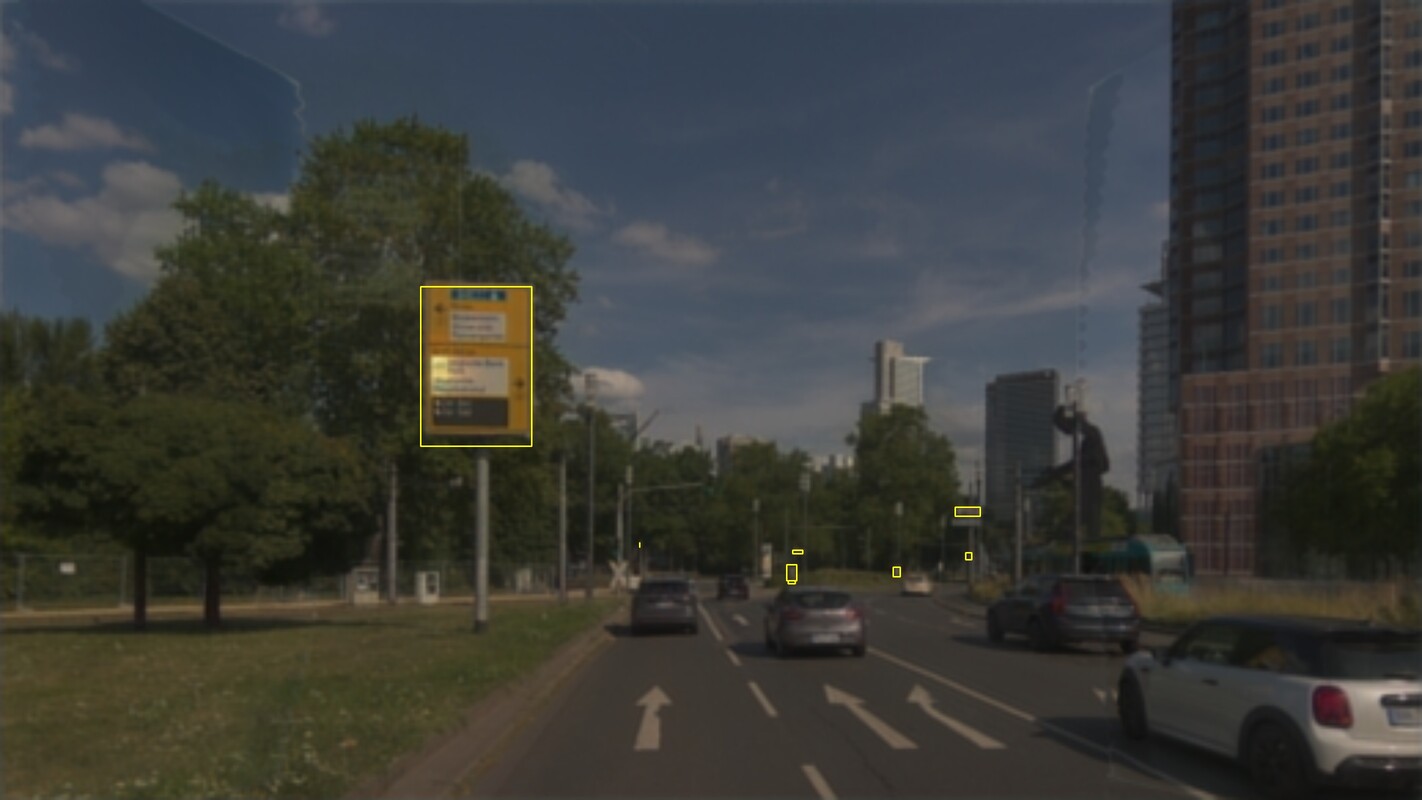}
    \caption{Reconstruction}
    \label{fig:qualitative_lateral_nvs_recon2}
  \end{subfigure}
  \hfill
  \begin{subfigure}{0.24\textwidth}
    \includegraphics[width=\textwidth]{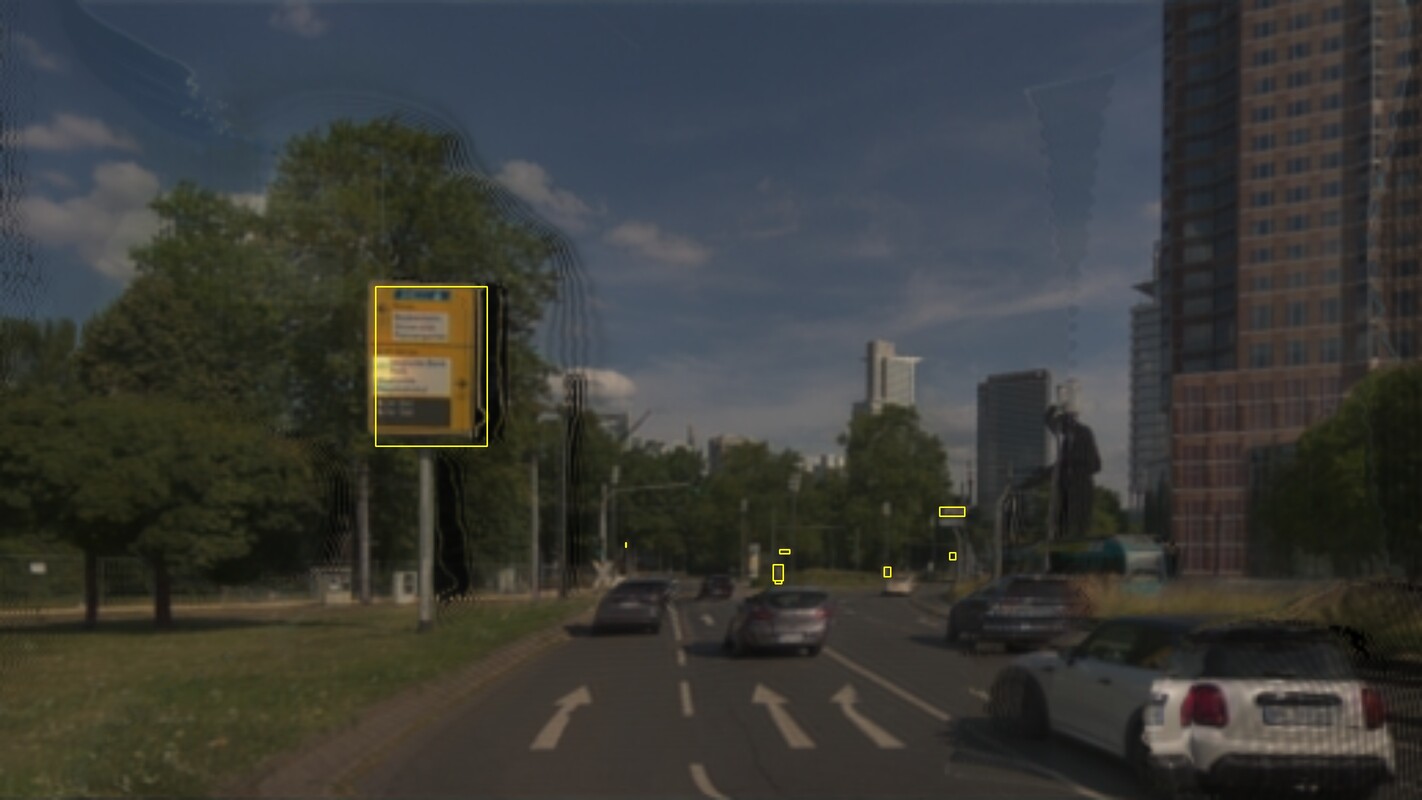}
    \caption{Right \SI{1}{\meter}}
  \end{subfigure}
  \hfill
  \begin{subfigure}{0.24\textwidth}
    \includegraphics[width=\textwidth]{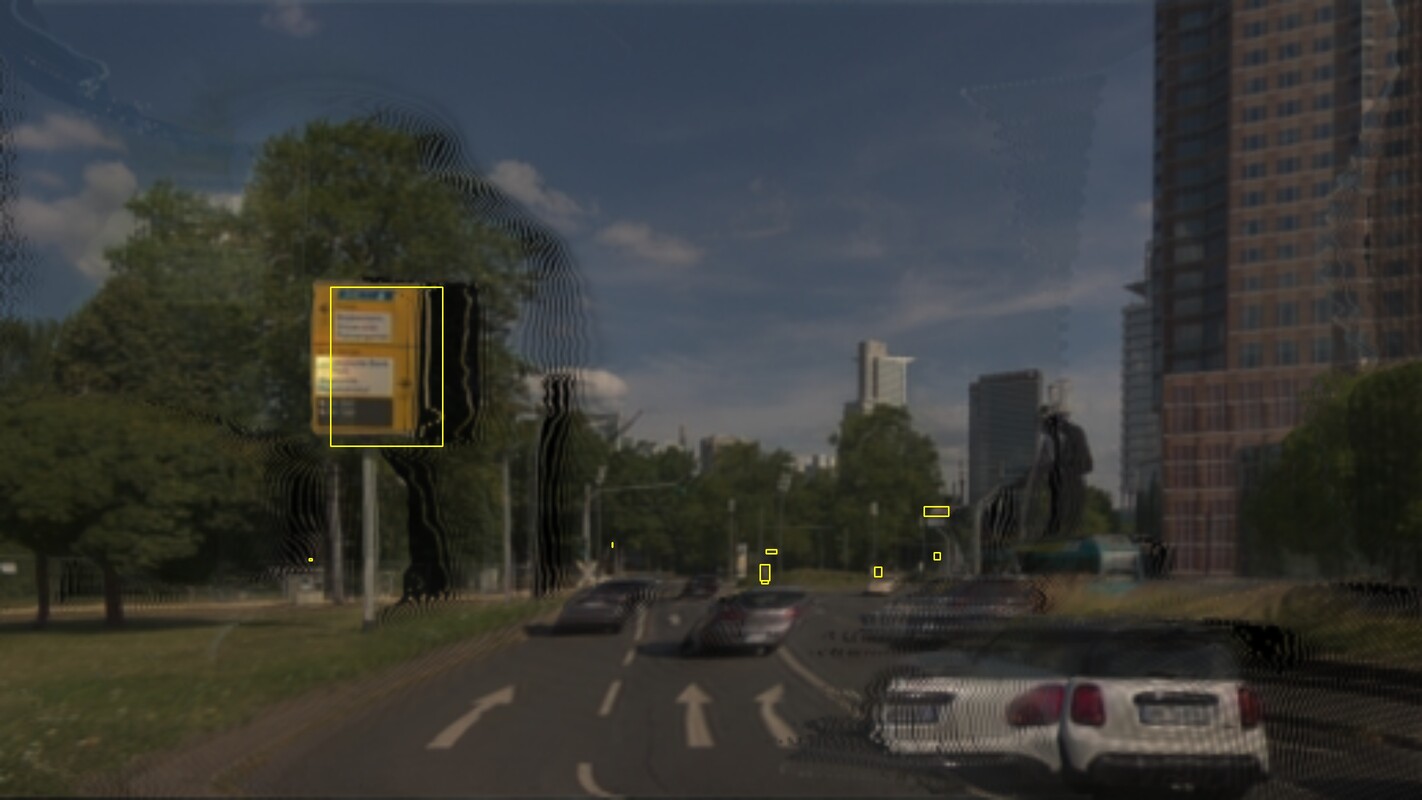}
    \caption{Right \SI{2}{\meter}}
  \end{subfigure}
  \hfill
  \begin{subfigure}{0.24\textwidth}
    \includegraphics[width=\textwidth]{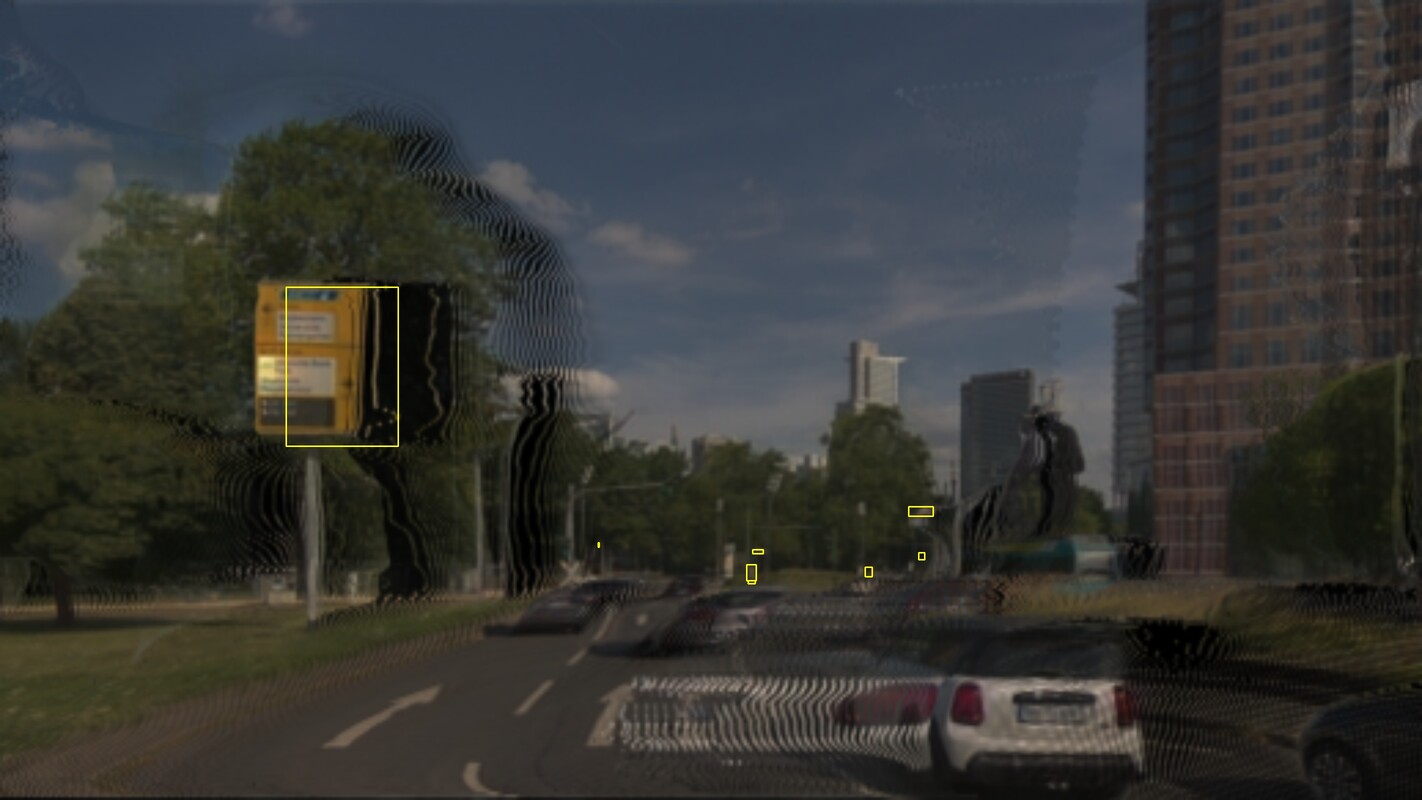}
    \caption{Right \SI{3}{\meter}}
  \end{subfigure}
  \caption{%
    Qualitative comparison of traffic-sign recall under lateral viewpoint shifts.
    (a), (i) Real photograph with projected GT annotations (yellow).
    (b--h), (j--p) Lateral NVS rendered by ReconDrive~\cite{yu2026_recondrive} with detections.
    Note the deviation between rendered traffic signs and GT projections in the lateral views.
  }
  \label{fig:qualitative_lateral_nvs}
\end{figure*}

\paragraph{Qualitative Lateral NVS Results.}
We further showcase qualitative results on the map-based NVS evaluation benchmark proposed in \Cref{sec:benchmarks:nvs}.
\Cref{fig:qualitative_lateral_nvs} illustrates the impact of lateral translation on structural fidelity.
We observe that while the reconstruction at the driven trajectory (\Cref{fig:qualitative_lateral_nvs_recon1}, \Cref{fig:qualitative_lateral_nvs_recon2}) shows reasonable alignment with the HD map projections, lateral shifts reveal significant geometric inaccuracies. 
Specifically, the traffic signs rendered in the novel views synthesized by ReconDrive do not fully align with the ground-truth projections. 
This misalignment, which becomes more pronounced as the lateral offset increases, suggests that the underlying geometry lacks the precision required for consistent projection at novel viewpoints. 
These artifacts degrade the object's visual signature, causing the detector to miss signs that are clearly visible in the ground-truth photograph. 
This highlights that photometric consistency on the training distribution does not guarantee geometric accuracy in novel spatial views, a gap that is critical for safety-oriented simulation.

\paragraph{Photometric NVS Benchmarks.}
We complement the map-based evaluation with a photometric held-out benchmark that measures complementary aspects of reconstruction quality.
It targets the independently mounted \SI{16.2}{\mega\pixel} \texttt{camera\_base\_front\_center}, which provides the novel-view target at the same instant. 
This benchmark measures spatial extrapolation to a viewpoint the method never observes as input.
We report PSNR, SSIM, and LPIPS at $280{\times}498$, averaged over the eight targets of each clip and then over clips, in \Cref{tab:nvs_results}.

\begin{table}[h]
  \centering
  \small
  \caption{%
    Held-out cross-mount NVS on the front-center camera over 543 clips, evaluated at $280{\times}498$.
    Parentheses give 95\% scenario-bootstrap confidence intervals, and bold marks the better method.
    $\uparrow$ and $\downarrow$ denote that higher and lower are better, respectively.
  }
  \label{tab:nvs_results}
  \setlength{\tabcolsep}{4pt}
  \begin{tabular}{l ccc}
    \toprule
    & PSNR$\uparrow$ & SSIM$\uparrow$ & LPIPS$\downarrow$ \\
    \midrule
    ReconDrive~\cite{yu2026_recondrive}
      & \textbf{23.12} \scriptsize{(22.90--23.33)} & \textbf{0.719} \scriptsize{(0.712--0.725)} & \textbf{0.384} \scriptsize{(0.378--0.390)} \\
    AnyView~\cite{anyview}
      & 19.62 \scriptsize{(19.30--19.96)} & 0.571 \scriptsize{(0.558--0.582)} & 0.416 \scriptsize{(0.410--0.423)} \\
    \bottomrule
  \end{tabular}
\end{table}

\FloatBarrier

\subsection{End-to-End Driving}
\label{sec:appx:benchmark_details:e2e}

\paragraph{Evaluation Sample Construction.}
The 483 e2e samples are drawn from the \texttt{val}~$\cup$~\texttt{overlap-train-val} scenes; each is a non-overlapping \SI{9}{\second} window of \SI{4}{\second} past observation and up to \SI{5}{\second} of future trajectory anchored at a keyframe.
Headline numbers use the \SI{3}{\second} horizon for parity with nuScenes- and nuPlan-trained baselines; the full \SI{5}{\second} protocol is offered as the long-horizon challenge. The samples stem from clips of 10 to \SI{60}{\second} recorded at \SI{10}{\hertz}, see \Cref{tab:splits_summary}. The held-out \texttt{test-e2e} split contains 107 scenes and \SI{24.5}{\kilo\meter} (112 scenes and \SI{25.2}{\kilo\meter} with \texttt{map-change-test-e2e}) with future sensor data and global pose withheld after the keyframe, and is reserved for a future leaderboard. 

\begin{table*}[t]
\centering
\caption{Inputs of the evaluated end-to-end models as run on KITScenes Multimodal. Resolution is per camera after resampling into the training camera (for the NAVSIM planners: of the panorama stitched from the three cropped views); megapixels (MP) are per time step over all cameras. \emph{Ego state} lists ego-motion information reaching the model; \emph{CAN bus} denotes velocity, acceleration, and yaw rate from the vehicle bus. \emph{CAN bus} denotes the nuScenes CAN vector (velocity, acceleration, yaw rate, heading), here derived from the recorded past poses; the NAVSIM planners receive velocity and acceleration directly as planner input.}
\label{tab:app:e2e_inputs}
\scriptsize
\setlength{\tabcolsep}{3.5pt}
\renewcommand{\arraystretch}{1.12}
\resizebox{\textwidth}{!}{%
\begin{tabular}{l l c c c c l l l c}
\toprule
\multirow{2}{*}{Model} & \multirow{2}{*}{\makecell[l]{Training\\data}}
& \multicolumn{3}{c}{Cameras}
& \multirow{2}{*}{\makecell[l]{Lidar\\BEV}}
& \multirow{2}{*}{Temporal context}
& \multirow{2}{*}{Ego state}
& \multirow{2}{*}{\makecell[l]{Driving\\command}}
& \multirow{2}{*}{Output}
\\
\cmidrule(lr){3-5}
& & \# & Res.\ [px] & Mpx & & & & &
\\
\midrule
UniAD~\cite{hu2023uniad}         & nuScenes & 6 (ring) & $1600\times900$ & 8.6 & -- & 2 past frames & CAN bus & 3-way & \SI{3}{\second} @ \SI{2}{\hertz} \\
DMAD~\cite{shen2025dmad}         & nuScenes & 6 (ring) & $1600\times900$ & 8.6 & -- & 2 past frames & CAN bus & 3-way & \SI{3}{\second} @ \SI{2}{\hertz} \\
SSR~\cite{li2025ssr} (non-temp.) & nuScenes & 6 (ring) & $640\times360$  & 1.4 & -- & --                         & CAN bus & 3-way & \SI{3}{\second} @ \SI{2}{\hertz} \\
SSR~\cite{li2025ssr} (temporal)  & nuScenes & 6 (ring) & $640\times360$  & 1.4 & -- & 2 past frames (\SI{1}{\second}) & CAN bus & 3-way & \SI{3}{\second} @ \SI{2}{\hertz} \\
\midrule
Epona~\cite{zhang2025epona} (AR / SS) & nuPlan & 1 (front) & $1024\times512$ & 0.5 & -- & 10 frames, 11 poses & past poses & -- & \SI{3}{\second} @ \SI{5}{\hertz} \\
\midrule
TransFuser~\cite{chitta2023transfuser}       & NAVSIM & 3 (stitched) & $1024\times256$ & 0.26 & $256^2$, \SI{64}{\meter} & -- & velocity, accel. & 4-way & \SI{4}{\second} @ \SI{2}{\hertz} \\
LTF~\cite{chitta2023transfuser}              & NAVSIM & 3 (stitched) & $1024\times256$ & 0.26 & --                          & -- & velocity, accel. & 4-way & \SI{4}{\second} @ \SI{2}{\hertz} \\
DiffusionDrive~\cite{liao2025diffusiondrive} & NAVSIM & 3 (stitched) & $1024\times256$ & 0.26 & $256^2$, \SI{64}{\meter} & -- & velocity, accel. & 4-way & \SI{4}{\second} @ \SI{2}{\hertz} \\
\bottomrule
\end{tabular}%
}
\vspace{-3pt}
{\parbox{1.0\linewidth}{
    \vspace{1pt}
    \scriptsize
    3-way: turn left / go straight / turn right; 4-way adds \emph{unknown}, which is never issued here. Commands are derived from the next \SI{3}{\second} for UniAD, DMAD, and SSR and from the next \SI{5}{\second} for the NAVSIM planners; in \Cref{tab:app:kitscenes_5s_results}, UniAD, DMAD, and SSR also receive the \SI{5}{\second} command.}
}
\end{table*}

\begin{figure}[!thp]
  \centering
    \includegraphics[width=\linewidth]{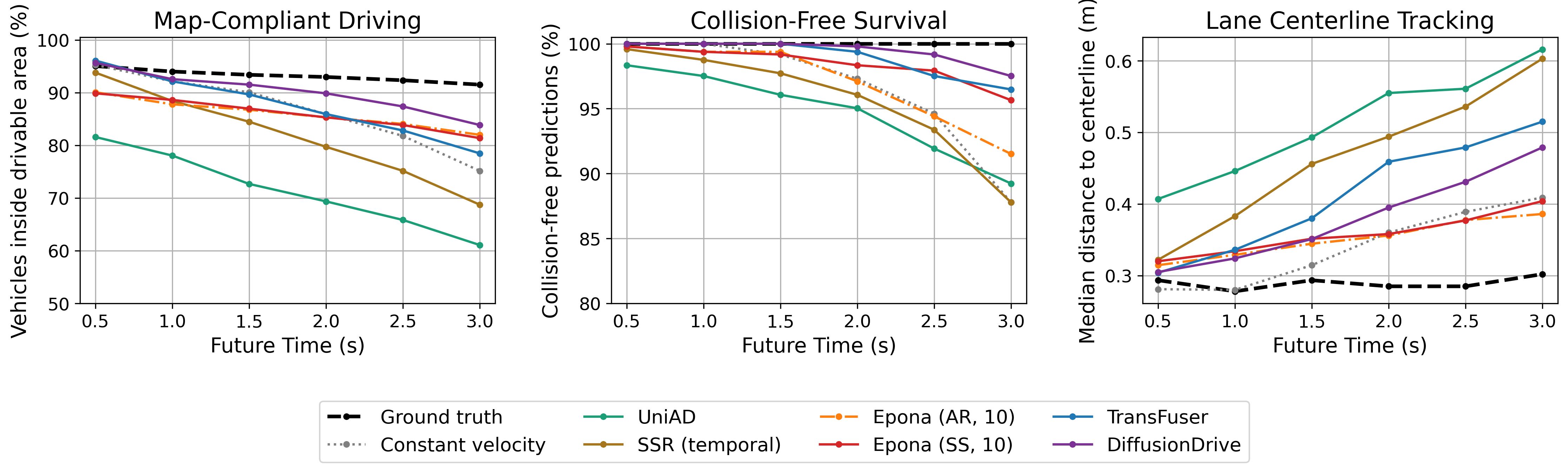}
    \caption{Per-horizon profiles of the map-grounded metrics on the 483 e2e samples for a subset of the evaluated end-to-end models, plotted over the prediction horizon. From left to right: map-compliant driving, collision-free driving against lidar-derived occupancy layer, and centerline-tracking error.}
  \label{fig:app:statistics}
\end{figure}

\paragraph{Baselines.}
We zero-shot evaluate the public checkpoints of UniAD~\cite{hu2023uniad}, DMAD~\cite{shen2025dmad}, SSR~\cite{li2025ssr}, Epona~\cite{zhang2025epona}, TransFuser and LTF~\cite{chitta2023transfuser}, and DiffusionDrive~\cite{liao2025diffusiondrive}; no fine-tuning on KITScenes Multimodal is performed. 
Their inputs differ considerably and are summarised in \Cref{tab:app:e2e_inputs}. UniAD, DMAD, and SSR consume the six ring cameras and are conditioned on a discrete navigation command, namely \texttt{turn left}, \texttt{turn right}, or \texttt{go straight}, derived from the ground-truth future trajectory over the next \SI{3}{\second}, the window used in their nuScenes training. 
Epona consumes the front-view camera only and runs without navigation commands. SSR is evaluated in two configurations: \emph{non-temp.} uses only the current keyframe whereas \emph{temporal} aggregates past BEV features. Epona is evaluated both as single-step (SS) prediction and autoregressive (AR) rollout, each at 10 and 100 diffusion denoising steps.
The NAVSIM-trained TransFuser, LTF, and DiffusionDrive see the three front-facing cameras, cropped and stitched into one panorama, and the current ego velocity, acceleration, and driving command; TransFuser and DiffusionDrive additionally rasterise the lidar sweep into a bird's-eye-view grid, whereas LTF replaces it by a learned latent.
NAVSIM's command is route-based, i.e., the next maneuver along the route without a fixed horizon; we use the \SI{5}{\second} window, the closer of our two command labels.
To bridge the different sensor rigs, every camera image is resampled into the corresponding camera of the training dataset, mimicing intrinsics and mounting orientation via a rotation homography, and lidar points are shifted vertically so that the road lies at the ground height of the NAVSIM vehicle frame.

\begin{figure*}[t]
  \centering
  \includegraphics[width=\linewidth]{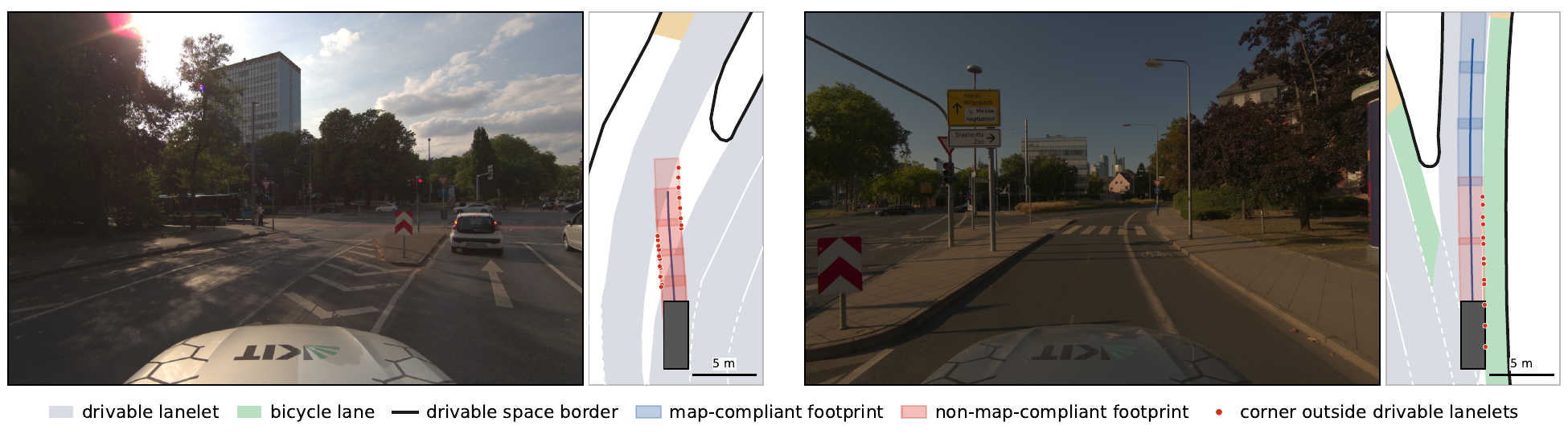}
  \caption{Recorded human drives from the e2e eval set that are not fully map-compliant yet serving as ground-truth. Each example shows the front camera and a top view in the vehicle frame with the vehicle-passable lanelets, bicycle lanes, and the drivable-space borders tagged in the HD map; and red dots mark footprint corners outside the vehicle-passable lanelets. Left: at a lane split, the vehicle crosses the unmapped gap between two diverging lanelets during a late lane change. Right: in a curve, the right corners ride up over the marking of the adjacent cycle lane.}
  \label{fig:app:map_compliance_examples}
\end{figure*}

\paragraph{Map-based and Occupancy-derived Metrics.}
All map-grounded metrics are evaluated on the predicted ego trajectory at its native output rate, with the ego footprint oriented along the predicted heading at every timestamp.
\emph{Map-compliant driving} at a given horizon is the fraction of e2e samples for which all four corners of the ego footprint stay inside the union of vehicle-passable Lanelet2 lanelets of the scene's local map at every timestamp up to that horizon; (most) bicycle lanes, crosswalks, and emergency lanes are not vehicle-passable, and the drivable-space borders tagged in the map, mostly curbs, delimit the road.
\emph{Collision-free driving} is defined analogously: a sample counts if the ego footprint never intersects the lidar-derived occupancy layer up to that horizon. The occupancy layer is built from the keyframe lidar sweep after removing the ego vehicle's own returns and compensating the ego motion during the sweep; the recorded ground truth is collision-free in all 483 samples.
\emph{Centerline distance} is the median over samples of the lateral offset between the ego centre and the closest drivable centerline at the evaluation horizon.
Per-horizon profiles for the map-grounded metrics are plotted up to \SI{3}{\second} in \Cref{fig:app:statistics}.
Map compliance is a deliberately strict criterion, and \SI{100}{\percent} is not the target: more than half of the eval samples are intersections, lane changes, night drives, or other non-nominal scenes, and human drivers leave the vehicle lanes for short moments as part of normal driving, for example by keeping onto an advisory cycle lane to pass a stopped truck or by brushing a cycle lane where the lane narrows. In additon, the lanelets do not always cover the full paved road width and some non-map-complient flags can also arise through map artifacts. 
The recorded ground truth is therefore map-compliant in \SI{91.5}{\percent} of the samples at \SI{3}{\second}. Two non-map-complient human driving scenarios from our e2e eval set are visualized in \Cref{fig:app:map_compliance_examples}.

\begin{table*}[t]
\centering
\caption{Displacement errors up to the full \SI{5}{\second} horizon on the 483 e2e samples. All models that take a driving command receive the \SI{5}{\second} command. Predictions shorter than \SI{5}{\second} are extended at constant velocity from their last two points and resampled to \SI{5}{\hertz}. Best values are bold, second-best underlined; the reference row is not ranked.}
\label{tab:app:kitscenes_5s_results}
\scriptsize
\setlength{\tabcolsep}{3.2pt}
\renewcommand{\arraystretch}{1.08}
\setlength{\aboverulesep}{0pt}
\setlength{\belowrulesep}{0pt}
\setlength{\extrarowheight}{2pt}
\resizebox{\textwidth}{!}{%
\begin{tabular}{l|ccccc|cc|cccccc}
\toprule
\multirow{2}{*}{Model}
& \multicolumn{5}{c|}{ADE $\downarrow$}
& \multicolumn{2}{c|}{FDE $\downarrow$}
& \multicolumn{6}{c}{ADE@\SI{5}{\second} by category $\downarrow$}
\\
\cmidrule{2-6}
\cmidrule{7-8}
\cmidrule{9-14}
& \SI{1}{\second} & \SI{2}{\second} & \SI{3}{\second} & \SI{4}{\second} & \SI{5}{\second}
& \SI{3}{\second} & \SI{5}{\second}
& \makecell{selec.\\(3)} & \makecell{constr.\\(3)} & \makecell{overt.\\(47)} & \makecell{inters.\\(153)} & \makecell{night\\(63)} & \makecell{nom.\\(212)}
\\
\midrule
UniAD~\cite{hu2023uniad}
& 0.91 & 1.55 & 2.30 & 3.18 & 4.15
& 4.55 & 9.03
& 4.51 & 3.58 & 3.79 & 3.57 & 5.67 & 4.22
\\
DMAD~\cite{shen2025dmad}
& 0.93 & 1.55 & 2.26 & 3.11 & 4.08
& 4.38 & 9.02
& 5.53 & 3.26 & 3.78 & 3.73 & 5.41 & 4.01
\\
SSR~\cite{li2025ssr} (non-temp.)
& 1.32 & 2.53 & 3.82 & 5.15 & 6.49
& 7.49 & 12.99
& 3.24 & 3.58 & 7.51 & 5.01 & 9.37 & 6.59
\\
SSR~\cite{li2025ssr} (temporal)
& 0.89 & 1.57 & 2.30 & 3.13 & 4.06
& 4.45 & 8.72
& 7.66 & 6.71 & 3.87 & 4.30 & 5.03 & 3.50
\\
\midrule
Epona~\cite{zhang2025epona} (AR, 10)
& 0.94 & 2.01 & 3.31 & 4.77 & 6.31
& 7.19 & 13.83
& 4.03 & 8.39 & 6.43 & 5.97 & 8.24 & 6.00
\\
Epona~\cite{zhang2025epona} (AR, 100)
& 0.76 & 1.58 & 2.60 & 3.75 & 5.00
& 5.62 & 11.16
& 4.43 & 7.01 & 5.41 & 4.64 & 6.62 & 4.69
\\
Epona~\cite{zhang2025epona} (SS, 10)
& 0.66 & 1.19 & 1.83 & 2.60 & 3.48
& 3.78 & 7.93
& 3.98 & 6.42 & 3.92 & 3.32 & 4.82 & 3.05
\\
Epona~\cite{zhang2025epona} (SS, 100)
& 0.63 & 1.17 & 1.83 & 2.62 & 3.53
& 3.82 & 8.16
& 4.25 & 5.94 & 3.98 & 3.35 & 4.85 & 3.13
\\
\midrule
TransFuser~\cite{chitta2023transfuser}
& 0.25 & 0.54 & 0.95 & \underline{1.48} & \underline{2.11}
& \underline{2.24} & \underline{5.31}
& \underline{1.76} & \underline{2.17} & 2.27 & \underline{2.20} & \underline{2.24} & \underline{1.97}
\\
LTF~\cite{chitta2023transfuser}
& \underline{0.24} & \underline{0.53} & \underline{0.94} & \underline{1.48} & 2.13
& \underline{2.24} & 5.42
& 1.89 & 2.28 & \textbf{2.01} & 2.22 & 2.48 & 1.98
\\
DiffusionDrive~\cite{liao2025diffusiondrive}
& \textbf{0.22} & \textbf{0.48} & \textbf{0.87} & \textbf{1.39} & \textbf{2.00}
& \textbf{2.11} & \textbf{5.14}
& \textbf{1.64} & \textbf{1.57} & \underline{2.02} & \textbf{2.11} & \textbf{2.16} & \textbf{1.88}
\\
\midrule
\textit{Constant velocity (ref.)}
& \textit{0.26} & \textit{0.72} & \textit{1.39} & \textit{2.25} & \textit{3.28}
& \textit{3.50} & \textit{8.50}
& \textit{3.13} & \textit{8.26} & \textit{3.20} & \textit{4.34} & \textit{2.53} & \textit{2.66}
\\
\bottomrule
\end{tabular}%
}
\end{table*}
\begin{table*}[t]
\centering
\caption{Multi-Maneuver Score~\cite{wagner2026longtail} on the 483 samples drawn from the \texttt{val}~$\cup$~\texttt{overlap-train-val} scenes of KITScenes Multimodal, broken down by scene category, evaluated using the \SI{5}{\second} horizon following \cite{wagner2026longtail}. MMS is higher-is-better; best values are bold, second-best underlined with ties sharing the mark).}
\label{tab:app:kitscenes_mms_results}
\setlength{\tabcolsep}{3.2pt}
\renewcommand{\arraystretch}{1.08}
\resizebox{0.75\textwidth}{!}{%
\begin{tabular}{lcccccccc}
\toprule
\multirow[c]{2}{*}[-0.3ex]{Model} & \multicolumn{8}{c}{MMS $\uparrow$} \\
\cmidrule(lr){2-9}
& avg & selec. & rain & constr. & overt. & inters. & night & nom. \\
\midrule
UniAD \cite{hu2023uniad}                     & 3.59               & \underline{2.83}   & 2.50               & 2.83               & 3.43               & 3.23               & 3.70               & 3.89 \\
DMAD \cite{shen2025dmad}                     & 3.62               & \textbf{5.00}      & 1.75               & \underline{3.17}   & 3.28               & 3.43               & 4.00               & 3.71 \\
SSR \cite{li2025ssr} (non-temp.)             & 3.66               & 1.54               & 3.00               & \underline{3.17}   & 3.71               & 3.46               & 3.85               & 3.79 \\
SSR \cite{li2025ssr} (temporal)              & 4.01               & \underline{2.83}   & \underline{3.50}   & \textbf{3.48}      & 4.19               & 3.68               & 3.95               & 4.25 \\
\midrule
Epona \cite{zhang2025epona} (AR, 10)         & 2.92               & 0.83               & \textbf{4.23}      & 1.87               & 3.19               & 2.46               & 3.11               & 3.16 \\
Epona \cite{zhang2025epona} (AR, 100)        & 3.18               & 0.83               & 1.25               & 1.67               & 3.30               & 2.80               & 3.38               & 3.44 \\
Epona \cite{zhang2025epona} (SS, 10)         & 3.28               & 0.83               & 1.25               & 1.67               & 3.06               & 2.58               & 3.47               & 3.85 \\
Epona \cite{zhang2025epona} (SS, 100)        & 3.26               & 0.83               & 1.25               & 1.67               & 2.95               & 2.81               & 3.33               & 3.72 \\
\midrule
TransFuser~\cite{chitta2023transfuser}       & \underline{4.45}   & 0.83               & \underline{3.50}   & \underline{3.17}   & \underline{5.24}   & 3.94               & \textbf{4.87}      & 4.60 \\
LTF~\cite{chitta2023transfuser}              & 4.43               & 1.17               & \underline{3.50}   & \underline{3.17}   & \textbf{5.27}      & \underline{3.97}   & 4.46               & \underline{4.65} \\
DiffusionDrive~\cite{liao2025diffusiondrive} & \textbf{4.61}      & 1.93               & \underline{3.50}   & 2.83               & 5.04               & \textbf{4.20}      & \underline{4.79}   & \textbf{4.83} \\
\bottomrule
\end{tabular}%
}
\scriptsize
\vspace{-3pt}
{\parbox{0.75\linewidth}{
    \vspace{1pt}
    \scriptsize
    Epona uses single-step (SS) or autoregressive (AR) rollouts with 10 or 100 diffusion denoising steps. Predictions shorter than \SI{5}{\second} (\SI{3}{\second}; \SI{4}{\second} for TransFuser, LTF, and DiffusionDrive) are linearly extrapolated to match the \SI{5}{\second} evaluation horizon of~\cite{wagner2026longtail}.
}}
\end{table*}

\paragraph{Long-horizon Results.}
\Cref{tab:app:kitscenes_5s_results} extends the displacement errors to \SI{5}{\second}. The ranking of \Cref{tab:kitscenes_e2e_results} is preserved: the NAVSIM planners stay well ahead (ADE@\SI{5}{\second} of \SIrange{2.00}{2.13}{\meter}), followed by the single-step Epona variants (\SI{3.5}{\meter}) and by SSR (temporal), DMAD, and UniAD (\SIrange{4.06}{4.15}{\meter}). The constant-velocity reference (\SI{3.28}{\meter}) again outperforms every model that does not observe the ego velocity directly. Errors roughly double between \SI{3}{\second} and \SI{5}{\second} for all models, partly because most predictions end at \SI{3}{\second} and are extrapolated beyond. Switching UniAD, DMAD, and SSR from the \SI{3}{\second} to the \SI{5}{\second} command changes the command in 41 samples and their ADE@\SI{3}{\second} by at most \SI{0.03}{\meter}.

\paragraph{Multi-Maneuver Score and Results.}
We follow the protocol of~\cite{wagner2026longtail}: each scene is self-annotated with at least three \SI{5}{\second} reference maneuvers spanning the recorded path together with scored alternative paths. MMS scores a prediction against the best-matching reference following \cite{wagner2026longtail}. Predictions shorter than \SI{5}{\second} are linearly extrapolated to \SI{5}{\second} to match the evaluation horizon. We report MMS overall and per scene category in \Cref{tab:app:kitscenes_mms_results}.

Under MMS, the NAVSIM-trained planners lead (DiffusionDrive 4.61, TransFuser 4.45, LTF 4.43), followed by the nuScenes-trained models (3.59--4.01, best SSR temporal). The command-free Epona variants score lowest (2.92--3.28), although they are competitive in map-compliant driving, likely because MMS explicitly rewards following the instructed maneuver, which Epona does not receive. 

\clearpage

\begin{figure*}[h]
  \centering
  \def\cellW{0.243\linewidth}
  \def\pairgap{0.02\linewidth} 
  \def\rowgap{0.02\linewidth} 
    \includegraphics[width=0.368\linewidth]{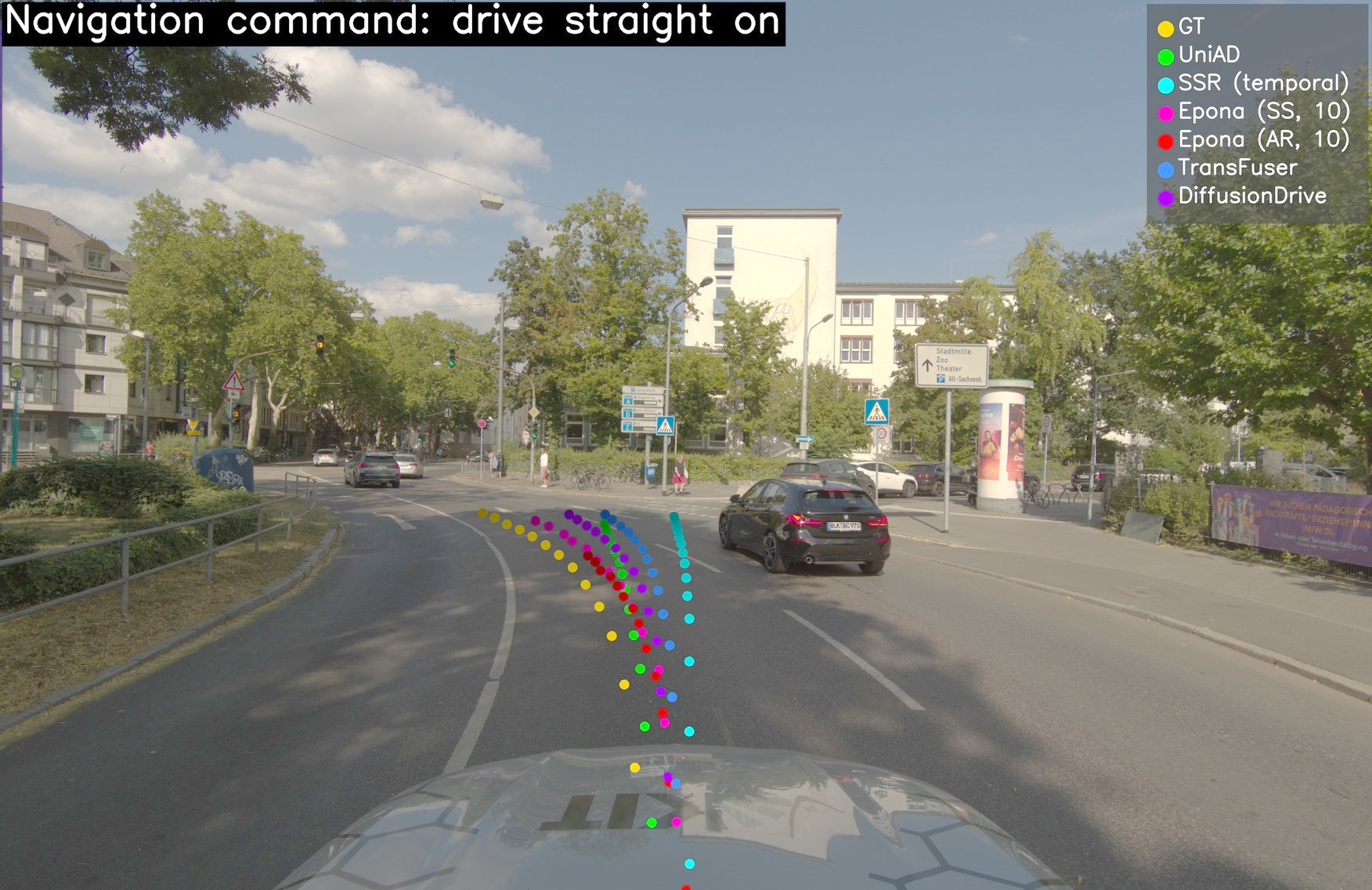}%
  \includegraphics[width=0.119\linewidth]{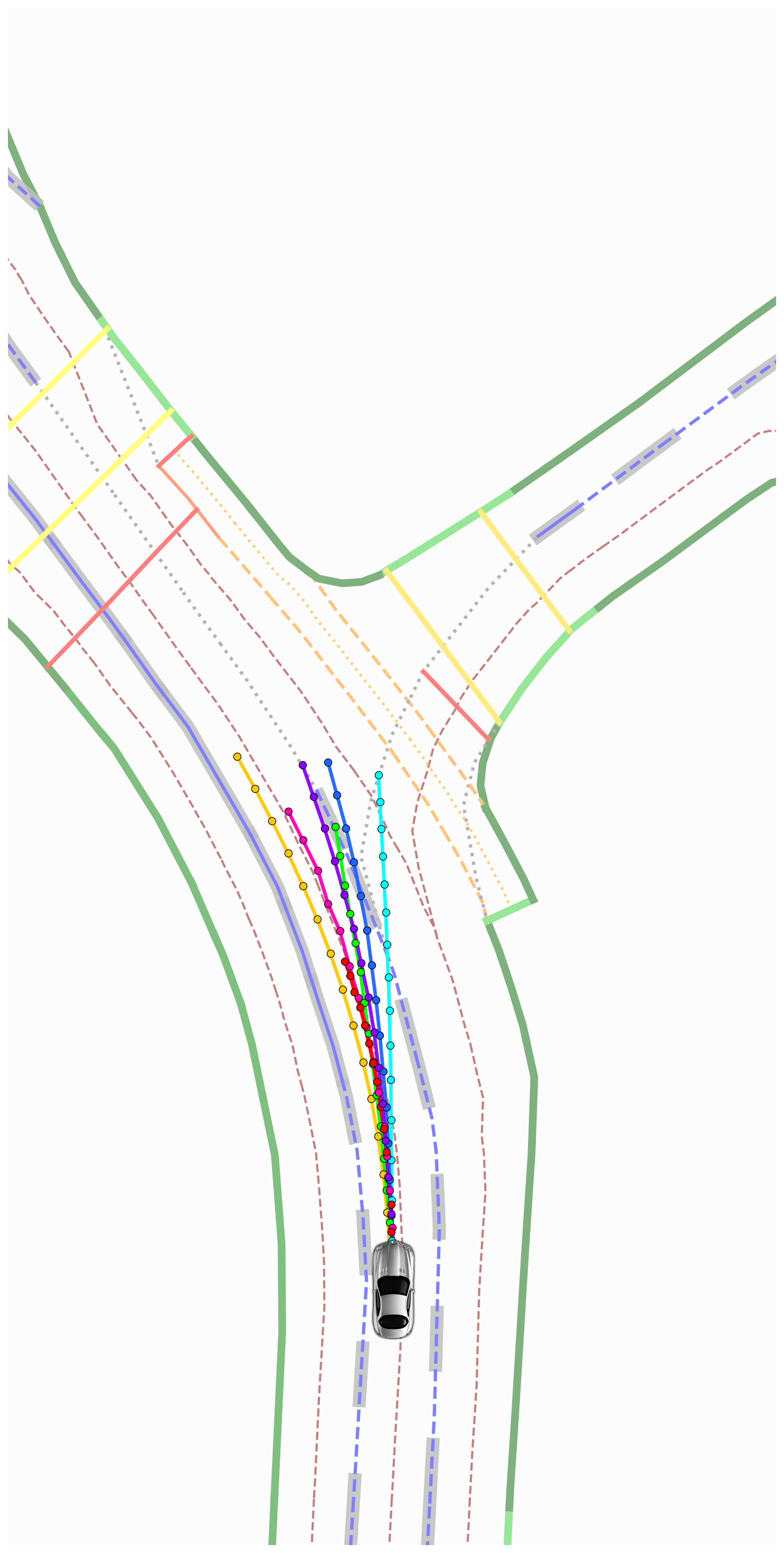}
  \hspace{\pairgap}%
  \includegraphics[width=0.368\linewidth]{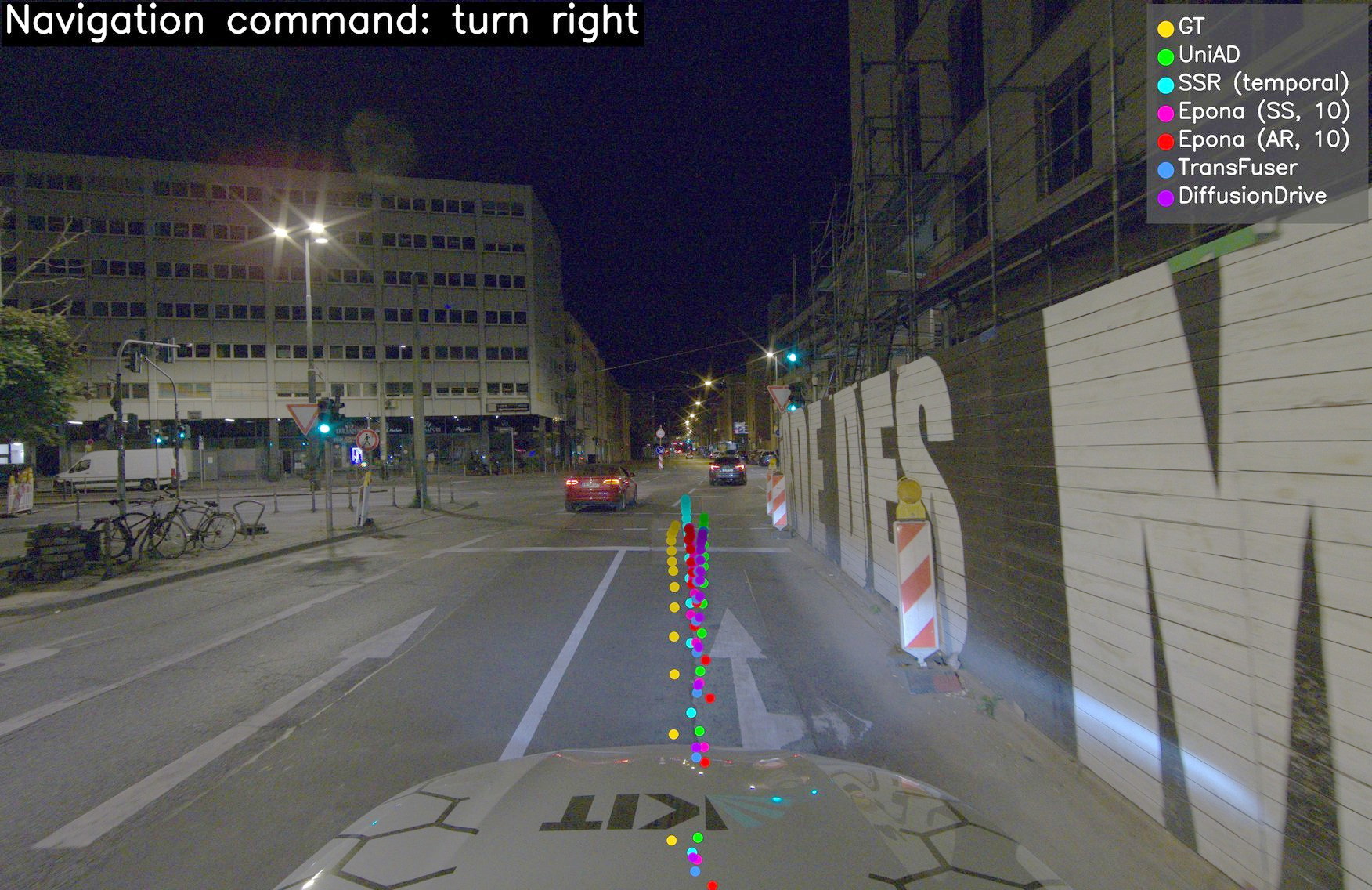}%
  \includegraphics[width=0.119\linewidth]{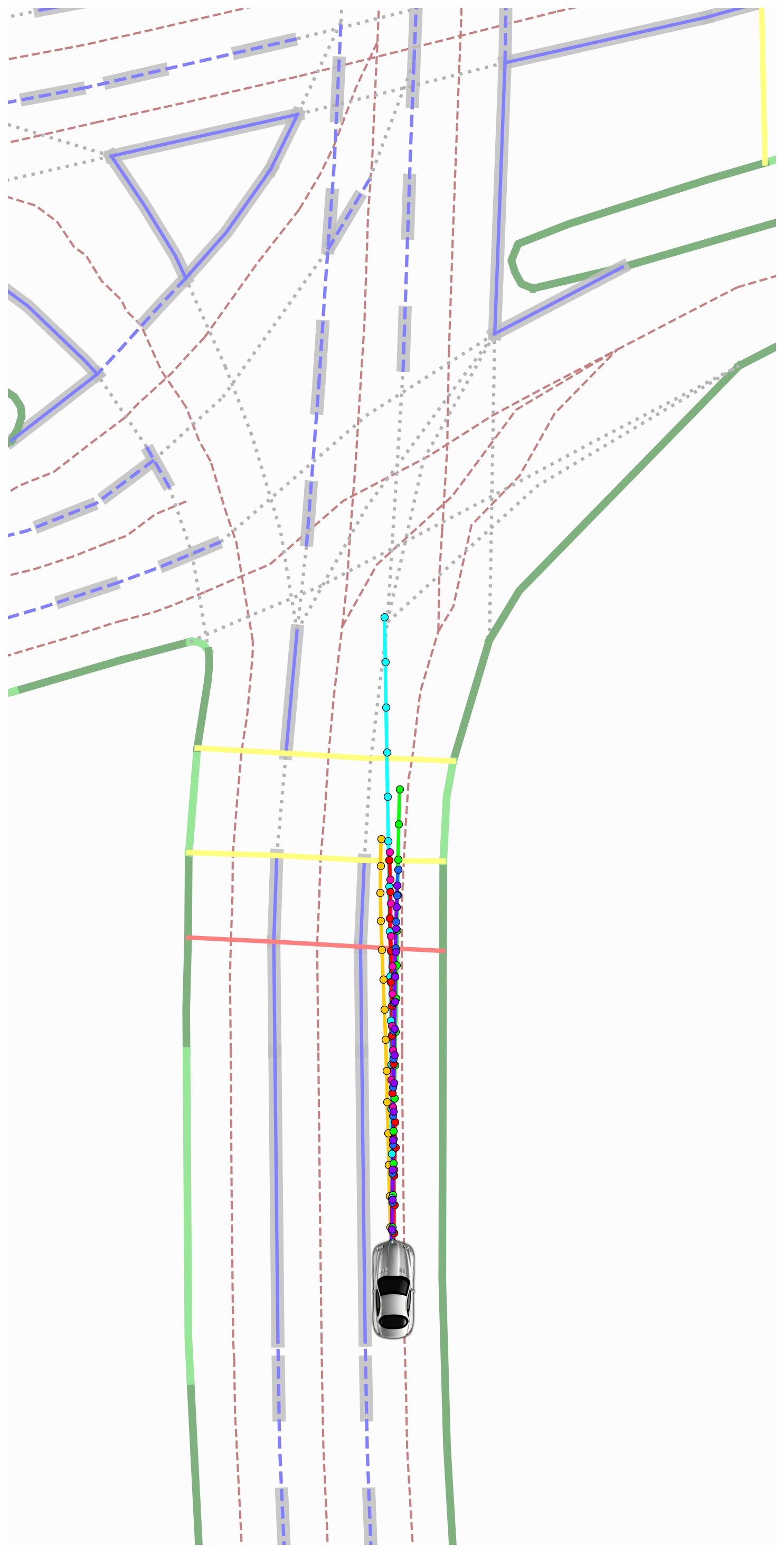}\\[\rowgap]

    \includegraphics[width=0.368\linewidth]{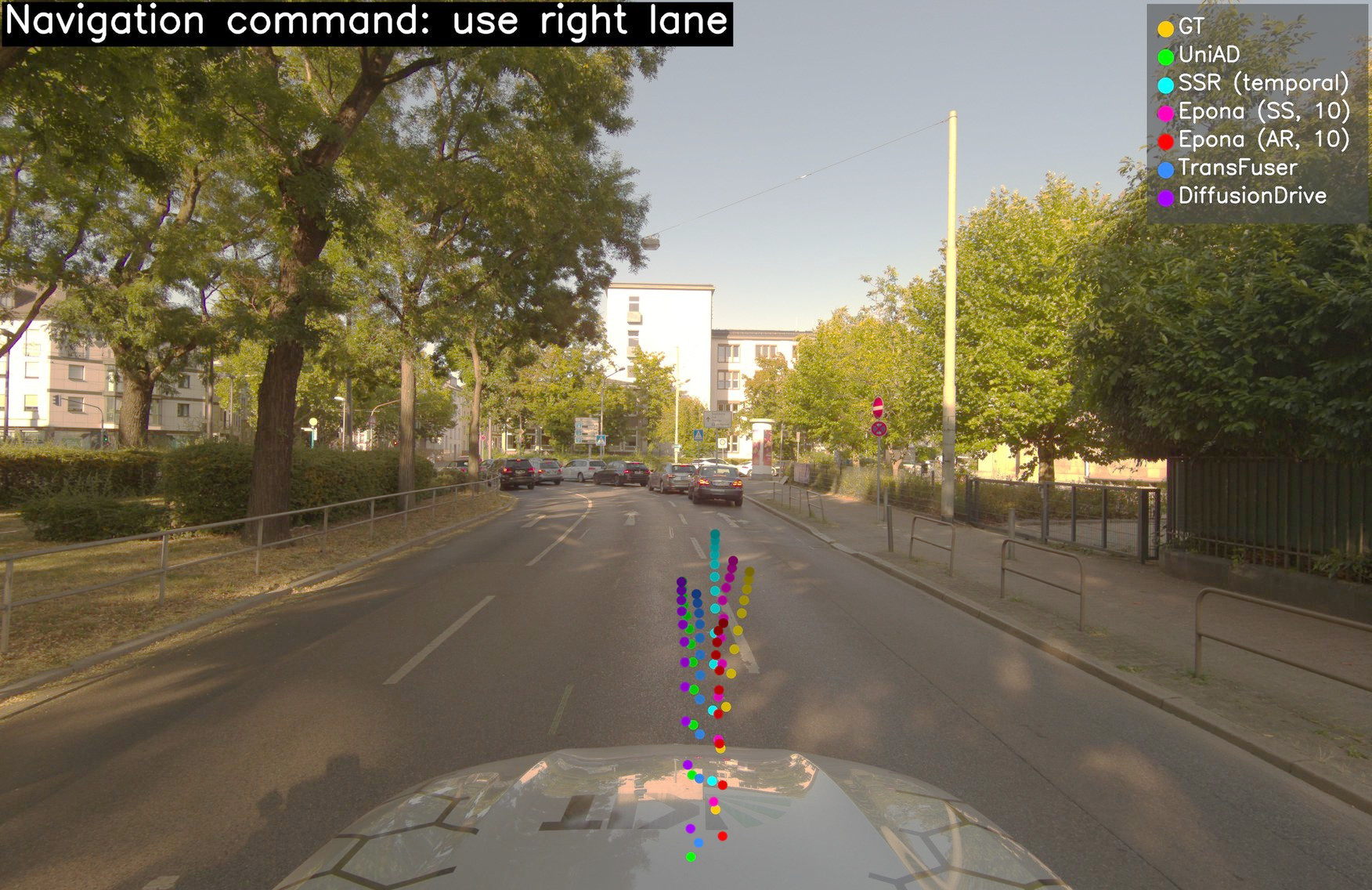}%
   \includegraphics[width=0.119\linewidth]{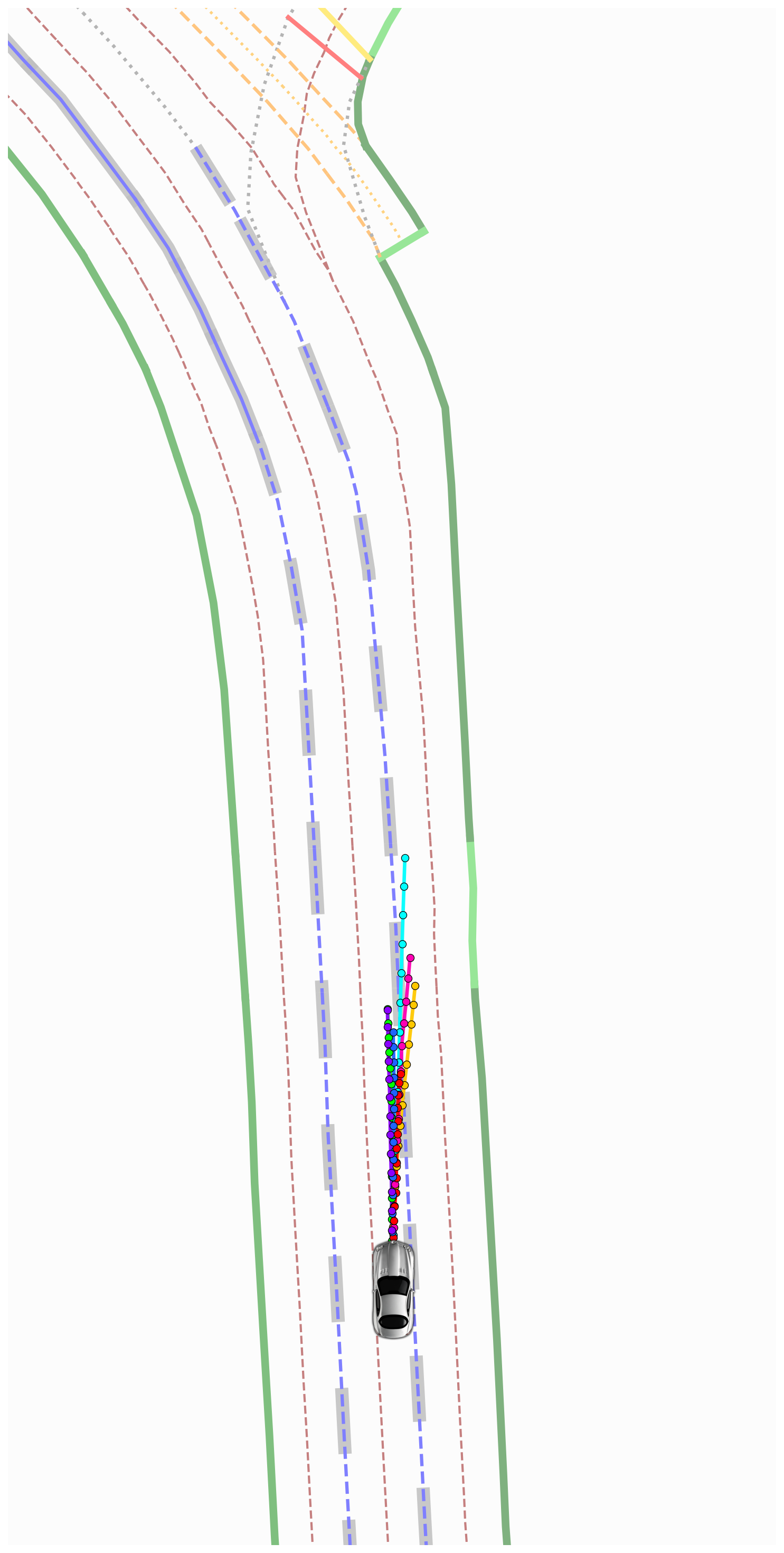}
  \hspace{\pairgap}%
  \includegraphics[width=0.368\linewidth]{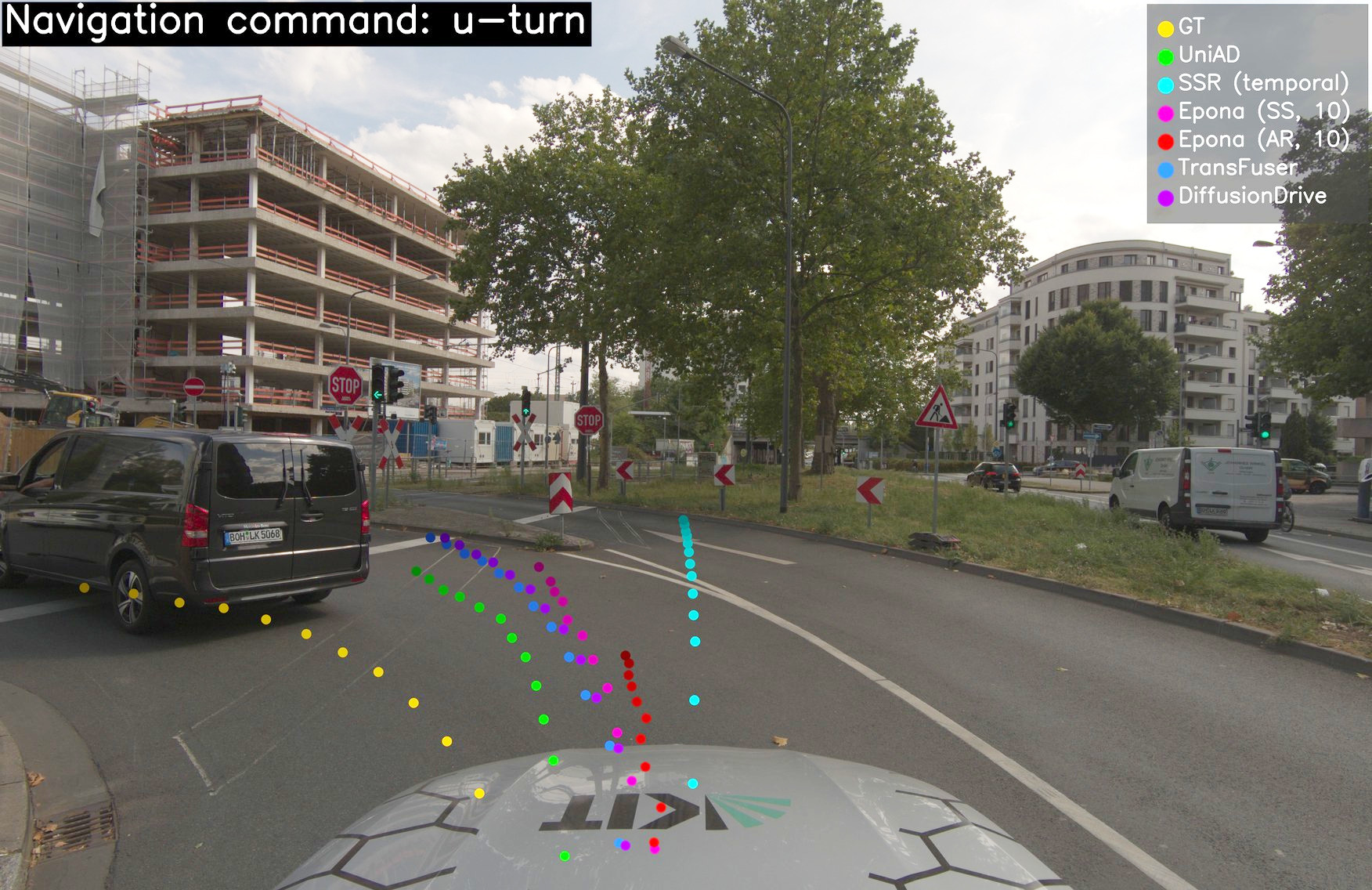}%
  \includegraphics[width=0.119\linewidth]{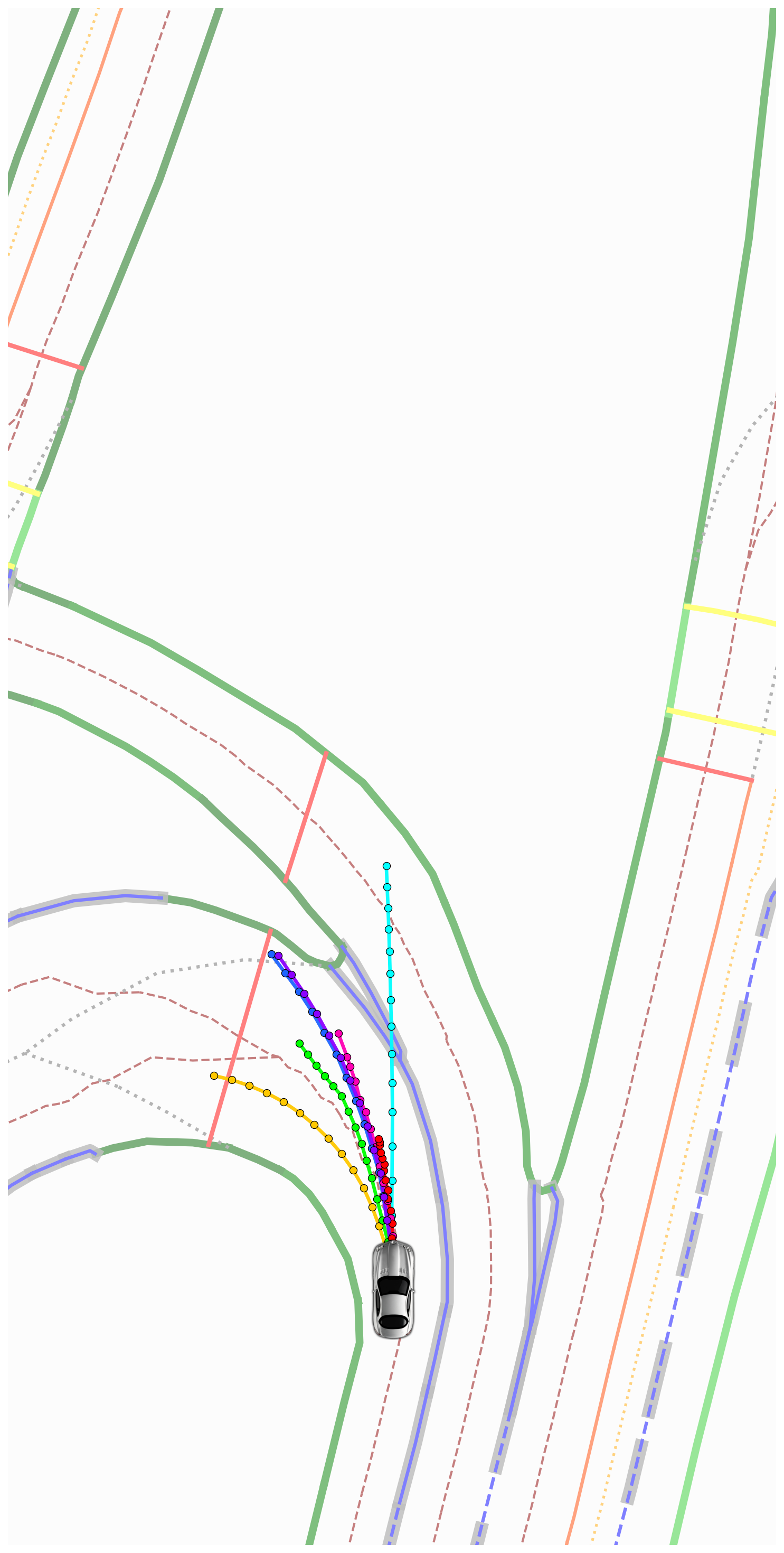}

\caption{Additional qualitative end-to-end predictions on KITScenes Multimodal, complementing \Cref{fig:en2end_quali}. Each scene pairs the front-view camera image with a top-down view of all model trajectories overlaid on the HD map and ground truth.}

  \label{fig:app:en2end_quali}
\end{figure*}

\paragraph{Additional qualitative examples.}
\Cref{fig:app:en2end_quali} shows four further scenes complementing \Cref{fig:en2end_quali}.

\FloatBarrier

\section{Compute Resources}
\label{subsec:appx:compute_resources}

The models were trained and evaluated  on 8 Nvidia  RTX 6000 Ada (StreamMapNet, SDTagNet) or 4 Nvidia A100 (MapTRv2 and MapQR).

\clearpage
\newpage
\makeatletter
\if@preprint
\else
    \section*{NeurIPS Paper Checklist}

\begin{enumerate}

\item {\bf Claims}
    \item[] Question: Do the main claims made in the abstract and introduction accurately reflect the paper's contributions and scope?
    \item[] Answer: \answerYes{} 
    \item[] Justification: The abstract and introduction summarize the main contributions of the paper: the KITScenes Multimodal dataset, its production-grade HD maps, and the four benchmark tracks. These claims are further elaborated in Sections 1-4 and scoped by the dedicated limitations discussion in Section 5.1.

\item {\bf Limitations}
    \item[] Question: Does the paper discuss the limitations of the work performed by the authors?
    \item[] Answer: \answerYes{} 
    \item[] Justification: The paper includes a dedicated Limitations subsection discussing the limitations in Section 5.1.

\item {\bf Theory assumptions and proofs}
    \item[] Question: For each theoretical result, does the paper provide the full set of assumptions and a complete (and correct) proof?
    \item[] Answer: \answerNA{} 
    \item[] Justification: The paper does not present theoretical theorems or formal proofs; its primary contributions are a new dataset, accompanying HD maps, and empirical benchmark definitions and evaluations.

    \item {\bf Experimental result reproducibility}
    \item[] Question: Does the paper fully disclose all the information needed to reproduce the main experimental results of the paper to the extent that it affects the main claims and/or conclusions of the paper (regardless of whether the code and data are provided or not)?
    \item[] Answer: \answerYes{} 
    \item[] Justification: The paper describes the dataset construction pipeline, benchmark definitions, split protocol, released assets, and per-benchmark implementation details needed to reproduce the reported results.

\item {\bf Open access to data and code}
    \item[] Question: Does the paper provide open access to the data and code, with sufficient instructions to faithfully reproduce the main experimental results, as described in supplemental material?
    \item[] Answer: \answerYes{} 
    \item[] Justification: Access links for the dataset and accompanying assets are provided through the submission materials on OpenReview. 

\item {\bf Experimental setting/details}
    \item[] Question: Does the paper specify all the training and test details (e.g., data splits, hyperparameters, how they were chosen, type of optimizer) necessary to understand the results?
    \item[] Answer: \answerYes{} 
    \item[] Justification: The manuscript specifies the dataset splits, evaluation protocols, baselines, and metrics in the main paper, while the detailed preprocessing and benchmark-specific implementation choices are documented in the Appendix.

\item {\bf Experiment statistical significance}
    \item[] Question: Does the paper report error bars suitably and correctly defined or other appropriate information about the statistical significance of the experiments?
    \item[] Answer: \answerNo{} 
    \item[] Justification: We do not provide error bars or statistical metrics of our experimental results.
Considering dataset sizes, training stability and compute requirements, established practice in sensor-related tasks for autonomous driving relies on the number of ablations and evaluation on large datasets (as is the case for our dataset).

\item {\bf Experiments compute resources}
    \item[] Question: For each experiment, does the paper provide sufficient information on the computer resources (type of compute workers, memory, time of execution) needed to reproduce the experiments?
    \item[] Answer: \answerYes{}  
    \item[] Justification: The paper describes the compute resources on which the experiments were conducted in the Appendix.
    
\item {\bf Code of ethics}
    \item[] Question: Does the research conducted in the paper conform, in every respect, with the NeurIPS Code of Ethics \url{https://neurips.cc/public/EthicsGuidelines}?
    \item[] Answer: \answerYes{} 
    \item[] Justification: To the best of our knowledge, the research conforms to the NeurIPS Code of Ethics. The paper describes privacy-preserving anonymization, a gated release process, dataset terms of use, and a takedown and maintenance procedure.

\item {\bf Broader impacts}
    \item[] Question: Does the paper discuss both potential positive societal impacts and negative societal impacts of the work performed?
    \item[] Answer: \answerNA{} 
    \item[] Justification: Due to mitigation of negative risks related to privacy by anonymization, gated access and takedown procedures we expect no positive/negative societal impact outside of the field of autonomous driving.
    
\item {\bf Safeguards}
    \item[] Question: Does the paper describe safeguards that have been put in place for responsible release of data or models that have a high risk for misuse (e.g., pre-trained language models, image generators, or scraped datasets)?
    \item[] Answer: \answerYes{}  
    \item[] Justification: All data is anonymized, including anonymization of faces and license plates and a process for errata and data-removal requests exists.

\item {\bf Licenses for existing assets}
    \item[] Question: Are the creators or original owners of assets (e.g., code, data, models), used in the paper, properly credited and are the license and terms of use explicitly mentioned and properly respected?
    \item[] Answer: \answerYes{} 
    \item[] Justification: Existing datasets, map formats, software stacks, annotation sources, and benchmark baselines are credited in the related work and method sections.

\item {\bf New assets}
    \item[] Question: Are new assets introduced in the paper well documented and is the documentation provided alongside the assets?
    \item[] Answer: \answerYes{} 
    \item[] Justification: The paper introduces new dataset assets and maps and documents them through the main paper and the associated Croissant file.

\item {\bf Crowdsourcing and research with human subjects}
    \item[] Question: For crowdsourcing experiments and research with human subjects, does the paper include the full text of instructions given to participants and screenshots, if applicable, as well as details about compensation (if any)? 
    \item[] Answer: \answerNA{} 
    \item[] Justification: The paper does not report crowdsourcing experiments or human-subject studies. The annotation effort described in the paper is in-house dataset curation.

\item {\bf Institutional review board (IRB) approvals or equivalent for research with human subjects}
    \item[] Question: Does the paper describe potential risks incurred by study participants, whether such risks were disclosed to the subjects, and whether Institutional Review Board (IRB) approvals (or an equivalent approval/review based on the requirements of your country or institution) were obtained?
    \item[] Answer: \answerNA{} 
    \item[] Justification: The paper does not describe a human-subject study or crowdsourcing protocol that would require IRB-style reporting within the scope of this checklist.

\item {\bf Declaration of LLM usage}
    \item[] Question: Does the paper describe the usage of LLMs if it is an important, original, or non-standard component of the core methods in this research? Note that if the LLM is used only for writing, editing, or formatting purposes and does \emph{not} impact the core methodology, scientific rigor, or originality of the research, declaration is not required.
    \item[] Answer: \answerNA{} 
    \item[] Justification: LLMs are not an important, original, or non-standard component of the dataset construction, benchmark design, or scientific methodology presented in this work.

\end{enumerate}
\fi
\makeatother

\end{document}